\documentclass[letterpaper,11pt]{article}
\usepackage{parskip}
\usepackage[english]{babel}
\usepackage[nottoc]{tocbibind}

\usepackage[square]{natbib}

\usepackage{diagbox}
\usepackage{makecell}
\usepackage{booktabs}
\usepackage{tikz}
\usepackage{amsmath}
\usepackage{amsfonts}
\usepackage{amssymb}
\usepackage{amsthm}
\usepackage{url,ifthen}
\usepackage{enumitem}
\usepackage{srcltx}
\usepackage{multirow}
\usepackage{boxedminipage}
\usepackage[margin=1.1in]{geometry}
\usepackage{nicefrac}
\usepackage{xspace}
\usepackage{graphicx}
\usepackage{color}
\usepackage{colortbl}
\usepackage{setspace}
\usepackage{soul}

\usepackage{hyperref}
\usepackage{cleveref}
\usepackage{mathptmx}

\definecolor{DarkGreen}{rgb}{0.1,0.5,0.1}
\definecolor{DarkRed}{rgb}{0.5,0.1,0.1}
\definecolor{DarkBlue}{rgb}{0.1,0.1,0.5}
\definecolor{Gray}{rgb}{0.2,0.2,0.2}

\usepackage{listings}
\lstdefinestyle{mystyle}{
    commentstyle=\color{DarkBlue},
    keywordstyle=\color{DarkRed},
    numberstyle=\tiny\color{Gray},
    stringstyle=\color{DarkGreen},
    basicstyle=\footnotesize,
    breakatwhitespace=false,         
    breaklines=true,                 
    captionpos=b,                    
    keepspaces=true,                 
    numbers=left,                    
    numbersep=5pt,                  
    showspaces=false,                
    showstringspaces=false,
    showtabs=false,                  
    tabsize=2
}
\usepackage[small]{caption}
\usepackage{hyperref}
\hypersetup{
    unicode=false,          
    pdftoolbar=true,        
    pdfmenubar=true,        
    pdffitwindow=false,      
    pdfnewwindow=true,      
    colorlinks=true,       
    linkcolor=DarkGreen,          
    citecolor=DarkGreen,        
    filecolor=DarkRed,      
    urlcolor=DarkBlue,          
    pdftitle={},
    pdfauthor={},
}

\def\draft{1}

\def\submit{0}

\ifnum\draft=1 
    \def\ShowAuthNotes{1}
\else
    \def\ShowAuthNotes{0}
\fi

\ifnum\submit=1
\newcommand{\forsubmit}[1]{#1}
\newcommand{\forreals}[1]{}
\else
\newcommand{\forreals}[1]{#1}
\newcommand{\forsubmit}[1]{}
\fi

\ifnum\ShowAuthNotes=1
\newcommand{\authnote}[2]{{ \footnotesize \bf{\color{DarkRed}[#1's Note:
{\color{DarkBlue}#2}]}}}
\else
\newcommand{\authnote}[2]{}
\fi

\newtheorem{theorem}{Theorem}[section]
\newtheorem{remark}[theorem]{Remark}
\newtheorem{lemma}[theorem]{Lemma}

\theoremstyle{definition}
\newtheorem{definition}[theorem]{Definition}
\newtheorem{assumption}[theorem]{Assumption}
\newtheorem{example}[theorem]{Example}

\newtheoremstyle{example_contd}
{\topsep} {\topsep}%
{}
{}
{\bfseries}
{.}
{1em}
{\thmname{#1} \thmnumber{ #2}\thmnote{#3} (continued)}

\theoremstyle{example_contd}

\newcommand{\chapterref}[1]{\hyperref[ch:#1]{Chapter~\ref{ch:#1}}}

\newcommand{\claimref}[1]{\hyperref[claim:#1]{Claim~\ref{claim:#1}}}

\newcommand{\corollaryref}[1]{\hyperref[cor:#1]{Corollary~\ref{cor:#1}}}

\newcommand{\definitionref}[1]{\hyperref[def:#1]{Definition~\ref{def:#1}}}

\newcommand{\equationref}[1]{\hyperref[eq:#1]{Equation~\ref{eq:#1}}}

\newcommand{\factref}[1]{\hyperref[fact:#1]{Fact~\ref{fact:#1}}}

\newcommand{\figureref}[1]{\hyperref[fig:#1]{Figure~\ref{fig:#1}}}

\newcommand{\tableref}[1]{\hyperref[tab:#1]{Table~\ref{tab:#1}}}

\newcommand{\itemref}[1]{\hyperref[item:#1]{Item~(\ref{item:#1})}}

\newcommand{\lemmaref}[1]{\hyperref[lem:#1]{Lemma~\ref{lem:#1}}}

\newcommand{\propref}[1]{\hyperref[prop:#1]{Proposition~\ref{prop:#1}}}

\newcommand{\propositionref}[1]{\hyperref[prop:#1]{Proposition~\ref{prop:#1}}}

\newcommand{\remarkref}[1]{\hyperref[rem:#1]{Remark~\ref{rem:#1}}}

\newcommand{\sectionref}[1]{\hyperref[sec:#1]{Section~\ref{sec:#1}}}

\newcommand{\theoremref}[1]{\hyperref[thm:#1]{Theorem~\ref{thm:#1}}}

\usepackage[utf8]{inputenc}
\usepackage[T1]{fontenc}
\usepackage{kpfonts}
\usepackage{microtype}

\renewcommand{\hat}{\widehat}

\newcommand{\cA}{{\cal A}}

\newcommand{\cC}{{\cal C}}

\newcommand{\cG}{{\cal G}}
\newcommand{\cH}{{\cal H}}

\newcommand{\cL}{{\cal L}}

\newcommand{\cO}{{\cal O}}
\newcommand{\cP}{{\cal P}}

\newcommand{\cR}{{\cal R}}
\newcommand{\cS}{{\cal S}}

\newcommand{\cX}{{\cal X}}
\newcommand{\cY}{{\cal Y}}
\newcommand{\cZ}{{\cal Z}}

\renewcommand{\leq}{\leqslant}
\renewcommand{\le}{\leqslant}
\renewcommand{\geq}{\geqslant}
\renewcommand{\ge}{\geqslant}

\usepackage{bm}

\newcommand{\ignore}[1]{}

\renewcommand{\epsilon}{\varepsilon}

\newcommand{\remove}[1]{}

\usepackage{color-edits}
\addauthor{hui}{cyan}

\newcommand{\bE}{\mathbb{E}}

\usepackage{bm}
\usepackage{amsfonts}
\usepackage[utf8]{inputenc} 
\usepackage[T1]{fontenc}    
\usepackage{hyperref}       
\usepackage{url}            
\usepackage{booktabs}       
\usepackage{amsfonts}       
\usepackage{nicefrac}       
\usepackage{microtype}      
\usepackage{xcolor}         

\usepackage{algorithm}
\usepackage{algpseudocode}
\usepackage{diagbox}
\usepackage{makecell}
\usepackage{booktabs}
\usepackage{tikz}
\usepackage{amsmath}
\allowdisplaybreaks
\usepackage{amssymb}
\usepackage{amsthm}
\usepackage{url,ifthen}
\usepackage{enumitem}
\usepackage{srcltx}
\usepackage{multirow}
\usepackage{boxedminipage}
\usepackage{nicefrac}
\usepackage{xspace}
\usepackage{graphicx}
\usepackage{array,booktabs,longtable}
\usepackage{color}
\usepackage{colortbl}
\usepackage{setspace}
\usepackage{soul}
\usepackage{graphicx}
\usepackage{wrapfig}
\usepackage{subcaption}
\usepackage{enumitem}
\usepackage{cleveref}
\definecolor{DarkGreen}{rgb}{0.1,0.5,0.1}
\definecolor{DarkRed}{rgb}{0.5,0.1,0.1}
\definecolor{DarkBlue}{rgb}{0.1,0.1,0.5}
\definecolor{Gray}{rgb}{0.2,0.2,0.2}

\title{OmniMechanism Design for Human-AI Collaboration with Impressionable Minds}

\usepackage{authblk}

\author[1]{Tianwei Gao}
\author[1]{Yifei Pang}
\author[2]{Lunjia Hu}
\author[3]{Kate Donahue}
\author[4]{Lihua Lei}
\author[5]{Vahid Tarokh}
\author[1]{Zhun Deng}
\affil[1]{University of North Carolina at Chapel Hill}
\affil[2]{Northeastern University}
\affil[3]{University of Illinois at Urbana-Champaign}
\affil[4]{Stanford University}
\affil[5]{Duke University}
\date{}

\begin{document}

\maketitle

\begin{abstract}
As AI systems are increasingly deployed in high-stakes domains such as healthcare, effectively integrating human expertise with AI outputs is critical for reliable decision-making. A substantial body of theoretical work studies optimal mechanisms for human-AI collaboration (HAC), focusing on when to involve humans and how to combine their expertise with AI predictions. However, these approaches largely overlook a key factor highlighted by empirical evidence: human judgments are impressionable and can vary depending on how AI outputs are presented. For example, when AI systems defer decisions to humans, revealing or withholding the AI prediction can significantly influence human judgments. Moreover, existing work neglects deployment-time variability, where conditions, particularly the cost of eliciting human input, may differ substantially from those observed during training.
 To address these challenges, we introduce a rigorous and theoretically grounded framework for human-AI collaboration, termed \emph{collaborative omnimechanism design}, based on a novel and computationally efficient contextual bandit variant of boosting for outcome indistinguishability.
 In particular, we reduce a calibration-type outcome indistinguishability condition to weighted calibration, providing a novel time-efficiency result, bypassing the high-dimension curse induced by complicated collaboration methods and outcome space.
 Our framework provides a principled characterization of human impressionability and introduces a unified optimization scheme that enables robust and adaptable decision-making across diverse query costs and evaluation objectives without retraining.  We evaluate the framework in three human--AI collaboration settings, examining adaptation to query costs and prediction objectives without retraining. Across these settings, our omnimechanism induces policies that achieve performance comparable to or better than the evaluated baseline policies.

\end{abstract}

\section{Introduction}

Despite rapid advances in machine learning, human judgment remains essential in many high-stakes decision-making tasks. In domains such as healthcare, AI systems can provide accurate predictions, yet often lack access to the full context available to human experts. Conversely, human decision-makers can incorporate rich contextual information but may benefit from algorithmic guidance. This complementary structure has led to growing interest in human-AI collaboration (HAC) as a framework for combining the strengths of both \citep{jacobs2021machine,bondi2022role,geifman2017selective,bansal2019beyond,alur2023auditing,agarwal2023combining,feng2024large,alur2024human,agarwal2025designing}.

A key but often overlooked aspect of human-AI collaboration is the \textit{impressionability} of human decision-makers. In HAC settings, when an AI system defers a task to a human while providing a prediction or auxiliary signal, this information can influence the human’s perception of the task and alter their subsequent behavior and final decision. For example, \cite{bondi2022role} show that, when deferring a task to a human, it is possible to improve human performance by informing the human of the deferral decision without revealing the AI prediction. \cite{vodrahalli2022uncalibrated} further demonstrate that manipulating the displayed confidence score of an AI system can alter subsequent human judgments. More recently, \cite{bo2025rely} reveal that, in collaboration with large language models, different reliance interventions can significantly affect both decision quality and the time required for users to engage with AI advice and answer questions. We use \textbf{impressionability} to mean that a human's response distribution depends on how AI advice is presented. This impressionability creates a fundamental challenge: \textbf{\textit{the presentation of AI outputs becomes a decision variable to optimize, while the resulting distribution of human responses depends on this choice}}. This dependence is reminiscent of performative prediction \citep{perdomo2020performative,lin2023plug,li2025statistical}. Despite growing empirical evidence of such effects \citep{bansal2019beyond,bondi2022role,li2024understanding,behzad2025beyond,bo2025rely,duetting2025information}, many theoretical HAC models treat human responses as independent of the presentation of AI advice.



Another fundamental challenge in human–AI collaboration arises from deployment-time variability. In practice, the downstream environment in which a collaboration mechanism is used can differ significantly from the conditions assumed during training. In particular, \textbf{\textit{the evaluation criteria and the cost of querying human expertise may vary across applications, time, and resource constraints}} \citep{mozannar2020consistent,mao2023two,alves2024cost,casacuberta2025selective}. For example, in a clinical setting, the cost of consulting a specialist may vary depending on workload or urgency, while the evaluation criterion may shift between minimizing diagnostic error and prioritizing high-risk patients, depending on the application. Moreover, once a system is deployed, access to training data is often limited or unavailable, making it impractical to retrain the model for each new setting. Addressing this challenge requires frameworks that can adapt collaboration policies to varying deployment-time prediction objectives and human-query costs without retraining the learned outcome model. However, existing approaches typically optimize for a single objective and do not directly support such adaptation \cite{mozannar2020consistent,mao2023two,verma2023learning,ramgolam2025learning}.



\textbf{Our contributions.} To address the challenges of human impressionability and deployment-time variability, we introduce a unified theoretical framework for human-AI collaboration based on \emph{collaborative omnimechanism design}. Our main contribution is the theoretical foundation of \emph{Collaborative Outcome Indistinguishability (COI)}, a loss-matching condition underlying a computationally tractable omniprediction framework tailored for human-AI collaboration with impressionable human minds. Formally, the COI condition ensures that expected losses under our learned predictive model closely match those under the true data-generating environment for the policies in a prescribed comparison class and the policies induced by the learned model across supported loss-cost pairs. This enables a central planner to select a policy whose expected loss is close to that of the best policy in the comparison class for each supported loss-cost pair, without retraining or requiring an explicit parametric model of impressionable human behavior. The core technical contributions of our work are as follows:



 (a). \textbf{Efficient COI learning via boosting.} We extend the omniprediction framework to the HAC setting and propose an efficient boosting algorithm, COI-Boost, that achieves outcome indistinguishability using a reliable auditor. We establish finite-sample guarantees for both the auditor and the boosting procedure, and validate the effectiveness of this approach empirically. Further, we reduce Decision-OI auditing to weighted calibration and obtain a computational guarantee under the coverage and smoothing assumptions. We also derive a sample-complexity lower bound for offline omnimechanism learning.



(b). \textbf{Capturing impressionable human behavior.} We do not posit a parametric model of human responses or require pointwise recovery of their true distribution. Instead, we learn a surrogate outcome model whose expected losses agree with those of the environment for the relevant policies and objectives under standard coverage and complexity assumptions. This enables near-optimal decision-making without requiring a precise model of human behavior.



(c). \textbf{Extensions to richer collaboration settings.} We extend our framework to settings with multiple human experts and to feature-based methods, demonstrating its flexibility and applicability to more complex human-AI collaboration scenarios.

(d). \textbf{Evaluation across decision regimes.} We evaluate cost adaptation with human responses generated by a simulator fitted to real interaction data in binary classification, observed radiologist--AI interactions, and house-price prediction. Together, these experiments test whether a learned predictive model can support input-dependent collaboration policies across distinct decision regimes and adapt to different costs without retraining.

\section{Setup }
In this section, we formally introduce our mathematical models and desiderata.

\textbf{Basic mechanism design with a single loss.} For clarity, we first formulate collaboration for a fixed prediction loss and human-query cost, then extend the objective to a family of losses and costs.


\begin{figure}[!ht]
  \centering
  \includegraphics[width=\textwidth]{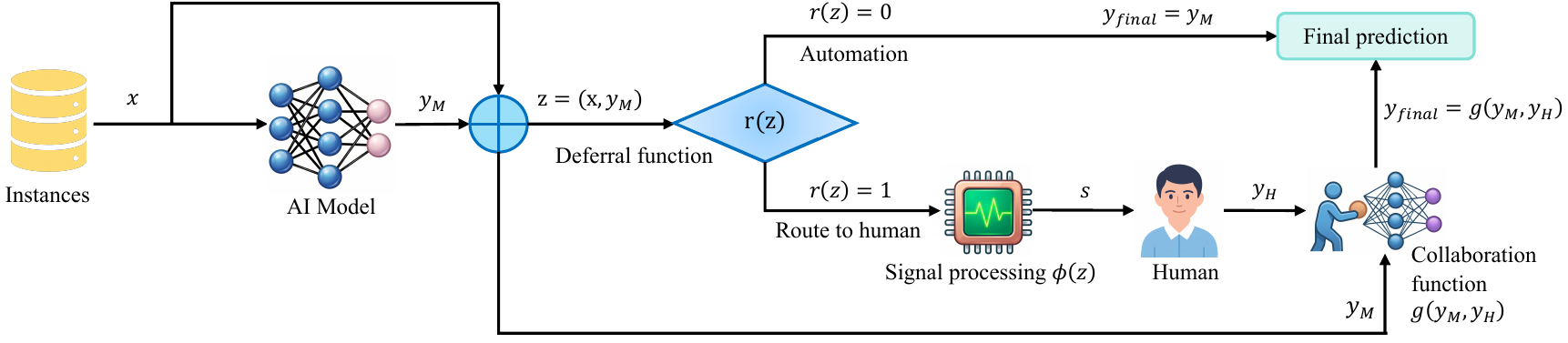}
     \vspace{-1.5em}
    \caption{OmniMechanism setup and workflow diagram.}
    \label{fig:workflow}
  \vspace{-0.5em}
\end{figure}



As illustrated in Figure~\ref{fig:workflow}, for an input $x\in \mathcal{X}$, a black-box AI model $M$ produces a prediction $y_M\in \mathcal{Y}$. In human-AI collaboration, given the context $z=(x,y_M)$, the mechanism first decides whether to request human input using a deferral rule $r:\cZ\rightarrow \{0,1\}$, where $\cZ=\cX\times\cY$. This leads to two cases:



When $r(z)=0$, the process is fully automated, and we do not rely on any human input; the AI model's prediction $y_M$ serves as the final output. 

When $r(z)=1$, our mechanism presents the transformed (possibly augmented, curated, or reframed) AI output, which we also refer to as the signal, $s=\phi(z)\in\cS$, to a human decision-maker, resulting in a human output $y_H\in \cY$. Here, the function $\phi:\cZ\to \cS$ denotes the signal-presentation rule. Examples include the parameterized  confidence score manipulation function  \citep{vodrahalli2022uncalibrated} and  different types of reliance intervention \citep{bo2025rely} mentioned in the introduction. Importantly, because the human decision-maker is influenced by the presented signal, \textbf{\textit{different choices of the signal $s$ induce different distributions over the human output}}. Our mechanism then combines the AI model's prediction and the human output through a collaboration function $g$, yielding the final output $g(y_M,y_H)\in\cY$. In this process, we must choose both $\phi$ and $g$ in order to optimize the quality of the final output.  Let $\cR$, $\Phi$, and $\cG$ denote prescribed classes of deferral rules $r:\cZ\to\{0,1\}$, signal-presentation rules $\phi:\cZ\to\cS$, and collaboration functions $g:\cY\times\cY\to\cY$, respectively.



In both cases, the final output quality is evaluated against the ground truth $y_{\text{true}} \in \cY_{\text{true}}$ corresponding to input $x$. For notational convenience, we write the final prediction as
$y_{\text{final}}$, which is set to be $y_M$ when $r(z)=0$, and $g(y_M,y_H)$ when $r(z)=1$. We define $\cH=\{(r,\phi,g)|r\in\cR,\phi\in\Phi, g\in\cG\}$. Each triple $h=(r,\phi,g)\in\cH$ specifies a collaboration mechanism consisting of a deferral rule, a signal-presentation rule, and a collaboration function. Our objective must account for both the cost of human querying and the quality of the final prediction. Thus, given a fixed function pair $\omega=(\tilde{l},c)$, where loss $\tilde{l}$ measures prediction error and cost function $c(x)$ represents the cost of querying the human on input $x$, we consider the following total loss at $h\in\cH$:  
\[
\ell^{\omega}(z, y_H, y_{\text{true}};h) 
= (1 - r(z))\, \tilde{l}(y_M, y_{\text{true}}) 
+ r(z)\, \big[ \tilde{l}(g(y_M, y_H), y_{\text{true}}) + c(x) \big].
\]


Here, the expected loss depends on $r,g$ and most importantly, \textit{depends on the signal through the distribution of $y_{H}$}. We also note that, in general, $\cY_{\text{true}}$ may differ from $\cY$.  For example, $\mathcal Y_{\mathrm{true}}=\{0,1\}$ may represent binary ground-truth labels, while $\mathcal Y=[0,1]$ represents probability-valued predictions.



\begin{example}[Illustration of Notations]
\emph{Given a patient’s medical image and clinical records $x$, a black-box AI model first produces a diagnosis prediction $y_M$. The system then decides whether to rely directly on the AI prediction or defer the case to a physician. If the task is deferred, the physician is shown an additional signal $s=\phi(z)$ derived from the AI output, such as a displayed confidence score or a reliability warning. For example, presenting a highly confident AI prediction may increase the physician’s reliance on the AI recommendation, while a cautionary intervention may encourage more careful review. Because human judgments are impressionable, different presentations can lead to different physician decisions. The final diagnosis is then produced by combining the AI prediction and the physician’s assessment through a collaboration function $g$ \citep{alur2024human}.}
\end{example}

From the previous discussion, we see that, to design an (optimal) mechanism for human-AI collaboration with impressionable human decision-makers, given a black-box AI model, we need to choose a deferral function $r\in\cR$, a signal-presentation rule $\phi\in\Phi$, and a collaboration function $g\in\cG$. In a typical supervised learning framework with a fixed prediction loss $\tilde{l}$ and a cost function $c$, the goal will be achieving
$$\min_{h\in\cH}\bE_{z,y_H,y_{\text{true}}}\ell^{\omega}(z, y_H, y_{\text{true}};h),$$
The expectation is taken under the distribution induced by the candidate signal presentation rule $\phi$, whose effect on human responses is unknown, while the context distribution is held fixed throughout.





\textbf{Our ultimate goal: omnimechanism design for a loss family.} As motivated in the introduction, in human-AI collaboration, in many cases, neither the prediction loss $\tilde{l}$ nor the querying cost $c$ is fixed or known \emph{a priori}, as both depend on the downstream task, resource constraints, and deployment context. To address this, we adopt an omniprediction perspective \cite{gopalan2021omnipredictors,gopalan2022loss,dwork2021outcome,dwork2022beyond} and seek a mechanism design that \textbf{\textit{simultaneously}} performs well across a class of losses $\tilde{l}$ and cost functions $c$ without retraining, ensuring robustness and broad applicability without needing to tailor the model to a specific downstream objective.



Let us consider a family of prediction losses $\cL$ and a family of query-cost functions  $\cC$, and let $\Omega=\cL\times \cC$. We further denote $\Delta(T)$ as the set of probability distributions over a set $T$. Inspired by the omniprediction theory \citep{casacuberta2025selective,gopalan2021omnipredictors}, we learn a \textit{\textbf{base predictive distribution}} ${p}:\mathcal{Z}\times \{0,1\}\times \mathcal{S}\times \mathcal{G}\to \Delta{(\mathcal{Y}\times\mathcal{Y}_{\text{true}})}$ modeling the joint distribution of the resulting final prediction and the ground-truth label, such that, for any pair $\omega=(\tilde{l},c)\in \Omega$ of loss and cost, there is a data-free post-processing map $k^{\omega}$ satisfying $[k^{\omega}({p})](z)=({r}^{\omega}(z),{\phi}^{\omega}(z),{g}^{\omega}(z,\cdot,\cdot))$, which maps from $\cZ$ to $\{0,1\}\times\mathcal{S}\times\mathcal{G}$ (i.e., a tuple specifying a deferral decision, signaling strategy, and a $z$-dependent collaboration function), thereby defining 
$$\ell^{\omega}(z, y_H, y_{\text{true}};k^{\omega}({p}))= (1 - r^{\omega}(z))\, \tilde{l}(y_M, y_{\text{true}}) 
+ r^{\omega}(z)\, \big[ \tilde{l}(g^{\omega}(z, y_M, y_H), y_{\text{true}}) + c(x) \big],$$
where human output $y_H$ implicitly depends on $\phi^{\omega}(z)$. We highlight a notational distinction: while any base collaboration function $g \in \mathcal{G}$ operates exclusively on the predictions, mapping $\mathcal{Y} \times \mathcal{Y} $ to $ \mathcal{Y}$, our post-processing map extracts the collaboration function rule pointwise using the base predictive distribution $p(z)$ for each state $z$. Therefore, for each context \(z\), the selected collaboration function \(g^\omega(z,\cdot,\cdot)\) belongs to \(\mathcal G\) while the selected function may vary with \(z\). 





Loosely speaking, for any $\varepsilon>0$, we say that $p$ is an $(\Omega, \mathcal{H}, \varepsilon)$-\textit{\textbf{collaborative omnimechanism}}
if for any loss-cost pair $\omega\in \Omega$,
\begin{equation*}
    \mathbb{E}_{z,y_H,y_{\text{true}}}\ell^\omega(z, y_H, y_{\text{true}};k^\omega(p))\leq \min_{h\in\cH}\bE_{z,y_H,y_{\text{true}}}\ell^{\omega}(z, y_H, y_{\text{true}};h) + \varepsilon.
\end{equation*}
The guarantee compares the induced policy with the best mechanism in $\mathcal H$. The induced policy can be more expressive because it may choose a different aggregation rule for each context. We remark that the left hand side expectation uses the response distribution induced by  the signal-presentation rule $\phi^\omega$ while each right-hand expectation uses the distribution induced by the corresponding prescribed base policy's signal-presentation rule.



Our goal is to develop an omnimechanism that can simultaneously achieve approximate optimality across different choices of $\tilde{l}$ and $c$ through simple, data-free postprocessing, without requiring retraining. For clarity, this section describes deterministic post-processing. Section~3.1 introduces a smoothed randomized version and Theorem~\ref{thm:coi_implies_omni} quantifies the additional approximation error. Building on the omniprediction framework \citep{gopalan2021omnipredictors,kim2022making,gopalan2024computationally}, we will develop a sample-efficient boosting algorithm for the smoothed collaborative omnimechanism construction.

\section{Methodology}\label{sec:method}

To formally tackle the omnimechanism design objective, we must address the fundamental challenge of our setting: the impressionability of human behavior. Because the human output $y_H$ shifts based on the signal $s$ they are shown, we recast our framework in the language of contextual bandits. Building on outcome indistinguishability \citep{dwork2021outcome,kim2022making}, we employ a boosting algorithm to construct the base predictive distribution $p$, thereby validating the omnimechanism. We then show how to post-process $p$ in a data-free manner to obtain near-optimality guarantees.

\subsection{Policy Learning and State-Action Space}
We treat omnimechanism design as a policy learning problem. We first learn an outcome model from offline human-AI interaction data. At deployment, we use this model to evaluate the expected total losses of candidate actions under a supplied prediction loss and human-query cost, and obtain a collaboration policy by post-processing without retraining. We express this procedure in the language of contextual bandits, with the policies $\pi_h$ induced by mechanisms $h\in\mathcal H$ serving as the comparison class.



For that purpose, we treat $z=(x,y_M)$ as the state context and consider the action space $\cA:=\{0,1\}\times\cS\times\cG$. In addition, we define the (augmented) output space as $\cO:=\cY\times\cY_{\text{true}}$. Now, let us illustrate that given $z$, how each action $a\in \cA$ and outcome $o=(y_{\text{final}}, y_{\text{true}})\in \cO$ are connected. With a slight abuse of notation, given $z$, for an action $a=(r,s,g)\in \cA$ \footnote{Here $r\in\{0,1\}$ denotes a binary decision. A general policy may select different collaboration functions across contexts, whereas each base mechanism fixes one $g\in\mathcal G$.}, it leads to the final augmented outcome $o=(y_{\text{final}},y_{\text{true}})$ in the following way: if $r=0$, it leads to the final decision $y_{\text{final}} = y_M$; if $r=1$, the signal $s$ is presented to the human, and the final decision is given by $y_{\text{final}} = g(y_M,y_H)$. Crucially, we need to keep in mind that $y_H$ is implicitly dependent on $s$ (i.e., the impressionability of human minds). For a pair $\omega=(\tilde{l},c)$ of prediction error loss $\tilde{l} \in \mathcal{L}$ and cost $c \in \mathcal{C}$, we denote the loss value at state $z$ under action $a=(r,s,g)$ and outcome $o$ as:
\begin{equation*}
 \begin{aligned}
     \ell_{a}^{\omega}(z, o) &= (1-r)\tilde{l}(y_M,y_{\text{true}}) +  r\big[\tilde{l}(g(y_M,y_H), y_{\text{true}}) + c(x)\big]= \tilde{l}(o) + rc(x),
 \end{aligned}
 \end{equation*}
 where the last equation follows the definition of $y_{\text{final}}$ and enables us to unify the outcome in automation and deferral processes.



In the following experimental and theoretical analysis, we assume finite action and outcome spaces; the context space need not be finite. For continuous-valued prediction problems, a finite approximation can be obtained by restricting candidate signals and collaboration functions to finite sets and discretizing the continuous coordinates of the outcome pair \((y_{\mathrm{final}},y_{\mathrm{true}})\) into bins. We define a policy as a map $\pi:\mathcal{Z}\to\Delta{\mathcal{A}}$. We also define a base predictive distribution as a map $p:\mathcal{Z}\times \mathcal{A}\to \Delta(\mathcal{O})$, and let $p^*:\mathcal{Z}\times \mathcal{A}\to \Delta(\mathcal{O})$ denote the true data-generating base predictive distribution.  We remark that, as a special case, each $h=(r,\phi,g)\in \mathcal{H}$ can induce a deterministic base policy $\pi_{h}$ by setting $\pi_{h}(z) = (r(z),\phi(z),g(\cdot,\cdot))$ where $g$ is globally applied to all $z$. We identify deterministic policies with point-mass distributions when writing $\pi(a\mid z)$. Then under a pair $\omega=(\tilde{l},c)$, the typical supervised learning objective can be rewritten in the following form: 
$$\min _{h\in \mathcal{H}}\mathbb{E}_{z,y_H,y_{\text{true}}}\ell^\omega(z,y_H,y_{\text{true}};h)=\min _{h\in \mathcal{H}}\mathbb{E}_z\mathbb{E}_{o\sim p^*(\cdot\mid z,\pi_h(z))}\ell_{\pi_{h}(z)}^\omega(z,o).$$ For given $\varepsilon>0$, our $\varepsilon$-omnimechanism design \textbf{\textit{is equivalent to}} searching for a base predictive distribution $p$ that enables a possibly randomized data-free post-processing $\pi^\omega := k^{\omega}(p)$ such that the following near-optimal guarantee holds:
$$\mathbb{E}_{z}\mathbb{E}_{a\sim \pi^{\omega}(\cdot\mid z)}\mathbb{E}_{o\sim p^*(\cdot\mid z,a)}\ell_{a}^\omega(z,o)\leq \min _{h\in \mathcal{H}}\mathbb{E}_z\mathbb{E}_{o\sim p^*(\cdot\mid z,\pi_{h}(z))}\ell_{\pi_{h}(z)}^\omega(z,o) + \varepsilon.$$


To constructively instantiate this scheme with provable efficiency guarantees, we propose the following post-processing rule. We introduce and fix an inverse-temperature parameter $\tau>0$. 
Let $\mathcal{P} = \{ p : \mathcal{Z} \times \mathcal{A} \to \Delta(\mathcal{O}) \}$  denote the space of base predictive distributions, for $p\in \cP$, we write $p(\cdot|z,a)$ the predicted outcome distribution given state $z$ and action $a$. Specifically, for any given loss-cost pair $\omega \in \Omega$, we define the post-processing procedure as the functional mapping $k^\omega: \mathcal{P} \to \Delta(\mathcal{A})^{\mathcal{Z}}$, such that for a given state $z$ and action $a$, its probability mass at $(z,a)$ is:
\begin{equation}\label{eq:postprocessing}
        (k^{{\omega}}(p))_{z,a} :{=} \dfrac{\exp\Big(-\tau\mathbb{E}_{o \sim p(\cdot \mid z,a)}[\ell^\omega_a(z,o)]\Big)}{\sum_{a'\in \mathcal{A}}\exp\Big(-\tau\mathbb{E}_{o \sim p(\cdot \mid z,a')}[\ell^\omega_{a'}(z,o)]\Big)}.
\end{equation} 


Given a state $z$, the probability of taking action $a$ is defined to be  $\pi^{\omega}(a|z):=(k^{{\omega}}(p))_{z,a}$. When $\tau=\infty$, $k^\omega(p)$ is a degenerate distribution at each state $z$ given a fixed tie-breaking rule and the induced policy $\pi^{\omega}(\cdot|z)$ deterministically takes loss-minimizing action, which recovers the post-processing method in our setup as a special case. Taking $\tau<\infty$ enables us to prove time-efficiency result, bypassing the hardness of performative decision-calibration in \cite{kim2022making}.

By relying entirely on the simulated loss via base predictive distribution $p$, this formulation enables data-free adaptation to any prescribed downstream cost-loss pair without accessing fresh samples.
With this explicit post-processing mechanism in place, we can now formally characterize the conditions under which a base predictive distribution satisfies the requirements of our omnimechanism.




\subsection{Our Algorithm}
To algorithmically construct a collaborative omnimechanism, we extend the concept of outcome indistinguishability in \citep{dwork2021outcome,kim2022making} to our human-AI collaboration setting, and introduce a boosting algorithm to construct a base predictive distribution $p$ that achieves the omnimechanism. 

\textbf{Collaborative outcome indistinguishability (COI).} A key property for obtaining a collaborative omnimechanism is what we call collaborative outcome indistinguishability.

\begin{definition}[Collaborative Outcome Indistinguishability]\label{def:coi}
    A base predictive distribution $p$ satisfies $(\Omega,\cH,\varepsilon)$-Collaborative Outcome Indistinguishability (COI) if the following conditions hold:\\
    $(a)$. (Hypothesis-OI) For each loss-cost pair $\omega\in \Omega$, and deterministic base collaboration policy $\pi_{h}$ induced by $h \in \mathcal{H}$, the simulated expected loss under $p$ approximately matches the true expected loss under $p^*$:
    $$\Big| \mathbb{E}_{z} \mathbb{E}_{o \sim p^*(\cdot \mid z, \pi_h(z))} \big[ \ell_{\pi_h(z)}^\omega(z, o) \big] - \mathbb{E}_{z} \mathbb{E}_{o \sim p(\cdot \mid z, \pi_h(z))} \big[ \ell^\omega_{\pi_h(z)}(z, o) \big] \Big| \le \varepsilon$$
    $(b)$. (Decision-OI) For each loss-cost pair $\omega\in \Omega$, given the randomized post-processing policy $\pi^\omega$ defined in Eq.~\eqref{eq:postprocessing} 
    , the simulated expected loss approximately matches the true expected loss:
    $$\Big| \mathbb{E}_{z} \mathbb{E}_{a \sim \pi^{\omega}(\cdot \mid z)} \mathbb{E}_{o \sim p^*(\cdot \mid z, a)} \big[ \ell^\omega_a(z, o) \big] - \mathbb{E}_{z} \mathbb{E}_{a \sim \pi^{\omega}(\cdot \mid z)} \mathbb{E}_{o \sim p(\cdot \mid z, a)} \big[ \ell^\omega_a(z, o) \big] \Big| \le \varepsilon$$
\end{definition}


COI requires approximate agreement of expected losses under the specified policies and loss–cost pairs, without requiring pointwise recovery of the true outcome distribution \(p^*(\cdot\mid z,a)\). Crucially, satisfying the COI condition is sufficient to guarantee that a base predictive distribution $p$ acts as a valid omnimechanism (we defer the formal proof of this sufficiency to Section~\ref{app:theory}). Therefore, our algorithmic goal reduces to finding a base predictive distribution $p$ that satisfies Definition \ref{def:coi}. Here we provide an intuitive overview of these introduced key objects in our framework and their roles.

\paragraph{Intuitive walkthrough.}
Four objects connect offline learning and the COI-condition to deployment:
\begin{enumerate}
    \item \textbf{True outcome distribution $p^*(\cdot\mid z,a)$.}
    This describes the outcome $o=(y_{\mathrm{final}},y_{\mathrm{true}})$
    generated by action $a$ in context $z$, including the effect of
    the displayed signal on human responses.

    \item \textbf{Logging policy $\pi_{\mathrm{log}}(\cdot\mid z)$.}
    This determines which actions are taken during data collection.
    Offline observations follow
    \[
        z\sim\mathcal D_Z,\qquad
        a\sim\pi_{\mathrm{log}}(\cdot\mid z),\qquad
        o\sim p^*(\cdot\mid z,a),
    \]
    where $\mathcal D_Z$ is the fixed context distribution.
    Sufficient logging coverage allows us to estimate the expected
    losses of candidate policies.

    \item \textbf{Learned predictive distribution $p(\cdot\mid z,a)$.}
    COI-Boost constructs a reusable surrogate outcome model,
    updating it when an auditor detects a discrepancy between
    simulated and true expected losses. COI requires agreement
    for the tested policies and loss-cost pairs, without requiring
    pointwise recovery of $p^*$.

    \item \textbf{Deployed policy $\pi^\omega=k^\omega(p)$.}
    Given a supported loss-cost pair $\omega=(\tilde l,c)$,
    Eq.~\eqref{eq:postprocessing} converts the predicted action
    losses into a randomized policy. Changing $\omega$ requires
    only this post-processing step, without retraining $p$ or
    collecting new outcome samples; the context distribution
    and action-conditional outcome law remain fixed.
\end{enumerate}

The near-optimality argument shall use the following three comparisons:
First, Decision-OI transfers the deployed policy's true expected loss to its simulated loss under $p$; Second, the softmax rule makes this simulated loss close to the minimum, and hence no greater than that of any comparator $\pi_h$, $h\in\mathcal H$, up to the smoothing error; Finally, Hypothesis-OI then transfers the comparator's simulated loss back to its true loss. The two COI conditions contribute at most $2\varepsilon$, while smoothing contributes at most $\log|\mathcal A|/\tau$. Below, we present our algorithmic approach to produce a base predictive distribution satisfying the COI condition.

\textbf{COI-boosting algorithm.} We present here our algorithm assuming we have a reliable auditor, which we call a COI-auditor that defined below. To validate this auditor in practice, we assume access to an offline dataset collected via a fixed logging policy, denoted as $\pi_{\text{log}}$. We will defer the details of data collection using $\pi_{\text{log}}$ and construction of such a reliable auditor in Section~\ref{sec:theory}.  

\begin{definition}[COI Auditor]
\label{def:auditor}
Fix $\varepsilon>0$, $\delta\in(0,1)$, and the data
collection policy $\pi_{\mathrm{log}}$. An
$(\Omega,\mathcal H,\varepsilon,\delta,m)$-COI auditor
takes as input a base predictive distribution $p$
and $m$ independent fresh samples
$\{(z_i,a_i,o_i)\}_{i=1}^{m}$ drawn using
$\pi_{\mathrm{log}}$ that are independent of $p$. It outputs either $\perp$ or a certificate $(\pi,\omega,\sigma)$, where $\omega\in\Omega$,
$\sigma\in\{-1,1\}$, and $\pi$ is either a base
policy $\pi_h$ for some $h\in\mathcal H$ or the
post-processed policy $\pi^\omega$, such that: with probability
at least $1-\delta$, the following
guarantees hold:
\begin{enumerate}
    \item If the auditor returns $(\pi,\omega,\sigma)$,
    then
    \[\sigma\mathbb E_z\mathbb E_{a\sim\pi(\cdot\mid z)}
    \left[
    \mathbb E_{o\sim p^\ast(\cdot\mid z,a)}
    [\ell_a^\omega(z,o)]
    -
    \mathbb E_{o\sim p(\cdot\mid z,a)}
    [\ell_a^\omega(z,o)]
    \right]
    >\frac{\varepsilon}{2}.\]
    \item If the auditor returns $\perp$, then $p$
    satisfies $(\Omega,\mathcal H,\varepsilon)$-COI.
\end{enumerate}
In particular, if $p$ does not satisfy
$(\Omega,\mathcal H,\varepsilon)$-COI, the auditor
must return a certificate on this event. However, when $p$ satisfies $(\Omega,\mathcal H,\varepsilon)$-COI but not $(\Omega,\mathcal H,\varepsilon/2)$-COI, either valid output is permitted.
\end{definition}



\begin{algorithm}[h]
\caption{COI-Boosting Algorithm}\label{alg:coiboosting}
    \begin{algorithmic}[1]
        \State Given $\mathcal{H}$,  $\mathcal{L}$, $\mathcal{C}$, $\varepsilon$, $\pi_{\text{log}}$, $\delta$, and a $(\Omega,\mathcal{H},\varepsilon,\delta,m)$-COI auditor.
         Set learning rate $\eta$, maximal iteration round $T_{\max}$ and the initial base predictive distribution $p^{(0)}(o\mid z,a) = 1/|\mathcal{O}|$ for all $(z,a,o)$.
        \For{$0\leq t< T_{\text{max}}$}
            \State Draw $m$-samples $\{(z_i,a_i,o_i)\}_{i=1}^m$ and call the $( \Omega, \mathcal{H},\varepsilon, \delta, m)$-COI auditor with $p^{(t)}$.
            \If{the auditor outputs $\perp$} 
            Terminate.
            \Else 
            \State 
            The auditor outputs $(\pi^{(t)}, \omega^{(t)},\sigma)$. Set $g^{(t)}(z,a,o) = \sigma  \frac{\pi^{(t)}(a\mid z)}{\pi_{\text{log}}(a\mid z)} \ell^{\omega^{(t)}}_{a}(z,o)$;
            \State Update the base predictive distribution with learning rate $\eta$, $$p^{(t+1)}(o\mid z,a) = \dfrac{
p^{(t)}(o\mid z,a)\exp(\eta g^{(t)}(z,a,o))}{ \sum_{o'\in\mathcal O}
p^{(t)}(o'\mid z,a)\exp(\eta g^{(t)}(z,a,o'))
}.$$
            \EndIf
        \EndFor
        \State \Return{$p^{(t)}$} if it terminates or reaches the maximal iteration round $t=T_{\text{max}}$.
    \end{algorithmic}
\end{algorithm}







Table~\ref{tab:notation} summarizes the main notation.
Throughout, $\Delta(S)$ denotes the set of probability
distributions over a set $S$.

\begin{table}[t]
\centering
\caption{Main notation for collaborative omnimechanism design.}
\label{tab:notation}
\small
\renewcommand{\arraystretch}{1.08}
\begin{tabular}{@{}p{0.25\linewidth}p{0.75\linewidth}@{}}
\toprule
\textbf{Notation} & \textbf{Description} \\
\midrule

$x\in\mathcal X,\ z=(x,y_M)\in\mathcal Z$
& Input and observed context;
$\mathcal Z=\mathcal X\times\mathcal Y$. \\

$M;\ y_M,y_H,y_{\mathrm{final}}\in\mathcal Y$
& Black-box AI model; AI, human, and final predictions. \\

$y_{\mathrm{true}}\in\mathcal Y_{\mathrm{true}}$
& Ground-truth label and its space. \\

$r\in\mathcal R,\ \phi\in\Phi,\ g\in\mathcal G$
& Deferral, signal-presentation, and collaboration rules;
$s=\phi(z)\in\mathcal S$ is the displayed signal. \\

$a=(r,s,g)\in\mathcal A$
& Collaboration action;
$\mathcal A=\{0,1\}\times\mathcal S\times\mathcal G$.
Here $r$ is a binary decision. \\

$o\in\mathcal O$
& Outcome $o=(y_{\mathrm{final}},y_{\mathrm{true}})$;
$\mathcal O=\mathcal Y\times\mathcal Y_{\mathrm{true}}$. \\

$h\in\mathcal H,\ \pi_h$
& Base mechanism $h=(r,\phi,g)$ in
$\mathcal H=\mathcal R\times\Phi\times\mathcal G$,
inducing $\pi_h(z)=(r(z),\phi(z),g)$. \\

$\pi:\mathcal Z\to\Delta(\mathcal A)$
& Randomized collaboration policy. \\

$p^*,p\in\mathcal P$
& True and learned outcome models;
$\mathcal P=\{p:\mathcal Z\times\mathcal A\to\Delta(\mathcal O)\}$.
Their conditional distributions are written
$p^*(\cdot\mid z,a)$ and $p(\cdot\mid z,a)$. \\

$\omega=(\tilde l,c)\in\Omega$
& Prediction loss and query-cost pair;
$\Omega=\mathcal L\times\mathcal C$. \\

$\ell_a^\omega(z,o)$
& Total loss:
$\tilde l(y_{\mathrm{final}},y_{\mathrm{true}})+rc(x)$. \\

$k^\omega,\ \pi^\omega=k^\omega(p)$
& Loss-specific post-processing map and its induced policy. \\

$\mathcal D_Z,\ \pi_{\mathrm{log}}$
& Fixed context distribution and historical logging policy. \\

$\tau>0$
& Softmax inverse temperature;
larger values favor lower predicted losses more strongly. \\

$\gamma$
& Logging coverage bound:
$\pi_{\mathrm{log}}(a\mid z)\geq\gamma$. \\

$B_c,\ B$
& Query-cost magnitude bound and total-loss bound,
respectively. \\

$d_{\mathcal H},d_{\mathcal R},d_\Phi$
& Natarajan dimensions of the induced base-policy class,
deferral-rule class, and signal-rule class, respectively. \\

$m,\ \varepsilon,\ \delta$
& Offline sample size, accuracy tolerance,
and failure-probability bound. \\

$p^{(t)},\ \eta,\ T_{\max}$
& Predictor at boosting iteration $t$, learning rate,
and maximum iteration index. \\

\bottomrule
\end{tabular}
\end{table}

{\color{red} 

}

\section{Theoretical Guarantees }\label{sec:theory}
In this section, we establish the theoretical foundations of our collaborative omnimechanism framework. We proceed in two steps. First, we prove that satisfying the Collaborative Outcome Indistinguishability (COI) condition is sufficient to guarantee a valid omnimechanism. Second, we establish sample-complexity bounds for COI auditing and evaluate practical implementations of our framework empirically in Section~\ref{sec:exp}.



We begin by formally establishing the sufficiency of COI. Theorem~\ref{thm:coi_implies_omni} shows that if our boosting algorithm yields a base predictive distribution that is indistinguishable from the true environment under our specified test classes, the resulting post-processed policy will be near-optimal for any downstream loss and cost, and thus guarantees that the base predictive distribution is a valid omnimechanism.
\begin{theorem}[COI Implies Collaborative Omnimechanism.]
\label{thm:coi_implies_omni}
If a base predictive distribution ${p}:\mathcal{Z}\times\mathcal{A}\to \Delta\mathcal{O}$ satisfies $(\Omega,\mathcal{H},\varepsilon)$-COI (Definition~\ref{def:coi}), then it is an $(\Omega,\mathcal{H},2\varepsilon+\frac{\log|\mathcal{A}|}{\tau})$-collaborative omnimechanism. Specifically, for any loss-cost pair $\omega\in \Omega$,  we have:
\begin{equation*}
    \mathbb{E}_{z} \mathbb{E}_{a \sim \pi^\omega(\cdot \mid z)} \mathbb{E}_{o \sim p^*(\cdot \mid z, a)} \big[ \ell^\omega_a(z, o) \big] \le \min_{h \in \mathcal{H}} \mathbb{E}_{z} \mathbb{E}_{o \sim p^*(\cdot \mid z, \pi_h(z))} \big[ \ell^\omega_{\pi_h(z)}(z, o) \big] + 2\varepsilon + \frac{\log|\mathcal{A}|}{\tau}.
\end{equation*}
\end{theorem}



\vspace{-0.5em}
Theorem~\ref{thm:coi_implies_omni} confirms that collaborative omnimechanism can be achieved via a COI base predictive distribution. When $\tau$ is large, the near-optimality gap becomes negligible. We shall show that the convergence guarantee of  algorithm (Algorithm~\ref{alg:coiboosting})  holds for both $\tau<\infty$ and $\tau=\infty$ and further validate empirically that our algorithm is efficient even if we take $\tau=\infty$. On the other hand, the convergence of the COI-Boosting algorithm (Algorithm~\ref{alg:coiboosting}) relies on the existence of a reliable and efficient COI-auditor (Definition~\ref{def:auditor}). To practically operationalize an $(\Omega, \mathcal{H},\varepsilon, \delta, m)$-COI auditor, we assume access to a historical dataset of human-AI collaborations. Specifically, given a fixed data collection policy $\pi_{\text{log}}:\mathcal{Z}\to \Delta\mathcal{A}$, the samples for the auditor input are drawn via the following process: 
$$
z\sim \mathcal{D}_{\mathcal{Z}},\ a\sim \pi_{\text{log}}(\cdot \mid z),\ o\sim p^*(\cdot \mid z,a),
$$
where $\mathcal{D}_{\mathcal{Z}}$ denotes the marginal distribution over state $z\in\mathcal{Z}$. In Theorem~\ref{thm:upper}, we show that under standard data coverage and finite complexity assumptions, we can construct such a reliable, sample-efficient auditor. We defer the formal statements of  assumptions on data coverage (Assumption~\ref{as:log}) and finiteness of complexity and boundedness for $\mathcal{L},\mathcal{C},\mathcal{H}$ (Assumption~\ref{as:complexity}) to the Appendix~\ref{app:theory}.


\begin{theorem}[Sample Efficiency of Auditor, Informal]
\label{thm:upper}
Under Assumptions~\ref{as:log} and~\ref{as:complexity},
fix $\tau>0$ and $\varepsilon,\delta\in(0,1)$.
For any base predictive distribution $p$ independent
of the audit batch, there exists an
$(\Omega,\mathcal H,\varepsilon,\delta,m)$-COI auditor
using
\[
m = \mathcal O\left(
\frac{(1+B_c)^2}{\gamma^2\varepsilon^2}
\left(
(d_{\mathcal R}+d_\Phi+\log|\mathcal G|)\log|\mathcal A|
+(1+\tau)^2|\mathcal O|
+\log|\mathcal C|
+\log(1/\delta)
\right)
\right)
\]
independent offline samples collected under
$\pi_{\mathrm{log}}$.
With probability at least $1-\delta$, the auditor
outputs either a valid violation certificate
$(\pi,\omega,\sigma)$ or $\perp$, where returning
$\perp$ certifies that $p$ satisfies
$(\Omega,\mathcal H,\varepsilon)$-COI. Here $\gamma>0$ is the coverage constant for logging policy, $B_c$ is a uniform bound of the cost functions, and
$d_{\mathcal R}$ and $d_\Phi$ are the Natarajan
dimensions of $\mathcal R$ and $\Phi$ respectively. 
\end{theorem}

This bound reveals that the sample complexity of the auditor scales  linearly with the structural complexity of the collaboration policy classes (i.e., $d_{\mathcal{R}}$, $d_{\Phi}$, and $\log|\mathcal{G}|$), which is of the same order as in previous theoretical learning to defer work \cite{mozannar2020consistent}. Coupled with the convergence guarantees of the boosting procedure (Theorem~\ref{thm:convergence} in Appendix~\ref{app:theory}), this ensures the sample-efficient construction of a collaborative omnimechanism. 

We remark that, beyond introducing COI as a framework for human-AI collaboration with impressionable decision-makers, we establish new auditing guarantees that distinguish our setting from prior omniprediction and performative prediction frameworks. Inspired by \cite{gopalan2024computationally}, we establish in Appendix~\ref{app:timeeff} efficient witness-based Decision-OI auditing through a reduction to weighted calibration. This yields an auditor with runtime
$\operatorname{poly}(|\mathcal{A}\times\mathcal{O}|^{O(C(B,\tau)\log^{2}(1/\varepsilon))})$ under uniform logging policy with fixed scalar cost and finite inverse temperature. This runtime is polynomial in the action and outcome space cardinalities for fixed accuracy, and quasipolynomial in inverse accuracy when those cardinalities are fixed, which avoids the exponential dependence on loss-space dimension incurred by direct discretization, achieving polynomial dependence on the action and outcome cardinalities. It thereby addresses, in our smoothed setting, the computational bottleneck highlighted by \citet{kim2022making} for performative decision-calibration. Complementing our per-call auditing upper bound, we establish a sample-complexity lower bound for offline omnimechanism learning:
$m=\Omega\left(
\frac{d_{\mathcal H}+\log(1/\delta)}
{\gamma\varepsilon^2}
\right),$
this lower bound establishes unavoidable dependence on policy-class complexity, inverse-squared accuracy, and limited logging coverage. Together, these results show that COI is not only a conceptual extension of outcome indistinguishability to human-AI collaboration with impressionable decision-makers, but also admits efficient and nearly tight auditing guarantees.





\section{Extension}\label{sec:ext}
We further extend our framework to settings with multiple human experts and feature-based methods, highlighting its flexibility and applicability to more complex human-AI collaboration scenarios. Due to space limitations, we defer the technical details and analysis to the appendix.

\textbf{Incorporating opinions from multiple humans.} In human--AI collaboration, a task may be deferred to one of several human experts with different specialties and query costs~\citep{mao2023two,verma2023learning,ramgolam2025learning}. We consider routing each deferred task to a single expert, whose identity is included in the action. Appendix~\ref{app:ext} explains how our sample-complexity
analysis extends to this enlarged action space, with
the resulting guarantee depending on the logging coverage
and the complexity of the expanded collaboration policy class.



\textbf{Efficient auditor with feature-based methods.}
In previous human-AI collaboration frameworks, given a state $z=(x,y_M)$, the action $a=(r,s,g)\in \mathcal{A}$, consisting of the deferral decision, signal and collaboration function, is often determined based solely on the AI output $y_M$ \citep{vodrahalli2022uncalibrated,bondi2022role,alur2024human,casacuberta2025selective}. This can be viewed as a special case of feature-based methods, where we introduce a feature map that projects a state $z$ into a low-dimensional feature vector. We show that under standard Lipschitz continuity assumptions, our framework extends to this setting, preserving the sample-efficiency guarantees for the COI-auditor. We defer the theoretical details to Appendix~\ref{app:ext}.







\section{Experiments}\label{sec:exp}
We evaluate our framework in three settings.  Experiment~1 is our primary
experiment, in which we learn omnimechanisms using both COI-Boosting and a
neural network and evaluate their optimally post-processed policies and COI
violations.  Experiment~2 uses empirical outcome distributions from 
radiologist--AI interaction data, without relying on a behavioral simulator.
Experiment~3 extends the evaluation to continuous house-price prediction.
Together, these experiments support the central claim that a learned
omnimechanism can induce effective input-dependent policies for different
downstream objectives. Implementation details and additional results
are provided in the appendix.



\subsection{Experiment 1: Impressionable Human Decisions}\label{sec:exp-vodrahalli}

\textbf{Dataset.} We conduct experiments on the Art and Sarcasm tasks from \textbf{the real human-AI interaction dataset} of \citep{vodrahalli2022uncalibrated}, which contains over 30,000 interactions across several binary classification tasks. Each example records a real participant's initial response, an AI signal, and the participant's final response after observing the AI advice. Following \citep{vodrahalli2022uncalibrated}, we fit their two-stage behavioral model as a black-box simulator; as reported in \citep{vodrahalli2022uncalibrated}, the model was shown to align well with aggregate real human behavior on these tasks. 
The base predictive distribution is then trained and evaluated only on samples generated by the fitted simulator, without access to its parameters.
This setup does not require white-box access to the human-behavior model or the convexity and sensitivity assumptions used in prior work \citep{perdomo2020performative,lin2023plug,li2025statistical}.



\textbf{Implementation.} We instantiate the omnimechanism in a setting where the downstream cost varies. For each example, the system observes a state $z$\footnote{This is a task-specific representation of the general state $z=(x,y_M)$ used in the theoretical formulation.} that summarizes the pre-interaction information: the human's initial confidence, the AI's confidence, and their initial agreement indicator. The system then chooses an action $a$ consisting of a deferral decision, a signal, and a collaboration function. The action either directly outputs the AI prediction, or routes the instance to the human with a transformed AI-advice signal \citep{vodrahalli2022uncalibrated} and combines the human and AI predictions through a linear collaboration function. We treat the initial human judgment as given; cost $c$ and human
involvement refer only to additional consultation.


For each state-action pair $(z,a)$, we use a trained base predictive distribution $p(\cdot\mid z,a)$ to estimate the simulated expected loss. Given a pair $\omega = (\tilde{l},c)$ of a loss function $\tilde{l}$ and a constant human-involvement cost $c$\footnote{By a slight abuse of notation, we identify a constant cost function $c(z)$ with its value at each state.}, the optimal post-processing policy chooses $\pi^\omega(z) \in \arg\min_a \{ \mathbb{E}_{{p}(\cdot \mid z,a)} [\ell_a^\omega(z,o)]\}$\footnote{We evaluate the deterministic post-processing policy, which is equivalent to taking $\tau=\infty$ in Eq.~\ref{eq:postprocessing}. We provide efficiency guarantees for deterministic policies in Appendix~\ref{app:theory}.}. 

We implement our omnimechanism with the boosting algorithm (Algorithm~\ref{alg:coiboosting}) by discretizing $z$ and maintaining an outcome distribution for each discrete state-action pair. To extend this approach to continuous spaces, we further develop a neural base predictive distribution utilizing a three-layer neural network that preserves the continuous nature of $z$. This provides a function-approximation implementation of the same omnimechanism. 
We present implementation details in Appendix~\ref{app:experimental_details}.

\begin{table}[!t]
\centering
\caption{Policy performance under human-involvement costs $c \in \{0,0.1,0.2\}$. We report accuracy (Acc.), utility ($\mathrm{Util.}=\mathrm{Acc.}-c\cdot\mathrm{Hum.}$), and human-involvement
rate (Hum.).
The best and second-best utilities in each setting are shown in \textbf{bold} and \underline{underlined}, respectively.}
\label{tab:hac-results}
\vspace{0.5em}
\renewcommand{\arraystretch}{0.95}
\setlength{\tabcolsep}{5pt}
\resizebox{0.9\linewidth}{!}{
\begin{tabular*}{\textwidth}{@{\extracolsep{\fill}}llccc ccc}
\toprule
& & \multicolumn{3}{c}{\textbf{Art}} & \multicolumn{3}{c}{\textbf{Sarcasm}} \\
\cmidrule(lr){3-5} \cmidrule(lr){6-8}
$c$ & \textbf{Method}
& \textbf{Acc.} & \textbf{Util.} & \textbf{Hum.}
& \textbf{Acc.} & \textbf{Util.} & \textbf{Hum.} \\
\midrule
\multirow{5}{*}{0.0}
& No Action                 & 0.768 & 0.768 & 1.000 & 0.722 & 0.722 & 1.000 \\
& Manipulation             & 0.776 & 0.776 & 1.000 & 0.733 & 0.733 & 1.000 \\
& Global Best              & 0.777 & \underline{0.777} & 0.000 & 0.792 & \underline{0.792} & 0.227 \\
& Ours (Boosting)          & 0.778 & \textbf{0.778} & 0.178 & 0.785 & 0.785 & 0.378 \\
& Ours (NN)                & 0.777 & \underline{0.777} & 0.043 & 0.806 & \textbf{0.806} & 0.099 \\
\midrule
\multirow{5}{*}{0.1}
& No Action                 & 0.768 & 0.668 & 1.000 & 0.722 & 0.622 & 1.000 \\
& Manipulation             & 0.776 & \underline{0.676} & 1.000 & 0.733 & 0.633 & 1.000 \\
& Global Best              & 0.777 & \textbf{0.777} & 0.000 & 0.776 & 0.776 & 0.000 \\
& Ours (Boosting)          & 0.777 & \textbf{0.777} & 0.000 & 0.793 & \underline{0.779} & 0.141 \\
& Ours (NN)                & 0.777 & \textbf{0.777} & 0.000 & 0.796 & \textbf{0.792} & 0.047 \\
\midrule
\multirow{5}{*}{0.2}
& No Action                 & 0.768 & 0.568 & 1.000 & 0.722 & 0.522 & 1.000 \\
& Manipulation             & 0.776 & \underline{0.576} & 1.000 & 0.733 & 0.533 & 1.000 \\
& Global Best              & 0.777 & \textbf{0.777} & 0.000 & 0.776 & \underline{0.776} & 0.000 \\
& Ours (Boosting)          & 0.777 & \textbf{0.777} & 0.000 & 0.776 & \underline{0.776} & 0.000 \\
& Ours (NN)                & 0.777 & \textbf{0.777} & 0.000 & 0.780 & \textbf{0.777} & 0.016 \\
\bottomrule
\end{tabular*}}
\vspace{-1.5em}
\end{table}



\textbf{Evaluation and analysis.} Table~\ref{tab:hac-results} reports accuracy, utility, and human involvement rate for each method under different human cost $c$. We compare our post-processing policies with three global baselines: \textbf{No Action}, which always defers to the human with the original AI advice; \textbf{Manipulation}, which always routes to the human with a globally optimized signaling function $\phi$ \citep{vodrahalli2022uncalibrated}; and \textbf{Global Best}, which selects the policy in $\cH$ with the
highest overall utility for each cost $c$ and holds that policy fixed across inputs. Here, $\cH$ extends \textbf{Manipulation} with typical deferral functions and human-AI collaboration functions in \citep{bondi2022role,alur2024human}. More details are presented in Appendix~\ref{app:baselines_metrics}.



\emph{\textbf{The post-processed policies are competitive with
Global Best across human-cost regimes.}} On Art, both implementations
recover AI-only behavior at nonzero costs and match Global Best's
utility. On Sarcasm, the neural implementation achieves higher utility
at all reported costs, while Boosting also improves upon Global Best
at $c=0.1$. These results illustrate how a learned omnimechanism can
support effective collaboration policies across downstream costs
through post-processing.


As cost $c$ increases, the post-processing policies $\pi^\omega$ reduce or eliminate deferral, reflecting the trade-off between accuracy and growing human cost. We highlight that this adaptation is obtained through the post-processing of our omnimechanism, rather than by retraining a separate model for each cost parameter.
Beyond the original setting, where AI advice accuracy is around $80\%$, we further evaluate shifted AI-advice settings with average AI accuracy reduced to approximately $70\%$ (Table~\ref{tab:shift-ai-70}) and $60\%$ (Table~\ref{tab:shift-ai-60}) in Appendix~\ref{app:varying_ai_accuracy}. We retrain the omnimechanisms under each shifted advice distribution;
within each setting, the resulting predictors are reused across costs. The omnimechanism-based policies remain competitive with the strongest baselines across these settings, further supporting the effectiveness of our approach under different AI reliability levels.

\begin{figure}[!t]
  \centering
  \begin{subfigure}[t]{0.49\textwidth}
    \centering
    \includegraphics[width=\textwidth]{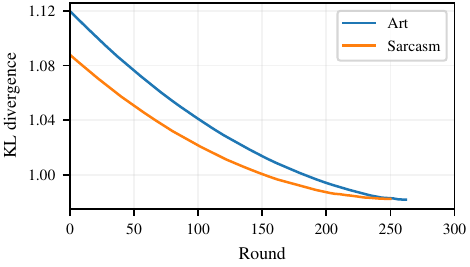}
     \vspace{-1.5em}
    \caption{KL divergence from the simulator target to the
    learned predictive distribution.}
    \label{fig:kl}
  \end{subfigure}
  \hfill
  \begin{subfigure}[t]{0.49\textwidth}
    \centering
    \includegraphics[width=\textwidth]{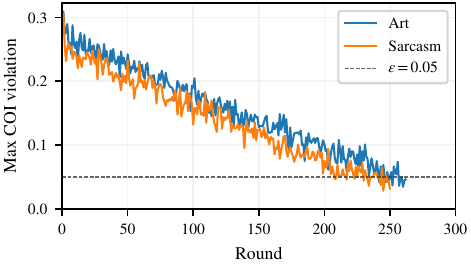}
     \vspace{-1.5em}
    \caption{Maximum empirical COI violation. The dashed line marks $\varepsilon=0.05$.}
    \label{fig:audit-gap}
  \end{subfigure}  
  \vspace{-0.5em}
  \caption{Training dynamics of the boosting-based omnimechanism over boosting rounds.}
  \label{fig:boosting-comparison}
  \vspace{-0.5em}
\end{figure}



\textbf{Convergence of boosting-based omnimechanism.} We find that \emph{\textbf{the boosting-based omnimechanism exhibits stable finite-sample training dynamics consistent with our convergence analysis in Theorem~\ref{thm:convergence}.}}
As shown in Figure~\ref{fig:boosting-comparison}, the KL divergence
from the smoothed simulator target to the learned predictive
distribution decreases over boosting rounds. The maximum empirical
COI violation also approaches the tolerance $\varepsilon=0.05$.
These diagnostics show improved distributional alignment and
agreement in expected loss over the audited policies and costs.

\subsection{Experiment 2: Radiologist--AI Collaboration}
\label{sec:exp-radiology}

\textbf{Data and setting.}
We use the Design~1 subset of Collab-CXR, comprising probabilistic assessments by 159 professional radiologists on 324 chest X-rays under randomized information conditions
\citep{moehring2025dataset,agarwal2023combining}. 
We focus on Airspace Opacity and Cardiomediastinal Abnormality,
the two pre-registered diagnostic categories in the original study. The action set consists of direct AI prediction (\textsc{AI}), unaided radiologist assessment
(\textsc{H}), and AI-assisted radiologist assessment (\textsc{AI+H}), matching the main collaboration modes considered in the original study.



\textbf{Implementation and evaluation.}
Following Experiment~1, we learn an action-conditional four-state predictive
distribution and obtain a policy by deterministic expected-loss-minimizing
post-processing. The training targets are constructed directly from the
observed AI and radiologist probabilities. The model receives a MedSigLIP-448
X-ray embedding \citep{sellergren2025medgemma}, the AI probability, the
pathology, and the candidate action.  Following
\citet{agarwal2023combining}, we evaluate diagnostic probabilities by their
average deviation from the diagnostic standard.  For a reported diagnostic
probability $\rho \in [0,1]$ and binary diagnostic-standard label $y \in \{0,1\}$,
the utility of action $a$ at human cost $c$ is
$u_c(\rho,y,a)=1-|\rho-y|-c\mathbf{1}\{a\ne\textsc{AI}\}$.  We use a patient-level
split with 259 training and 65 test X-rays, with model selection performed
using only the training set.  Post-processing selects the action with the
highest estimated expected utility.  For comparison, Global Best selects on
the training set a single action for each $c$ and applies it to all test inputs across both pathologies.  Further details are given in Appendix~\ref{app:radiology_experiment}.

\begin{table}[!t]
\centering
\caption{Collab-CXR test performance by pathology.
Score is $1-|\rho-y|$ and $\mathrm{Util.}=\mathrm{Score}-c\cdot\mathrm{Hum.}$.
Global Best is selected overall on training data, not separately by pathology.}
\label{tab:radiology-main}
\vspace{0.4em}
\renewcommand{\arraystretch}{0.95}
\setlength{\tabcolsep}{4pt}
\resizebox{0.85\linewidth}{!}{
\begin{tabular}{cl ccc ccc}
\toprule
& & \multicolumn{3}{c}{\textbf{Ours (NN)}} &
\multicolumn{2}{c}{\textbf{Global Best}} & \textbf{$\Delta$ Util.} \\
\cmidrule(lr){3-5}\cmidrule(lr){6-7}
$c$ & Pathology & Score & Hum. & Util. & Action & Util. &
Ours $-$ Global \\
\midrule
\multirow{2}{*}{0.00} & Airspace Opacity          & 0.795 & 0.938 & 0.795 & \textsc{AI+H} & 0.771 & $+0.024$ \\
                      & Cardiomediastinal Abnorm  & 0.768 & 0.785 & 0.768 & \textsc{AI+H} & 0.743 & $+0.025$ \\
\midrule
\multirow{2}{*}{0.02} & Airspace Opacity          & 0.774 & 0.538 & 0.763 & \textsc{AI+H} & 0.751 & $+0.012$ \\
                      & Cardiomediastinal Abnorm  & 0.767 & 0.015 & 0.767 & \textsc{AI+H} & 0.723 & $+0.044$ \\
\midrule
\multirow{2}{*}{0.05} & Airspace Opacity          & 0.720 & 0.077 & 0.716 & \textsc{AI}   & 0.697 & $+0.019$ \\
                      & Cardiomediastinal Abnorm  & 0.768 & 0.000 & 0.768 & \textsc{AI}   & 0.768 & $0.000$ \\
\midrule
\multirow{2}{*}{0.10} & Airspace Opacity          & 0.697 & 0.000 & 0.697 & \textsc{AI}   & 0.697 & $0.000$ \\
                      & Cardiomediastinal Abnorm  & 0.768 & 0.000 & 0.768 & \textsc{AI}   & 0.768 & $0.000$ \\
\midrule
\multirow{2}{*}{0.20} & Airspace Opacity          & 0.697 & 0.000 & 0.697 & \textsc{AI}   & 0.697 & $0.000$ \\
                      & Cardiomediastinal Abnorm  & 0.768 & 0.000 & 0.768 & \textsc{AI}   & 0.768 & $0.000$ \\
\bottomrule
\end{tabular}}
\vspace{-0.5em}
\end{table}



\textbf{Results.}
Table~\ref{tab:radiology-main} reports diagnostic score, human-involvement
rate, and utility separately for each pathology.
\emph{\textbf{The post-processed policy improves upon Global Best at low and
moderate human costs and recovers the AI-only policy as the cost increases.}}
When human input is free, it improves utility over \textbf{Global Best} for
both Airspace Opacity ($+0.024$) and Cardiomediastinal Abnormality ($+0.025$);
the overall gain is $0.025$.  At $c=0.02$, it adapts differently across
pathologies, retaining substantially more human involvement for Airspace
Opacity while improving utility for both.  At $c=0.05$, the remaining gain is
concentrated in Airspace Opacity.  For $c\in\{0.10,0.20\}$, the same trained
predictor selects \textsc{AI} for every test input and exactly matches
\textbf{Global Best}.  Averaged across pathologies, the utility improvements
are $0.028$ at $c=0.02$ and $0.010$ at $c=0.05$.


\subsection{Experiment 3: AI-Assisted House-Price Prediction}
\label{sec:exp-ames}

\textbf{Data and setting.}
To complement the classification settings in Experiments~1 and~2, we use a simulated experiment based on the AMES AI-assisted continuous-valued house-price prediction 
dataset introduced by \citet{guo2025value}. The released data contain 5,244
paired initial and assisted judgments from 437 participants on 240 houses.
The action set has nine total actions, consisting of directly using the initial human estimate
($\textsc{H}_0$), either of two AI predictions ($\textsc{AI}_1$ and
$\textsc{AI}_2$), or requesting a human revision under one of six assistance
conditions proposed by \citep{guo2025value}: either AI model paired with no explanation, SHAP, or ILIV+SHAP. The decision is made after observing the initial human estimate, and the human cost applies only to an additional revision.

\textbf{Implementation and evaluation.}
Following the NN method in Experiments~1 and~2, we learn an action-conditional
predictive distribution and obtain a policy by deterministic
expected-loss-minimizing post-processing. We represent outcomes by 13 bins of signed relative prediction error. Training targets are one-hot for the
three observed direct outputs and empirical bin distributions for simulated
human revisions. The behavioral simulator is fitted to the log-relative
human update and generates responses by sampling prediction residuals.
The omnimechanism receives six house features, the initial human estimate,
both AI predictions, and the candidate action; the sale price is used only
to construct training targets and evaluate losses.
Following \cite{guo2025value}, for a final estimate $v$ and sale price $y>0$, we use absolute percentage
error (APE), $d_{\mathrm{APE}}(v,y)=|(v-y)/y|$, with total loss
$d_{\mathrm{APE}}(v,y)+c\mathbf{1}\{a\in\mathcal A_1\}$, where
$\mathcal A_1$ is the set of six revision actions. 
We remark that the policy uses bin representatives to estimate losses when selecting actions; evaluation uses the continuous prediction errors. Appendix~\ref{app:ames_experiment} provides the discretization, logged-response evaluation, and sensitivity analyses.



\begin{table}[!t]
\centering
\caption{AMES simulator test performance on 1,044 contexts from 87 held-out
participants (seed 42). Error is continuous APE, Rev. is the fraction
of contexts requesting an additional human revision, and
$\mathrm{Total}=\mathrm{Error}+c\cdot\mathrm{Rev.}$.
Positive $\Delta$ Loss indicates lower total loss for Ours.
Global Best is selected separately for each cost on training data.}
\label{tab:ames-main}
\vspace{0.4em}
\renewcommand{\arraystretch}{0.95}
\setlength{\tabcolsep}{4pt}
{
\begin{tabular}{c ccc cc c}
\toprule
& \multicolumn{3}{c}{\textbf{Ours (NN)}} &
\multicolumn{2}{c}{\textbf{Global Best}} & \textbf{$\Delta$ Loss} \\
\cmidrule(lr){2-4}\cmidrule(lr){5-6}
$c$ & Error & Rev. & Total & Action & Total & Global $-$ Ours \\
\midrule
0.000 & 0.1053 & 0.0278 & 0.1053 & $\textsc{AI}_2$ & 0.1335 & $+0.0281$ \\
0.010 & 0.1052 & 0.0163 & 0.1054 & $\textsc{AI}_2$ & 0.1335 & $+0.0281$ \\
0.025 & 0.1054 & 0.0115 & 0.1057 & $\textsc{AI}_2$ & 0.1335 & $+0.0278$ \\
0.050 & 0.1054 & 0.0077 & 0.1057 & $\textsc{AI}_2$ & 0.1335 & $+0.0277$ \\
0.100 & 0.1056 & 0.0000 & 0.1056 & $\textsc{AI}_2$ & 0.1335 & $+0.0279$ \\
\bottomrule
\end{tabular}}
\vspace{-0.5em}
\end{table}

\textbf{Results.}
Table~\ref{tab:ames-main} reports continuous APE, additional human-revision rate, and total loss. \emph{\textbf{The post-processed policy improves upon Global Best throughout
the reported cost range and reduces additional human involvement as the  cost increases.}} At zero cost, the NN policy reduces APE from 0.1335 to 0.1053, while requesting revisions on 2.78\% of contexts. As cost increases, additional human involvement declines, reaching zero at $c=0.10$. The policy retains gains at this cost by choosing among the initial human estimate and the two AI predictions, whereas Global Best uses $\textsc{AI}_2$ throughout. These results illustrate our framework's ability to adapt to the human-involvement cost without retraining.



\section{Conclusion}

In this work, we present a novel collaborative omnimechanism framework that accounts for signal-dependent human responses and adapts collaboration policies to different deployment-time prediction losses and human-query costs.
 Unlike prior approaches that assume fixed human behavior or optimize a single objective, our method learns a unified predictive model that can be post-processed to adapt to different query costs and evaluation criteria without retraining. The framework may improve decision quality and reduce unnecessary human workload in high-stakes domains, but it also risks manipulating human judgments or amplifying overreliance on AI, requiring transparency, auditing, and safeguards against misuse. Our evaluation uses existing interaction datasets, and two experiments rely partly on fitted behavioral simulators. The results therefore depend on simulator fidelity, and selected human-revision interventions can have sparse logged support. The finite intervention menus also leave richer forms of presentation and interaction unexplored.

\section{Acknowledgements}
Vahid Tarokh was supported in part by the NSF under the National AI Institute for Edge Computing Leveraging Next Generation Wireless Networks, Grant CNS-2625580.

\clearpage
\bibliography{refs}
\bibliographystyle{plainnat}


\clearpage
\appendix

\section{Additional Related Work}
\textbf{Human-AI Collaboration.} 
Learning to defer (L2D) framework \cite{geifman2017selective,mozannar2020consistent} jointly optimize predictors and rejectors. Subsequent works have extended this to various settings, including  capacity constraints \cite{mozannar2020consistent,mao2023two}, and routing to multiple experts \cite{mao2023two,verma2023learning,ramgolam2025learning}. \cite{guo2024decision,alur2023auditing,alur2024human} studied the complementary information in human-AI collaboration and efficient aggregation method to improve human-AI team performance. 

\textbf{Impressionability of Human Behavior.}
A growing body of empirical research \cite{bondi2022role,vodrahalli2022uncalibrated,bo2025rely} demonstrates that human decision-makers are highly impressionable when interacting with AI systems. Performative  prediction \cite{perdomo2020performative,lin2023plug,li2025statistical} studied theoretically the situation  where the deployment of a model shifts the underlying data distribution under convexity and smoothness assumptions.

\textbf{Omniprediction and Outcome Indistinguishability.}
To handle the deployment-time variability of loss and cost functions without retraining, \cite{gopalan2021omnipredictors} initiated the theory of omniprediction. \cite{dwork2021outcome,dwork2022beyond} provide efficiency guarantees to achieve omniprediction via outcome indistinguishability (OI). Recently, omniprediction has been extended to selective classification \cite{casacuberta2025selective} and performative prediction \cite{kim2022making}.

\section{Theoretical Results and Proofs}\label{app:theory}

In this section, we formally state all the assumptions and theoretical results in Section~\ref{sec:theory}, and provide detailed proofs.




\subsection{Assumptions and Notations}
We make the following assumption on data collection policy $\pi_{\text{log}}$ for COI-auditor (Definition~\ref{def:auditor}) to guarantee that the offline dataset explores the action space and distribution shift induced by impressionable human behavior. Recall that we denote $\mathcal{D}_{\mathcal{Z}}$ the marginal distribution of state $z\in \mathcal{Z}$.
\begin{assumption}[Data Coverage]\label{as:log}
     We assume that offline dataset for the COI-auditor (Definition~\ref{def:auditor}) is collected by a historical logging policy $\pi_{\text{log}} : \mathcal{Z} \to \Delta{\mathcal{A}}$. Specifically, each sample $(z, a, o)$ is drawn independently from a joint distribution $\mathcal{D}_{\text{log}}$ defined by the following procedure:
$$
z \sim \mathcal{D}_{\mathcal{Z}}, \quad a \sim \pi_{\text{log}}(z), \quad o \sim p^*(\cdot \mid z, a).
$$
We assume there exists a strictly positive constant $\gamma \in (0, 1)$ such that for all states $z \in \mathcal{Z}$ and all actions $a \in \mathcal{A}$, the logging policy satisfies:
$$
\pi_{\text{log}}(a \mid z) \geq \gamma.
$$
\end{assumption}
We make the following assumptions on the loss family $\mathcal{L}$, cost family $\mathcal{C}$, and base collaboration policies $\mathcal{H}$. 
\begin{assumption}\label{as:complexity}
    We assume that the loss, cost, and policy families \((L,C,H)\) satisfy the following properties:
    
        $(a)$ (Boundedness of $\mathcal{L}$). The family of  losses $\mathcal{L}$ is the set of all bounded functions $\tilde{l}: \mathcal{O} \to [0,1]$. 

        $(b)$ (Finiteness and Boundedness of Cost Functions $\mathcal{C}$). The family of cost functions $\mathcal{C}$ is a finite set, where each $c \in \mathcal{C}$ is uniformly bounded such that $\sup_{x, c} |c(x)| \le B_c$.

        $(c)$ (Finite Complexity of Base Policy). We assume the deterministic policy family $\mathcal{R},\Phi,\mathcal{H}$ has finite Natarajan dimension $d_{\mathcal{R}}, d_{{\Phi}}$ and $d_{\mathcal{H}}$ respectively, and that the sets of combination rules $\mathcal{G}$ is finite. 
\end{assumption}

\textbf{Notations.} In the following context, we shall fix the logging policy $\pi_{\text{log}}$ and collect offline dataset with it. Given base policy family $\mathcal{H}$, loss family $\mathcal{L}$ and cost family $\mathcal{C}$ and a base predictive distribution $p$, we define  $\Pi_{\text{eval}}(p):=\{(\pi_{h},\omega):h\in \cH, \omega\in \Omega\}\cup \{(\pi^{\omega},\omega):\omega\in \Omega\}$. This set depends on base predictive distribution $p$ through post-processing policy $\pi^{\omega}=k^{\omega}(p)$ for a loss-cost pair $\omega=(\tilde{l},c)$.

For notation simplicity, given a base predictive distribution $p$, a pair $\omega=(\tilde{l},c)$ of an evaluation loss $\tilde{l} \in \mathcal{L}$ and a cost function $c \in \mathcal{C}$, the expected loss of a base collaboration policy $h = (r, \phi, g) \in \mathcal{H}$ is defined as:
$$V(h; \omega; p) \triangleq \mathbb{E}_{z} \mathbb{E}_{o\sim p(\cdot|z,\pi_h(z))}[\ell^\omega_{\pi_{h}(z)}(z,o)].$$

Similarly, for a random general policy $\pi:\mathcal{Z}\to \Delta{\mathcal{A}}$ the expected loss is:
$$V(\pi; \omega; p) \triangleq \mathbb{E}_{z} \mathbb{E}_{a \sim \pi(\cdot|z)} \mathbb{E}_{o \sim p(\cdot | z, a)} \ell^\omega_a(z,o).$$

\begin{remark}
This compact action-based notation simplifies the definitions
introduced in Section~\ref{sec:method}. Under this notation, a
base predictive distribution $p$ satisfies
$(\Omega,\mathcal H,\varepsilon)$-COI if and only if both of
the following conditions hold:
\begin{enumerate}
    \item \textbf{Hypothesis-OI.}
    The simulated and true expected losses differ by at most
    $\varepsilon$ across all predefined base collaboration
    policies and loss-cost pairs:
    \[
    \sup_{h\in\mathcal H,\,\omega\in\Omega}
    \left|
    V(\pi_h;\omega;p)-V(\pi_h;\omega;p^\ast)
    \right|
    \leq\varepsilon.
    \]

    \item \textbf{Decision-OI.}
    For every loss-cost pair $\omega\in\Omega$, the simulated
    and true expected losses of the post-processed policy
    $\pi^\omega=k^\omega(p)$ differ by at most
    $\varepsilon$:
    \[
    \left|
    V(\pi^\omega;\omega;p)-V(\pi^\omega;\omega;p^\ast)
    \right|
    \leq\varepsilon.
    \]
\end{enumerate}
\end{remark}

Further, for a possibly randomized policy $\pi$ and a loss-cost
pair $\omega=(\tilde l,c)\in\Omega$, we define the expected COI
gap as the true expected loss minus the predicted expected loss:
\[
\begin{aligned}
G(\pi;\omega;p)
&=
\mathbb E_z\mathbb E_{a\sim\pi(\cdot\mid z)}
\left[
\mathbb E_{o\sim p^\ast(\cdot\mid z,a)}
[\ell_a^\omega(z,o)]
-
\mathbb E_{o\sim p(\cdot\mid z,a)}
[\ell_a^\omega(z,o)]
\right]\\
&=
V(\pi;\omega;p^\ast)-V(\pi;\omega;p).
\end{aligned}
\]


Then we use the following Inverse Propensity Score (IPS) method to establish an empirical estimator  given $m$-samples $\{(z_i,a_i,o_i)\}_{i=1}^{m}$ collected via policy $\pi_{\text{log}}$. Specifically,  we define the empirical COI-violation for a base predictive distribution $p$ with respect to a  collaboration policy $\pi$ as:
$$\hat{G}_m(\pi;\omega; p) = \frac{1}{m} \sum_{i=1}^m {\left[ \frac{\pi(a_i|z_i)}{\pi_{\text{log}}(a_i|z_i)} \ell^\omega_{a_i}(z_i, o_i) - \sum_{a \in \mathcal{A}} \pi(a|z_i) \mathbb{E}_{o \sim p(\cdot|z_i, a)}[\ell^\omega_a(z_i, o)] \right]}.$$

\subsection{COI implies Omnimechanism}

In this subsection, we prove that $(\Omega,\mathcal{H},\varepsilon)$-COI implies $(\Omega,\mathcal{H},2\varepsilon+\frac{\log|\cA|}{\tau})$-omnimechanism.
\begin{theorem}[Restatement of Theorem~\ref{thm:coi_implies_omni}]
    Suppose a base predictive distribution $p:\mathcal{Z}\times\mathcal{A}\to \Delta{O}$ satisfies $(\Omega,\mathcal{H},\varepsilon)$-COI at level $\varepsilon > 0$. Then, for any evaluation loss $\tilde{l} \in \mathcal{L}$ and cost $c \in \mathcal{C}$, the extracted post-processing collaboration policy $\pi^\omega = k^\omega(p)$ achieves:$$V(\pi^\omega; \omega; p^*) \le \min_{h \in \mathcal{H}} V(h; \omega; p^*) + 2\varepsilon + \frac{\log(2 |\mathcal{S}| |\mathcal{G}|)}{\tau}.$$
\end{theorem}



\begin{proof}[Proof of Theorem~\ref{thm:coi_implies_omni}]
\label{pf:coi_implies_omni}
Fix an arbitrary evaluation loss $\tilde l\in\mathcal L$,
cost $c\in\mathcal C$, and base collaboration policy
$h\in\mathcal H$, with induced deterministic policy $\pi_h$.
Write $\omega=(\tilde l,c)$. For a state $z=(x,y_M)$ and
an action $a=(r,s,g)\in\mathcal A$, define the simulated
expected loss under $p$ as
\[
L_p(z,a)
\triangleq
\mathbb E_{o\sim p(\cdot\mid z,a)}[\ell_a^\omega(z,o)]
=
\mathbb E_{o\sim p(\cdot\mid z,a)}[\tilde l(o)+r c(x)].
\]

For finite $\tau>0$, the post-processed policy
$\pi^\omega=k^\omega(p)$ is given by
\[
\pi^\omega(a\mid z)
=
\frac{\exp(-\tau L_p(z,a))}
{\sum_{a'\in\mathcal A}\exp(-\tau L_p(z,a'))}.
\]
At each state $z$, its simulated expected loss satisfies
\[
\begin{aligned}
\sum_{a\in\mathcal A}\pi^\omega(a\mid z)L_p(z,a)
&=
-\frac1\tau
\log\left(\sum_{a'\in\mathcal A}\exp(-\tau L_p(z,a'))\right)
+\frac1\tau H(\pi^\omega(\cdot\mid z))\\
&\leq
-\frac1\tau
\log\left(\exp\left(-\tau\min_{a'\in\mathcal A}L_p(z,a')\right)\right)
+\frac{\log|\mathcal A|}{\tau}\\
&=
\min_{a'\in\mathcal A}L_p(z,a')
+\frac{\log|\mathcal A|}{\tau}\\
&\leq
L_p(z,\pi_h(z))+\frac{\log|\mathcal A|}{\tau},
\end{aligned}
\]
where $H(\pi^\omega(\cdot\mid z))$ denotes the entropy
of the action distribution and is at most
$\log|\mathcal A|$. Taking expectation over $z$ gives
\[
V(\pi^\omega;\omega;p)
\leq
V(\pi_h;\omega;p)+\frac{\log|\mathcal A|}{\tau}.
\]

By Decision-OI and Hypothesis-OI, respectively, we obtain
\[
\begin{aligned}
V(\pi^\omega;\omega;p^\ast)
&\leq V(\pi^\omega;\omega;p)+\varepsilon\\
&\leq V(\pi_h;\omega;p)+\varepsilon
+\frac{\log|\mathcal A|}{\tau}\\
&\leq V(\pi_h;\omega;p^\ast)+2\varepsilon
+\frac{\log|\mathcal A|}{\tau}.
\end{aligned}
\]
Since this holds for every $\tilde l\in\mathcal L$,
$c\in\mathcal C$, and $h\in\mathcal H$, the claimed
guarantee follows.

When $\tau=\infty$, the post-processing rule selects
an action minimizing $L_p(z,a)$ at each state.
The same comparison therefore holds with zero
smoothing error.
\end{proof}



\subsection{Convergence of the COI-Boost Algorithm}
In this section, we prove the convergence of the COI-boosting algorithm (Algorithm~\ref{alg:coiboosting}), showing that it will always terminate in finite round and output an $(\Omega,\mathcal{H},\varepsilon)$-COI base predictive distribution with high probability.
\begin{theorem}[Convergence of Boosting Algorithm~\ref{alg:coiboosting}]
\label{thm:convergence}
Under Assumption~\ref{as:log}, suppose
$0\leq\tilde l(o)\leq1$ for all $\tilde l\in\mathcal L$
and $o\in\mathcal O$. For $\varepsilon\in(0,1]$ and
$\delta\in(0,1)$, set $
\eta=\frac{\gamma\varepsilon}{4},\ T_{\max}
=
\left\lceil
\frac{16\log|\mathcal O|}{\gamma\varepsilon^2}
\right\rceil. $
Run Algorithm~\ref{alg:coiboosting} for $T_{\max}$ steps, using an independent batch
of logged samples at each auditing call and an auditor
with failure probability at most $\delta$ per call.
Then, with probability at least $1-T_{\max}\delta$,
the algorithm terminates within $T_{\max}$ auditing
calls and returns a valid
$(\Omega,\mathcal H,\varepsilon)$-COI predictor.
\end{theorem}

\begin{proof}[Proof of Theorem~\ref{thm:convergence}]
If $|\mathcal O|=1$, the initial predictor equals
$p^\ast$, and the conclusion is immediate.
Hence, assume $|\mathcal O|\geq2$. Conditional on the history before each auditing call,
the current predictor is fixed and its audit batch is
independent of that history. Thus, the auditor fails
with probability at most $\delta$ at each call.
Let $\mathcal E$ be the event that every auditing call
made by the algorithm succeeds. Since there are at most
$T_{\max}$ calls, the union bound gives
\[
\mathbb P(\mathcal E)\geq1-T_{\max}\delta.
\]
Condition on $\mathcal E$. If the auditor returns
$\perp$ at iteration $t$, the auditor guarantee implies that $p^{(t)}$ satisfies
$(\Omega,\mathcal H,\varepsilon)$-COI. We now show
that this must occur within $T_{\max}$ auditing calls.

We track the expected KL-divergence potential
\[
\Phi_t
=
\mathbb E_{z\sim\mathcal D_Z}
\mathbb E_{a\sim\pi_{\mathrm{log}}(\cdot\mid z)}
\left[
\operatorname{KL}\left(
p^\ast(\cdot\mid z,a)
\,\middle\|\,
p^{(t)}(\cdot\mid z,a)
\right)
\right].
\]
We have $\Phi_t\geq0$, and uniform initialization gives
$\Phi_0\leq\log|\mathcal O|$.

At an iteration where the auditor returns a certificate,
write its output as $(\pi^{(t)},\omega^{(t)},\sigma)$,
where $\omega^{(t)}=(\tilde l^{(t)},c^{(t)})$.
For $a=(r,s,g)$, the total loss is
\[
\ell_a^{\omega^{(t)}}(z,o)
=
\tilde l^{(t)}(o)+r c^{(t)}(x).
\]
The cost term is independent of $o$, so its exponential
factor cancels between the numerator and denominator
of the normalized update. Therefore, for the following
analysis, we may equivalently use the score
\[
g^{(t)}(z,a,o)
=
\sigma
\frac{\pi^{(t)}(a\mid z)}
{\pi_{\mathrm{log}}(a\mid z)}
\tilde l^{(t)}(o).
\]
This replacement leaves every predictor $p^{(t)}$
unchanged. By the auditor guarantee,
\[
\sigma
\mathbb E_{z\sim\mathcal D_Z}
\mathbb E_{a\sim\pi^{(t)}(\cdot\mid z)}
\left[
\mathbb E_{o\sim p^\ast(\cdot\mid z,a)}
[\ell_a^{\omega^{(t)}}(z,o)]
-
\mathbb E_{o\sim p^{(t)}(\cdot\mid z,a)}
[\ell_a^{\omega^{(t)}}(z,o)]
\right]
>
\frac{\varepsilon}{2}.
\]
The outcome-independent cost also cancels in this
difference. Consequently,
\[
\mathbb E_{z\sim\mathcal D_Z}
\mathbb E_{a\sim\pi_{\mathrm{log}}(\cdot\mid z)}
\left[
\mathbb E_{o\sim p^\ast(\cdot\mid z,a)}
g^{(t)}(z,a,o)
-
\mathbb E_{o\sim p^{(t)}(\cdot\mid z,a)}
g^{(t)}(z,a,o)
\right]
>
\frac{\varepsilon}{2}.
\]

Using the positive exponential update, we obtain
\[
\begin{aligned}
\Phi_t-\Phi_{t+1}
&=
\mathbb E_{z\sim\mathcal D_Z,\,a\sim\pi_{\mathrm{log}}(\cdot\mid z)}
\left[
\sum_{o\in\mathcal O}
p^\ast(o\mid z,a)
\log\frac{p^{(t+1)}(o\mid z,a)}{p^{(t)}(o\mid z,a)}
\right]\\
&=
\mathbb E_{z\sim\mathcal D_Z,\,a\sim\pi_{\mathrm{log}}(\cdot\mid z)}
\left[
\eta\mathbb E_{o\sim p^\ast(\cdot\mid z,a)}
g^{(t)}(z,a,o)
-
\log\left(
\mathbb E_{o\sim p^{(t)}(\cdot\mid z,a)}
\exp(\eta g^{(t)}(z,a,o))
\right)
\right].
\end{aligned}
\]

By Assumption~\ref{as:log} and the boundedness of
$\tilde l^{(t)}$, we have $|g^{(t)}|\leq1/\gamma$.
Since $\eta=\gamma\varepsilon/4$ and $\varepsilon\leq1$,
\[
|\eta g^{(t)}|\leq\frac14.
\]
Applying
$\exp(\eta g^{(t)})\leq
1+\eta g^{(t)}+\eta^2(g^{(t)})^2$
and $\log x\leq x-1$ yields
\[
\begin{aligned}
\Phi_t-\Phi_{t+1}
&\geq
\eta
\mathbb E_{z\sim\mathcal D_Z,\,a\sim\pi_{\mathrm{log}}(\cdot\mid z)}
\left[
\mathbb E_{o\sim p^\ast(\cdot\mid z,a)}g^{(t)}(z,a,o)
-
\mathbb E_{o\sim p^{(t)}(\cdot\mid z,a)}g^{(t)}(z,a,o)
\right]\\
&\quad-
\eta^2
\mathbb E_{z\sim\mathcal D_Z,\,a\sim\pi_{\mathrm{log}}(\cdot\mid z),\,o\sim p^{(t)}(\cdot\mid z,a)}
\left[(g^{(t)}(z,a,o))^2\right].
\end{aligned}
\]
Moreover,
\[
\begin{aligned}
&
\mathbb E_{z\sim\mathcal D_Z,\,a\sim\pi_{\mathrm{log}}(\cdot\mid z),\,o\sim p^{(t)}(\cdot\mid z,a)}
\left[(g^{(t)}(z,a,o))^2\right]\\
&=
\mathbb E_{z\sim\mathcal D_Z}
\left[
\sum_{a\in\mathcal A}
\frac{(\pi^{(t)}(a\mid z))^2}{\pi_{\mathrm{log}}(a\mid z)}
\mathbb E_{o\sim p^{(t)}(\cdot\mid z,a)}
\left[(\tilde l^{(t)}(o))^2\right]
\right]\\
&\leq
\frac1\gamma
\mathbb E_{z\sim\mathcal D_Z}
\left[
\sum_{a\in\mathcal A}(\pi^{(t)}(a\mid z))^2
\right]
\leq\frac1\gamma.
\end{aligned}
\]
Combining these bounds with the signed-gap guarantee,
every update satisfies
\[
\Phi_t-\Phi_{t+1}
>
\frac{\eta\varepsilon}{2}-\frac{\eta^2}{\gamma}
=
\frac{\gamma\varepsilon^2}{16}.
\]

If all $T_{\max}$ auditing calls returned certificates,
the algorithm would perform $T_{\max}$ updates, giving
\[
0\leq\Phi_{T_{\max}}
<
\Phi_0-\frac{\gamma\varepsilon^2}{16}T_{\max}
\leq
\log|\mathcal O|
-\frac{\gamma\varepsilon^2}{16}T_{\max}
\leq0,
\]
a contradiction. Hence, on $\mathcal E$, an auditor
returns $\perp$ within $T_{\max}$ calls, and the
returned predictor satisfies
$(\Omega,\mathcal H,\varepsilon)$-COI.
\end{proof}






\begin{remark}
    We remark here that the above convergence theorem does not depend on the  Lipschitzness of policy with smoothing constant $\tau<\infty$, so it applies to $\tau=\infty$.
\end{remark}

\subsection{Sample Efficiency of Auditor}
In this subsection, we prove our main efficiency results.  

\begin{theorem}[Restatement of Theorem~\ref{thm:upper}]
Under Assumptions~\ref{as:log} and~\ref{as:complexity},
fix $\tau>0$ and $\varepsilon,\delta\in(0,1)$.
For a base predictive distribution $p$ independent of
the audit batch, there exists an
$(\Omega,\mathcal{H},\varepsilon,\delta,m)$-COI auditor
using $m$ independent offline samples collected under
$\pi_{\mathrm{log}}$, provided that
\[
m \gtrsim
\frac{(1+B_c)^2}{\gamma^2\varepsilon^2}
\left(
(d_{\mathcal R}+d_{\Phi}+\log_2|\mathcal G|)\log|\mathcal A|
+(1+\tau)^2|\mathcal O|
+\log\frac{2|\mathcal C|}{\delta}
\right),
\]
with a sufficiently large universal implicit constant.
\end{theorem}



\begin{proof}
We first construct the auditor by defining its empirical
objective, then separately derive the sample complexity
for Hypothesis-OI and Decision-OI.
Throughout the proof, we use the sign convention
\[
G(\pi;\omega;p)
=
V(\pi;\omega;p^*)-V(\pi;\omega;p).
\]

\textbf{Step 1: Construction of Auditor.}

We denote the evaluation class by
\[
\Pi_{\mathrm{eval}}
=
\{(\pi^\omega,\omega):\omega\in\Omega\}
\cup
\{(\pi_h,\omega):h\in\mathcal H,\omega\in\Omega\}.
\]
The auditor approximately maximizes the absolute empirical
COI gap:
\[
\sup_{(\pi,\omega)\in\Pi_{\mathrm{eval}}}
|\widehat G_m(\pi;\omega;p)|,
\]
where $\widehat G_m$ is defined below.
It selects a pair $(\pi,\omega)\in\Pi_{\mathrm{eval}}$
whose objective value is within $\varepsilon/8$
of this supremum.

If
$|\widehat G_m(\pi;\omega;p)|>3\varepsilon/4$,
the auditor returns $(\pi,\omega,\sigma)$ with
$\sigma=\operatorname{sign}(\widehat G_m(\pi;\omega;p))$.
Otherwise, it returns $\perp$.
We will show that this rule is valid whenever
\[
\sup_{(\pi,\omega)\in\Pi_{\mathrm{eval}}}
|\widehat G_m(\pi;\omega;p)-G(\pi;\omega;p)|
\leq \frac{\varepsilon}{8}.
\]

\textbf{Step 2: Sample Complexity for Hypothesis-OI.}

Recall that the empirical COI gap is
\[
\widehat G_m(\pi;\omega;p)
=
\frac{1}{m}\sum_{i=1}^m
\left[
\frac{\pi(a_i\mid z_i)}{\pi_{\mathrm{log}}(a_i\mid z_i)}
\ell^\omega_{a_i}(z_i,o_i)
-
\sum_{a\in\mathcal A}\pi(a\mid z_i)
\mathbb E_{o\sim p(\cdot\mid z_i,a)}
[\ell^\omega_a(z_i,o)]
\right].
\]
This is an unbiased estimator of $G(\pi;\omega;p)$
for every fixed pair $(\pi,\omega)$.

For fixed $h$ and $c$, the estimation error
$\widehat G_m(\pi_h;\omega;p)-G(\pi_h;\omega;p)$
is affine in $\tilde l\in[0,1]^{|\mathcal O|}$.
Its maximum absolute value is therefore attained
at one of the $2^{|\mathcal O|}$ vertices of this cube.

We next control the complexity of the induced
deterministic policy class.
Since $\pi_h$ retains the three action components
associated with $h=(r,\phi,g)$, we have
\[
d_{\mathcal H}
\leq d_{\mathcal R}+d_\Phi+\log_2|\mathcal G|.
\]
Indeed, partition any Natarajan-shattered set according
to the first component in which its two witness action
labels differ.
The subsets assigned to the first two components are
Natarajan-shattered by $\mathcal R$ and $\Phi$,
respectively.
If the third subset contains $q$ points, realizing
all its binary choices requires at least $2^q$
distinct choices from $\mathcal G$.
The stated bound follows.

Let $\Pi_{\mathcal H}(n)$ denote the maximum number
of action labelings induced on $n$ contexts.
The multiclass Sauer bound gives, for
$n\geq d_{\mathcal H}\geq1$,
\[
\Pi_{\mathcal H}(n)
\leq
\sum_{j=0}^{d_{\mathcal H}}
\binom{n}{j}\binom{|\mathcal A|}{2}^{j}
\leq
\left(\frac{en|\mathcal A|^2}{d_{\mathcal H}}\right)^{d_{\mathcal H}}.
\]

For fixed $\omega$, consider the scalar IPS contributions
indexed by $h\in\mathcal H$:
\[
(z,a,o)\longmapsto
\frac{\mathbf 1\{\pi_h(z)=a\}}{\pi_{\mathrm{log}}(a\mid z)}
\ell^\omega_a(z,o)
-
\mathbb E_{o'\sim p(\cdot\mid z,\pi_h(z))}
[\ell^\omega_{\pi_h(z)}(z,o')].
\]
Each contribution has absolute value at most
$2(1+B_c)/\gamma$.
On any collection of logged observations, its values
are determined by the action labels assigned by $\pi_h$.
Consequently, if this scalar class pseudo-shatters
$n$ observations, then
\[
2^n\leq\Pi_{\mathcal H}(n).
\]
The growth bound above therefore implies that its
pseudo-dimension is
$\mathcal O(d_{\mathcal H}\log|\mathcal A|)$.
When $d_{\mathcal H}=0$, the induced policy class
contains only one distinct policy, and the same
concentration conclusion follows directly.

Applying the VC-subgraph covering bound and Dudley's
entropy integral, followed by the standard Rademacher
generalization bound, controls this scalar class
uniformly over $h$.
Taking a union bound over the $2^{|\mathcal O|}$
loss vertices and $c\in\mathcal C$ gives, with
probability at least $1-\delta/2$,
\[
\sup_{h\in\mathcal H,\,\omega\in\Omega}
\left|
\widehat G_m(\pi_h;\omega;p)-G(\pi_h;\omega;p)
\right|
\lesssim
\frac{1+B_c}{\gamma\sqrt m}
\sqrt{
d_{\mathcal H}\log|\mathcal A|
+|\mathcal O|
+\log\frac{2|\mathcal C|}{\delta}
}.
\]

\textbf{Step 3: Sample Complexity for Decision-OI.}

Fix a cost $c\in\mathcal C$ and write
$\omega=(\tilde l,c)$.
For an action $a=(r,s,g)$ and a logged observation
$(z_i,a_i,o_i)$, define
\[
g_i(a;\tilde l)
=
\frac{\mathbf 1\{a_i=a\}}{\pi_{\mathrm{log}}(a_i\mid z_i)}
\ell^\omega_a(z_i,o_i)
-
\mathbb E_{o\sim p(\cdot\mid z_i,a)}
[\ell^\omega_a(z_i,o)].
\]
For a policy $\pi$, set
\[
g_i(\pi,\tilde l)
=
\sum_{a\in\mathcal A}\pi(a\mid z_i)g_i(a;\tilde l).
\]
Thus
\[
\widehat G_m(\pi^\omega;\omega;p)
=
\frac1m\sum_{i=1}^m g_i(\pi^\omega,\tilde l).
\]

\begin{lemma}\label{lm:app1}
For fixed $c$, the function
$\tilde l\mapsto g_i(\pi^{(\tilde l,c)},\tilde l)$
is $\frac{4(1+B_c)(1+\tau)}{\gamma}$-Lipschitz
with respect to the $\ell_\infty$ norm on
$[0,1]^{|\mathcal O|}$.
\end{lemma}

\begin{proof}[Proof of Lemma~\ref{lm:app1}]
Using the product rule,
\[
\nabla_{\tilde l}g_i(\pi^\omega,\tilde l)
=
\sum_{a\in\mathcal A}
\left[
g_i(a;\tilde l)\nabla_{\tilde l}\pi^\omega(a\mid z_i)
+
\pi^\omega(a\mid z_i)\nabla_{\tilde l}g_i(a;\tilde l)
\right].
\]
By the definition of the softmax policy and the fact
that each $p(\cdot\mid z_i,a)$ is a probability distribution,
\[
\|\nabla_{\tilde l}\pi^\omega(a\mid z_i)\|_1
\leq 2\tau\pi^\omega(a\mid z_i).
\]
Also,
\[
|g_i(a;\tilde l)|
\leq (1+B_c)\left(1+\frac1\gamma\right)
\leq \frac{2(1+B_c)}{\gamma}.
\]
Hence the first term satisfies
\[
\sum_{a\in\mathcal A}
|g_i(a;\tilde l)|
\|\nabla_{\tilde l}\pi^\omega(a\mid z_i)\|_1
\leq \frac{4\tau(1+B_c)}{\gamma}.
\]

For the second term, differentiating the affine
dependence on $\tilde l$ gives
\[
\|\nabla_{\tilde l}g_i(a;\tilde l)\|_1
\leq
1+\frac{\mathbf 1\{a_i=a\}}
{\pi_{\mathrm{log}}(a_i\mid z_i)}
\leq 1+\frac1\gamma.
\]
Therefore,
\[
\|\nabla_{\tilde l}g_i(\pi^\omega,\tilde l)\|_1
\leq
\frac{4\tau(1+B_c)}{\gamma}
+1+\frac1\gamma
\leq
\frac{4(1+B_c)(1+\tau)}{\gamma}.
\]
Since the loss cube is convex, this gradient bound
establishes the claimed Lipschitz property with respect
to the $\ell_\infty$ norm.
\end{proof}

For fixed $c$, let $\mathcal F_c$ be the following
class of functions on logged triples:
\[
\mathcal F_c
=
\left\{
(z,a,o)\longmapsto
\frac{\pi^{(\tilde l,c)}(a\mid z)}
{\pi_{\mathrm{log}}(a\mid z)}
\ell^{(\tilde l,c)}_a(z,o)
-
\sum_{a'\in\mathcal A}\pi^{(\tilde l,c)}(a'\mid z)
\mathbb E_{o'\sim p(\cdot\mid z,a')}
[\ell^{(\tilde l,c)}_{a'}(z,o')]
:
\tilde l\in[0,1]^{|\mathcal O|}
\right\}.
\]

For $0<u\leq1$, the loss cube admits an
$\ell_\infty$ cover of size at most
$(1+2/u)^{|\mathcal O|}$.
By Lemma~\ref{lm:app1}, this induces a cover of
$\mathcal F_c$ under the normalized empirical
$L_2$ metric with radius
$4(1+B_c)(1+\tau)u/\gamma$.

Subtracting the member corresponding to $\tilde l=0$
does not change empirical Rademacher complexity.
Applying Dudley's entropy integral to the resulting
centered class gives
\[
\widehat{\mathfrak R}_m(\mathcal F_c)
\lesssim
\frac{(1+B_c)(1+\tau)}{\gamma\sqrt m}
\int_0^1
\sqrt{|\mathcal O|\log(1+2/u)}\,du
\lesssim
\frac{(1+B_c)(1+\tau)}{\gamma}
\sqrt{\frac{|\mathcal O|}{m}}.
\]

Every member of $\mathcal F_c$ has absolute value
at most $2(1+B_c)/\gamma$.
The standard Rademacher generalization bound and
a union bound over $c\in\mathcal C$ therefore imply,
with probability at least $1-\delta/2$,
\[
\sup_{\omega\in\Omega}
\left|
\widehat G_m(\pi^\omega;\omega;p)
-
G(\pi^\omega;\omega;p)
\right|
\lesssim
\frac{1+B_c}{\gamma\sqrt m}
\left(
(1+\tau)\sqrt{|\mathcal O|}
+
\sqrt{\log\frac{2|\mathcal C|}{\delta}}
\right).
\]

Combining Steps 2 and 3, with probability at least
$1-\delta$,
\[
\sup_{(\pi,\omega)\in\Pi_{\mathrm{eval}}}
\left|
\widehat G_m(\pi;\omega;p)-G(\pi;\omega;p)
\right|
\lesssim
\frac{1+B_c}{\gamma\sqrt m}
\sqrt{
d_{\mathcal H}\log|\mathcal A|
+(1+\tau)^2|\mathcal O|
+\log\frac{2|\mathcal C|}{\delta}
}.
\]
Under the stated sample-size condition, this error
is at most $\varepsilon/8$.

On this event, suppose the auditor returns
$(\pi,\omega,\sigma)$.
Then
\[
\sigma G(\pi;\omega;p)
\geq
|\widehat G_m(\pi;\omega;p)|-\frac{\varepsilon}{8}
>
\frac{3\varepsilon}{4}-\frac{\varepsilon}{8}
=
\frac{5\varepsilon}{8}
>
\frac{\varepsilon}{2}.
\]
Thus every returned tuple is a valid signed
violation certificate.

Otherwise, the selected pair has absolute empirical
gap at most $3\varepsilon/4$.
Since it is an $\varepsilon/8$-approximate maximizer,
\[
\sup_{(\pi,\omega)\in\Pi_{\mathrm{eval}}}
|\widehat G_m(\pi;\omega;p)|
\leq \frac{7\varepsilon}{8}.
\]
Consequently,
\[
\sup_{(\pi,\omega)\in\Pi_{\mathrm{eval}}}
|G(\pi;\omega;p)|
\leq
\frac{7\varepsilon}{8}+\frac{\varepsilon}{8}
=
\varepsilon.
\]
Returning $\perp$ therefore certifies
$(\Omega,\mathcal H,\varepsilon)$-COI.
This completes the proof.
\end{proof}

The following lemma establishes the sample efficiency when $\tau=\infty$.

\begin{lemma}[Sample Complexity of COI Auditing for $\tau=\infty$]
\label{lm:hard_auditing}
Suppose Assumptions~\ref{as:log} and \ref{as:complexity}
hold, the action and outcome spaces are finite, and the
logging propensities are known.
Fix a base predictive distribution $p$ independently of
the audit batch.
For $\tau=\infty$, we can define $k^\omega(p)$ to assign probability
one to the first action, under a fixed ordering of
$\mathcal A$, that minimizes
\[
\mathbb E_{o\sim p(\cdot\mid z,a)}
[\ell^\omega_a(z,o)].
\]
Let
\[
\Pi_{\mathrm{eval}}(p)
=
\{(\pi_h,\omega):h\in\mathcal H,\ \omega\in\Omega\}
\cup
\{(k^\omega(p),\omega):\omega\in\Omega\}.
\]
For $\varepsilon,\delta\in(0,1)$, whenever
\[
m\gtrsim
\frac{(1+B_c)^2}{\gamma^2\varepsilon^2}
\left[
\left(
d_{\mathcal R}+d_{\Phi}+\log_2|\mathcal G|
+|\mathcal O|
\right)\log(2|\mathcal A|)
+
\log\frac{2|\mathcal C|}{\delta}
\right],
\]
with a sufficiently large universal implicit constant,
$m$ independent samples from the logging distribution
satisfy, with probability at least $1-\delta$,
\[
\sup_{(\pi,\omega)\in\Pi_{\mathrm{eval}}(p)}
\left|
\widehat G_m(\pi;\omega;p)-G(\pi;\omega;p)
\right|
\leq \frac{\varepsilon}{8}.
\]
Consequently, the auditor that approximately maximizes
the absolute empirical COI gap to additive accuracy
$\varepsilon/8$ and uses threshold $3\varepsilon/4$
is an $(\Omega,\mathcal H,\varepsilon,\delta,m)$-COI auditor.
\end{lemma}

\begin{proof}
We condition on the predictor $p$, which is independent
of the audit batch. The Hypothesis-OI analysis is unchanged
from the proof of Theorem~\ref{thm:upper}.
We establish a uniform convergence bound for the
deterministic post-processing policies and then combine
the two bounds.

\textbf{Step 1: Sample Complexity for Decision-OI.}

Fix $c\in\mathcal C$. Consider the class of scalar IPS
contributions
\[
\mathcal F_c
=
\left\{
(z,a,o)\mapsto
\frac{k^{(\widetilde l,c)}(p)(a\mid z)}
{\pi_{\mathrm{log}}(a\mid z)}
\ell_a^{(\widetilde l,c)}(z,o)
-
\sum_{a'\in\mathcal A}
k^{(\widetilde l,c)}(p)(a'\mid z)
\mathbb E_{o'\sim p(\cdot\mid z,a')}
[\ell_{a'}^{(\widetilde l,c)}(z,o')]
:
\widetilde l\in[0,1]^{|\mathcal O|}
\right\}.
\]
We show that
\[
\operatorname{Pdim}(\mathcal F_c)
\lesssim |\mathcal O|\log(2|\mathcal A|).
\]

To this end, fix arbitrary logged observations
$(z_i,a_i,o_i)_{i=1}^{n}$ and thresholds
$t_1,\ldots,t_n\in\mathbb R$.
For each action $a\in\mathcal A$, define
\[
g_i(a;\widetilde l)
=
\frac{\mathbf 1\{a_i=a\}}
{\pi_{\mathrm{log}}(a_i\mid z_i)}
\ell_a^{(\widetilde l,c)}(z_i,o_i)
-
\mathbb E_{o\sim p(\cdot\mid z_i,a)}
[\ell_a^{(\widetilde l,c)}(z_i,o)].
\]
For fixed $c$, this expression is affine in
$\widetilde l\in\mathbb R^{|\mathcal O|}$.
The scalar contribution of the post-processing policy
at observation $i$ is
\[
g_i(k^{(\widetilde l,c)}(p),\widetilde l)
=
\sum_{a\in\mathcal A}
k^{(\widetilde l,c)}(p)(a\mid z_i)
g_i(a;\widetilde l).
\]
Since the policy is deterministic, this equals
$g_i(a;\widetilde l)$ for its selected action.

The selected action at $z_i$ is determined by the signs
of the pairwise predicted-loss differences
\[
\sum_{o\in\mathcal O}
p(o\mid z_i,a)\ell_a^{(\widetilde l,c)}(z_i,o)
-
\sum_{o\in\mathcal O}
p(o\mid z_i,a')\ell_{a'}^{(\widetilde l,c)}(z_i,o),
\qquad a\neq a'.
\]
Each difference is affine in $\widetilde l$.
Their signs, including zero signs, determine the minimizing
actions, and the fixed ordering determines which minimizer
is selected. Once that action is determined, whether
the scalar contribution exceeds $t_i$ is determined by
the sign of
\[
g_i(a;\widetilde l)-t_i.
\]

Thus all $n$ threshold comparisons are determined by the
signs of at most
\[
n\bigl(|\mathcal A|(|\mathcal A|-1)+|\mathcal A|\bigr)
=
n|\mathcal A|^2
\]
affine functions of the $|\mathcal O|$ loss coordinates.
For $n\geq|\mathcal O|$, the sign-pattern bound, including
zero signs, gives at most
\[
\left(
\frac{8e\,n|\mathcal A|^2}{|\mathcal O|}
\right)^{|\mathcal O|}
\]
possible sign patterns
\citep[Corollary~2.1]{goldberg1993bounding}.
Therefore, if $\mathcal F_c$ pseudo-shatters these
$n$ observations, then
\[
2^n
\leq
\left(
\frac{8e\,n|\mathcal A|^2}{|\mathcal O|}
\right)^{|\mathcal O|}.
\]
This implies
$n\lesssim|\mathcal O|\log(2|\mathcal A|)$.
The same bound holds trivially when $n<|\mathcal O|$,
which establishes the claimed pseudo-dimension bound.

Every member of $\mathcal F_c$ has absolute value at most
\[
(1+B_c)\left(1+\frac{1}{\gamma}\right)
\leq \frac{2(1+B_c)}{\gamma}.
\]
Applying the VC-subgraph covering bound and Dudley's
entropy integral, followed by the standard Rademacher
generalization bound and a union bound over $c\in\mathcal C$,
gives, with probability at least $1-\delta/2$,
\[
\sup_{\omega\in\Omega}
\left|
\widehat G_m(k^\omega(p);\omega;p)
-
G(k^\omega(p);\omega;p)
\right|
\lesssim
\frac{1+B_c}{\gamma\sqrt m}
\sqrt{
|\mathcal O|\log(2|\mathcal A|)
+
\log\frac{2|\mathcal C|}{\delta}
}.
\]

\textbf{Step 2: Combining Hypothesis-OI and Decision-OI.}

The Hypothesis-OI argument in the proof of
Theorem~\ref{thm:upper} gives, with probability at least
$1-\delta/2$,
\[
\sup_{h\in\mathcal H,\ \omega\in\Omega}
\left|
\widehat G_m(\pi_h;\omega;p)
-
G(\pi_h;\omega;p)
\right|
\lesssim
\frac{1+B_c}{\gamma\sqrt m}
\sqrt{
d_{\mathcal H}\log(2|\mathcal A|)
+
|\mathcal O|
+
\log\frac{2|\mathcal C|}{\delta}
}.
\]
Combining the two bounds, with probability at least
$1-\delta$,
\[
\sup_{(\pi,\omega)\in\Pi_{\mathrm{eval}}(p)}
\left|
\widehat G_m(\pi;\omega;p)-G(\pi;\omega;p)
\right|
\lesssim
\frac{1+B_c}{\gamma\sqrt m}
\sqrt{
(d_{\mathcal H}+|\mathcal O|)\log(2|\mathcal A|)
+
\log\frac{2|\mathcal C|}{\delta}
}.
\]
Using
$d_{\mathcal H}\leq
d_{\mathcal R}+d_{\Phi}+\log_2|\mathcal G|$,
the stated sample-size condition makes this error
at most $\varepsilon/8$.

On this event, let $(\pi,\omega)$ be a pair whose absolute
empirical gap is within $\varepsilon/8$ of the supremum
over $\Pi_{\mathrm{eval}}(p)$.
If its absolute empirical gap exceeds $3\varepsilon/4$,
return the pair with
$\sigma=\operatorname{sign}(\widehat G_m(\pi;\omega;p))$.
Then
\[
\sigma G(\pi;\omega;p)
\geq
|\widehat G_m(\pi;\omega;p)|-\frac{\varepsilon}{8}
>
\frac{5\varepsilon}{8}
>
\frac{\varepsilon}{2}.
\]
Otherwise, the supremum of the absolute empirical gap
is at most $7\varepsilon/8$, and hence
\[
\sup_{(\pi,\omega)\in\Pi_{\mathrm{eval}}(p)}
|G(\pi;\omega;p)|
\leq
\frac{7\varepsilon}{8}+\frac{\varepsilon}{8}
=
\varepsilon.
\]
Returning $\perp$ therefore certifies
$(\Omega,\mathcal H,\varepsilon)$-COI.
This completes the proof.
\end{proof}

To complement the upper bounds provided in Theorem \ref{thm:upper}, we establish a lower bound on sample complexity. This ensures that the sample complexity's dependence on the Natarajan dimension $d_{\mathcal{H}}$ and the logging policy's support constant $\gamma$ is inherent to the problem of offline collaborative omniprediction, rather than an artifact of our specific auditing algorithm.

\begin{theorem}[Sample Complexity Lower Bound for Offline Omnimechanism Learning]
Let $\mathcal A$ and $\mathcal O$ be finite action and outcome spaces,
with $|\mathcal A|,|\mathcal O|\geq2$.
Let $\mathcal H$ be a predefined hypothesis class of deterministic
policies with finite Natarajan dimension $d_{\mathcal H}\geq1$.
Assume that $\Omega$ contains a loss-cost pair
$\omega=(\tilde l,0)$, where
$\tilde l(o)=\mathbb I\{o=o_1\}$ for some $o_1\in\mathcal O$.

For any $\varepsilon\in(0,1/64]$,
$\delta\in(0,1/4)$, and
$\gamma\in(0,1/|\mathcal A|]$,
there exist a context distribution and a known logging policy
with minimum support $\gamma$ such that any algorithm that,
uniformly over the outcome environments, outputs an
$(\Omega,\mathcal H,\varepsilon)$-collaborative omnimechanism
with probability at least $1-\delta$ requires an offline
dataset of size
\[
m=\Omega\left(
\frac{d_{\mathcal H}+\log(1/\delta)}
{\gamma\varepsilon^2}
\right).
\]
\end{theorem}

\begin{proof}
The proof proceeds by reduction.
We construct a family of outcome distributions, which we refer
to as environments to distinguish them from the single
ground-truth distribution $p^*$ in the main text.
We show that an algorithm producing an
$(\Omega,\mathcal H,\varepsilon)$-collaborative omnimechanism
must identify these environments with small Hamming error.
We then establish the lower bound using information-theoretic
inequalities.

\textbf{Step 1: Construction of the State Space and Logging Policy.}

Write $d=d_{\mathcal H}$.
By the definition of Natarajan dimension, there exist states
$\mathcal Z_{\mathrm{hard}}=\{z_1,\ldots,z_d\}$
and two distinct actions $a_{0,j},a_{1,j}\in\mathcal A$
at each state $z_j$ such that every selection of one action
from each pair is realized by some policy in $\mathcal H$. Let the context distribution be uniform on these states:
\[
\mathbb P(z=z_j)=\frac1d,
\qquad j\in\{1,\ldots,d\}.
\]
Choose the logging policy
\[
\pi_{\mathrm{log}}(a\mid z_j)=
\begin{cases}
\gamma, & a=a_{1,j},\\[2pt]
\dfrac{1-\gamma}{|\mathcal A|-1}, & a\neq a_{1,j}.
\end{cases}
\]
Since $\gamma\leq1/|\mathcal A|$, every action has logging
probability at least $\gamma$.
Outside $\mathcal Z_{\mathrm{hard}}$, extend the logging policy
arbitrarily while preserving this coverage condition.

\textbf{Step 2: Construction of Hard Performative Environments.}

Choose an outcome $o_0\in\mathcal O$ distinct from $o_1$.
We parameterize the environments by
$v\in\{-1,1\}^d$.
Each environment is supported on $\{o_0,o_1\}$ and satisfies
\[
P_v(o_1\mid z_j,a)=
\begin{cases}
\frac12-16\varepsilon v_j, & a=a_{1,j},\\[2pt]
\frac12, & a\neq a_{1,j},
\end{cases}
\qquad
P_v(o_0\mid z_j,a)=1-P_v(o_1\mid z_j,a).
\]
Because $\varepsilon\leq1/64$, these probabilities lie in
$[1/4,3/4]$. Consider the loss-cost pair $\omega=(\tilde l,0)$ in the
theorem statement. Since $\tilde l(o)=\mathbb I\{o=o_1\}$,
the expected loss of action $a_{1,j}$ is
$1/2-16\varepsilon v_j$, while every other action has
expected loss $1/2$.

When $v_j=1$, the optimal action is $a_{1,j}$.
When $v_j=-1$, the action $a_{0,j}$ is optimal.
By Natarajan shattering, $\mathcal H$ contains a policy
making these optimal choices simultaneously at all selected
states.

Write $V(\pi;P_v)$ for the expected loss of policy $\pi$
under environment $P_v$ and the fixed loss-cost pair $\omega$.
For any randomized policy $\pi$,
\[
V(\pi;P_v)-\min_{h\in\mathcal H}V(h;P_v)
=
\frac{16\varepsilon}{d}
\sum_{j=1}^d
\left[
\mathbb I\{v_j=1\}\bigl(1-\pi(a_{1,j}\mid z_j)\bigr)
+
\mathbb I\{v_j=-1\}\pi(a_{1,j}\mid z_j)
\right].
\]

\textbf{Step 3: Reduction from Omnimechanism Learning to Environment Identification.}

Suppose the learner outputs a base predictive distribution
$p$, with post-processed policy $\pi^\omega=k^\omega(p)$.
Whenever $p$ is an
$(\Omega,\mathcal H,\varepsilon)$-collaborative omnimechanism,
\[
V(\pi^\omega;P_v)-\min_{h\in\mathcal H}V(h;P_v)
\leq\varepsilon.
\]
Define
\[
\hat v_j=
\begin{cases}
1, & \pi^\omega(a_{1,j}\mid z_j)\geq1/2,\\
-1, & \pi^\omega(a_{1,j}\mid z_j)<1/2,
\end{cases}
\]
and let $E=\{j:\hat v_j\neq v_j\}$. For every $j\in E$, the deployed policy places probability
at least $1/2$ on suboptimal actions at state $z_j$.
The excess-loss identity from Step 2 therefore gives
\[
\frac{8\varepsilon}{d}|E|
\leq
V(\pi^\omega;P_v)-\min_{h\in\mathcal H}V(h;P_v)
\leq\varepsilon.
\]
Consequently, whenever the learner succeeds,
\[
\mathrm{Hamming}(\hat v,v)=|E|\leq\frac d8.
\]
Since the learner succeeds with probability at least
$1-\delta$ under every environment and the Hamming error
is always at most $d$,
\[
\sup_v\mathbb E_v[\mathrm{Hamming}(\hat v,v)]
\leq\frac d8+\delta d
<\frac{3d}{8}.
\]

\textbf{Step 4: Information-Theoretic Distinguishability.}

Let $\mathbb P_v$ denote the distribution of one logged
sample $(z,a,o)$ under the context distribution, logging
policy, and environment $P_v$ constructed above.
Consider two environments $v,v'$ differing only in
coordinate $j$.
Their logged-sample distributions differ only when
$z=z_j$ and $a=a_{1,j}$, so
\[
D_{\mathrm{KL}}(\mathbb P_v\parallel\mathbb P_{v'})
=
\frac{\gamma}{d}
D_{\mathrm{KL}}\left(
\mathrm{Ber}\left(\frac12-16\varepsilon v_j\right)
\parallel
\mathrm{Ber}\left(\frac12+16\varepsilon v_j\right)
\right).
\]
For Bernoulli parameters in $[1/4,3/4]$, the inequality
\[
D_{\mathrm{KL}}(\mathrm{Ber}(p)\parallel\mathrm{Ber}(q))
\leq\frac{(p-q)^2}{q(1-q)}
\]
implies
\[
D_{\mathrm{KL}}(\mathbb P_v\parallel\mathbb P_{v'})
\lesssim\frac{\gamma\varepsilon^2}{d}.
\]
By additivity of KL divergence for independent samples,
\[
D_{\mathrm{KL}}(
\mathbb P_v^{\otimes m}
\parallel
\mathbb P_{v'}^{\otimes m})
\lesssim\frac{m\gamma\varepsilon^2}{d}.
\]
Here and below, the implicit constants are universal.

\textbf{Step 5: Applying Assouad's Lemma and a Two-Environment Testing Bound.}

Assouad's lemma together with Pinsker's inequality gives
\[
\sup_v\mathbb E_v[\mathrm{Hamming}(\hat v,v)]
\geq
\frac d2
\left(
1-
\sqrt{
\frac12
\max_{\mathrm{Hamming}(v,v')=1}
D_{\mathrm{KL}}(
\mathbb P_v^{\otimes m}
\parallel
\mathbb P_{v'}^{\otimes m})
}
\right).
\]
If the maximum adjacent-environment KL divergence is at most
$1/8$, this lower bound is at least $3d/8$, contradicting
Step 3.
Combining this observation with Step 4 yields
\[
m\gtrsim\frac{d}{\gamma\varepsilon^2}.
\]

To establish the dependence on $\delta$, consider the two
environments
\[
v=(1,\ldots,1),
\qquad
v'=(-1,\ldots,-1).
\]
Whenever the learner succeeds, at most $d/8$ coordinates of
$\hat v$ are incorrect.
Thus a majority vote over the coordinates of $\hat v$
distinguishes these two environments with error probability
at most $\delta$ under each environment. These environments differ only on the actions
$a_{1,j}$, whose total logging probability is $\gamma$.
The same KL calculation therefore gives
\[
D_{\mathrm{KL}}(
\mathbb P_v^{\otimes m}
\parallel
\mathbb P_{v'}^{\otimes m})
\lesssim m\gamma\varepsilon^2.
\]
On the other hand, the testing guarantee and the
Bretagnolle--Huber inequality imply
\[
1-2\delta
\leq
\operatorname{TV}(
\mathbb P_v^{\otimes m},
\mathbb P_{v'}^{\otimes m})
\leq
\sqrt{
1-\exp\left(
-D_{\mathrm{KL}}(
\mathbb P_v^{\otimes m}
\parallel
\mathbb P_{v'}^{\otimes m})
\right)
}.
\]
Consequently,
\[
m\gamma\varepsilon^2
\gtrsim
\log\frac{1}{4\delta(1-\delta)}
\gtrsim
\log\frac1\delta,
\]
where the last inequality holds uniformly for
$\delta\in(0,1/4)$.
Hence,
\[
m\gtrsim\frac{\log(1/\delta)}{\gamma\varepsilon^2}.
\]

Combining the two necessary bounds and using
$\max\{d,\log(1/\delta)\}\geq
(d+\log(1/\delta))/2$, we obtain
\[
m=\Omega\left(
\frac{d_{\mathcal H}+\log(1/\delta)}
{\gamma\varepsilon^2}
\right).
\]
This completes the proof.
\end{proof}




\section{Efficiency of Decision-OI Auditing}\label{app:timeeff}

In this section, we analyze the computational efficiency
of Decision-OI auditing, following
\citet{gopalan2024computationally}.
We consider a constant query cost $c$.
For a state $z$, an action $a=(r,s,g)$, and an outcome $o$,
the total loss is independent of $z$ and can therefore
be written as
\[
\ell^\omega_a(o)=\widetilde l(o)+rc,
\]
where $\omega=(\widetilde l,c)$ and, with a slight abuse
of notation, $c$ denotes the value of the constant
cost function. Recall that $\mathcal L=[0,1]^{|\mathcal O|}$.
Given a base predictive distribution $p$ and a loss-cost
pair $\omega\in\Omega$, define the signed Decision-OI gap as
\[
G(\pi^\omega;\omega;p)
:=
\mathbb E_z\mathbb E_{a\sim\pi^\omega(\cdot\mid z)}
\left[
\mathbb E_{o\sim p^*(\cdot\mid z,a)}[\ell^\omega_a(o)]
-
\mathbb E_{o\sim p(\cdot\mid z,a)}[\ell^\omega_a(o)]
\right],
\]
where the post-processing policy is
\[
\pi^\omega(a\mid z)
\propto
\exp\left(
-\tau\mathbb E_{o\sim p(\cdot\mid z,a)}
[\ell^\omega_a(o)]
\right),
\]
and $\tau>0$ is the inverse temperature.
A base predictive distribution $p$ satisfies
$(\Omega,\varepsilon)$-Decision-OI if
\[
\sup_{\omega\in\Omega}
|G(\pi^\omega;\omega;p)|
\leq\varepsilon.
\]

Throughout this section, let $m=|\mathcal A|$ and
$k=|\mathcal A\times\mathcal O|$.
We assume uniform logging,
$\pi_{\mathrm{log}}(a\mid z)=1/m$ for every $(z,a)$,
and a fixed constant query cost $c\geq0$.
Thus,
$\Omega=\mathcal L\times\{c\}=[0,1]^{|\mathcal O|}\times\{c\}$,
and we take $B=1+c$ as an upper bound on the total loss.

Following \citet{gopalan2024computationally}, we reduce
Decision-OI auditing to weighted-calibration auditing
over a class of polynomial weight functions in a
multinomial RKHS.
This reduction yields an \emph{auditor with witness}:
the auditor returns a bounded function whose correlation
with the prediction residual provides an update direction,
although this function need not be induced by a loss-cost
pair.
Such witnesses support convergence to Decision-OI through
the corresponding boosting procedure in
Lemma~\ref{lm:boosting}.

Theorem~\ref{thm:efficiency} establishes the following
guarantee.
Fix $\tau>0$, $\varepsilon\in(0,1/e)$, and
$\delta\in(0,1/3)$.
There exists a constant $C(B,\tau)>0$, depending only
on $B,\tau$ and independent of $|\mathcal A|,|\mathcal O|$,
such that, with
\[
D=\left\lceil C(B,\tau)\log^2(4/\varepsilon)\right\rceil,
\qquad
r=\sqrt{D+1}(mB\sqrt{ek})^D,
\]
there exists an
$\bigl(\Omega,\varepsilon,
\frac{\varepsilon}{6mBr\sqrt{D+1}}\bigr)$-auditor
with witness for Decision-OI.
For a fixed input predictor, this auditor has failure
probability at most $\delta$, sample complexity
\[
n=\mathcal O\left(
\frac{km^2B^2r^2(D+1)\log(1/\delta)}{\varepsilon^2}
\right),
\]
and running time $\operatorname{poly}(n,k,D)$.

For fixed $B,\tau,\varepsilon$, these complexities
are polynomial in $m$ and $k$.
The reduction therefore avoids the exponential dependence
on $|\mathcal O|$ incurred by exhaustive discretization
of the loss cube $\mathcal L=[0,1]^{|\mathcal O|}$.
For fixed $B,\tau,m,k$, the dependence on $1/\varepsilon$
is quasi-polynomial.




\subsection{Reduction to Weighted Calibration}

We first recall the notion of weighted calibration.
For a positive integer $k$, let $\Delta_k$ denote the
probability simplex on $\{1,\ldots,k\}$ and let
$\mathcal E_k$ denote the corresponding set of one-hot
vectors.

\begin{definition}[Weighted calibration,
\citep{gopalan2024computationally}]
Let $\mathcal W$ be a family of functions
$w:\Delta_k\to[-1,1]^k$.
The $\mathcal W$-calibration error of a distribution
$\mathcal D$ on $\Delta_k\times\mathcal E_k$ is
\begin{equation}
\mathrm{CE}_{\mathcal W}(\mathcal D)
=
\sup_{w\in\mathcal W}
\left|
\mathbb E_{(\mathbf v,\mathbf y)\sim\mathcal D}
[\langle\mathbf y-\mathbf v,w(\mathbf v)\rangle]
\right|.
\end{equation}
We say that $\mathcal D$ is
$(\mathcal W,\varepsilon)$-calibrated if
$\mathrm{CE}_{\mathcal W}(\mathcal D)\leq\varepsilon$.
\end{definition}

Recall that $m=|\mathcal A|$, $k=|\mathcal A\times\mathcal O|$,
and $B=1+c$ bounds the total loss.
We now reduce Decision-OI to a weighted-calibration
condition.
Under uniform logging, draw
\[
z\sim\mathcal D_{\mathcal Z},
\qquad
a\sim\mathrm{Unif}(\mathcal A),
\qquad
o\sim p^*(\cdot\mid z,a).
\]
Given $(z,a,o)$, define vectors $\mathbf v$ and
$\mathbf y$, indexed by $(a',o')\in\mathcal A\times\mathcal O$,
by
\begin{equation}\label{eq:Dcal}
\mathbf v_{(a',o')}(z)=\frac{p(o'\mid z,a')}{m},
\qquad
\mathbf y_{(a',o')}(z,a,o)
=\mathbf 1\{a=a',\,o=o'\}.
\end{equation}
Let $\mathcal D^{\mathrm{cal}}$ denote the induced
distribution of $(\mathbf v,\mathbf y)$.
Since
\[
\sum_{(a',o')\in\mathcal A\times\mathcal O}
\mathbf v_{(a',o')}(z)
=
\frac1m\sum_{a'\in\mathcal A}
\sum_{o'\in\mathcal O}p(o'\mid z,a')
=1,
\]
we have $\mathbf v\in\Delta_k$ and
$\mathbf y\in\mathcal E_k$.

For each $\omega\in\Omega$ and
$(a',o')\in\mathcal A\times\mathcal O$, define the
scalar weight function
\[
W_{(a',o')}^\omega(\mathbf v)
=
\frac{\ell_{a'}^\omega(o')}{B}
\frac{
\exp\left(-m\tau\sum_{o\in\mathcal O}
\ell_{a'}^\omega(o)\mathbf v_{(a',o)}\right)
}{
\sum_{a\in\mathcal A}
\exp\left(-m\tau\sum_{o\in\mathcal O}
\ell_a^\omega(o)\mathbf v_{(a,o)}\right)
}.
\]
Let
\[
\mathcal W_{\mathrm{dec}}
=
\left\{
W_{(a',o')}^\omega:
\omega\in\Omega,\,
(a',o')\in\mathcal A\times\mathcal O
\right\}.
\]
Each member of $\mathcal W_{\mathrm{dec}}$ maps
$\Delta_k$ into $[0,1]$.
We write $\mathcal W_{\mathrm{dec}}^k$ for its
$k$-fold Cartesian product, viewed as a family of
functions from $\Delta_k$ to $[0,1]^k$. At the encoded prediction $\mathbf v(z)$, the softmax
factor in $W_{(a',o')}^\omega$ equals
$\pi^\omega(a'\mid z)$.
Consequently,
\[
W_{(a',o')}^\omega(\mathbf v(z))
=
\frac{\ell_{a'}^\omega(o')}{B}\pi^\omega(a'\mid z).
\]

\begin{lemma}\label{applm:reduction}
For the distribution $\mathcal D^{\mathrm{cal}}$
constructed in Eq.~\eqref{eq:Dcal},
\[
\sup_{\omega\in\Omega}
|G(\pi^\omega;\omega;p)|
\leq
mB\,\mathrm{CE}_{\mathcal W_{\mathrm{dec}}^k}
(\mathcal D^{\mathrm{cal}}).
\]
In particular, if $\mathcal D^{\mathrm{cal}}$ is
$(\mathcal W_{\mathrm{dec}}^k,\varepsilon/(mB))$-calibrated,
then $p$ satisfies $(\Omega,\varepsilon)$-Decision-OI.
\end{lemma}

\begin{proof}
Fix $\omega\in\Omega$.
By construction, the tuple
$(W_{(a',o')}^\omega)_{(a',o')\in\mathcal A\times\mathcal O}$
belongs to $\mathcal W_{\mathrm{dec}}^k$.
Under uniform logging,
\[
\mathbb E[\mathbf y_{(a',o')}\mid z]
=
\frac1m p^*(o'\mid z,a').
\]
Therefore,
\begin{align*}
&\mathbb E_{(\mathbf v,\mathbf y)\sim\mathcal D^{\mathrm{cal}}}
\left[
\sum_{(a',o')\in\mathcal A\times\mathcal O}
W_{(a',o')}^\omega(\mathbf v)
\bigl(\mathbf y_{(a',o')}-\mathbf v_{(a',o')}\bigr)
\right]
\\
&=
\frac1m\mathbb E_z
\left[
\sum_{(a',o')\in\mathcal A\times\mathcal O}
W_{(a',o')}^\omega(\mathbf v(z))
\bigl(p^*(o'\mid z,a')-p(o'\mid z,a')\bigr)
\right]
\\
&=
\frac{1}{mB}\mathbb E_z
\left[
\sum_{a'\in\mathcal A}\pi^\omega(a'\mid z)
\sum_{o'\in\mathcal O}\ell_{a'}^\omega(o')
\bigl(p^*(o'\mid z,a')-p(o'\mid z,a')\bigr)
\right]
\\
&=
\frac{1}{mB}G(\pi^\omega;\omega;p).
\end{align*}
Taking absolute values and using the definition of
weighted calibration gives
\[
|G(\pi^\omega;\omega;p)|
\leq
mB\,\mathrm{CE}_{\mathcal W_{\mathrm{dec}}^k}
(\mathcal D^{\mathrm{cal}}).
\]
Since this holds for every $\omega\in\Omega$, taking
the supremum proves the first claim.
The calibration condition then implies
$\sup_{\omega\in\Omega}|G(\pi^\omega;\omega;p)|
\leq\varepsilon$, establishing Decision-OI.
\end{proof}

\begin{definition}[Auditor with witness for Decision-OI]
Fix a loss-cost family $\Omega$, $\varepsilon,\beta>0$,
and $\delta\in(0,1/3)$.
An $(\Omega,\varepsilon,\beta)$-auditor with witness
for Decision-OI takes a base predictive distribution
$p$ and independent samples from its induced
distribution $\mathcal D^{\mathrm{cal}}$.
It returns either $\perp$ or a function
$w':\Delta_k\to[-1,1]^k$. With probability at least $1-\delta$, the output
satisfies the following guarantees:
\begin{enumerate}
    \item If the auditor returns $\perp$, then
    $\sup_{\omega\in\Omega}
    |G(\pi^\omega;\omega;p)|\leq\varepsilon.$
    \item If the auditor returns $w'$, then $
    \mathbb E_{(\mathbf v,\mathbf y)\sim\mathcal D^{\mathrm{cal}}}
    [\langle\mathbf y-\mathbf v,w'(\mathbf v)\rangle]
    \geq\frac{\beta}{2}. $
\end{enumerate}
Here the returned witness need not be induced by a
loss-cost pair.
\end{definition}

To construct the auditor, first run a witness-search
procedure with failure probability $\delta/2$.
This procedure returns a bounded function
$w':\Delta_k\to[-1,1]^k$ and guarantees correlation
at least $\beta$ whenever $p$ violates
$(\Omega,\varepsilon)$-Decision-OI.

Next, use an independent validation batch of
$\mathcal O(\beta^{-2}\log(1/\delta))$ samples to
estimate
\[
\mathbb E_{\mathcal D^{\mathrm{cal}}}
[\langle\mathbf y-\mathbf v,w'(\mathbf v)\rangle]
\]
to accuracy $\beta/4$ with probability at least
$1-\delta/2$.
This follows from Hoeffding's inequality, since
\[
|\langle\mathbf y-\mathbf v,w'(\mathbf v)\rangle|
\leq\|\mathbf y-\mathbf v\|_1\leq2.
\]
The auditor returns $w'$ if its empirical correlation
is at least $3\beta/4$, and returns $\perp$ otherwise.

With probability at least $1-\delta$, both the search
and validation guarantees hold.
On this event, every returned witness has true
correlation at least
$3\beta/4-\beta/4=\beta/2$.
If the auditor returns $\perp$, the candidate's true
correlation is strictly less than
$3\beta/4+\beta/4=\beta$.
By the search guarantee, this can occur only if $p$
satisfies $(\Omega,\varepsilon)$-Decision-OI.
Thus the auditor satisfies the definition above
with failure probability at most $\delta$.

\begin{lemma}\label{lm:boosting}
Suppose an $(\Omega,\varepsilon,\beta)$-auditor with
witness for Decision-OI is available at any prescribed
failure probability.
Fix $\delta\in(0,1/3)$ and initialize
\[
p^{(0)}(o\mid z,a)=\frac1{|\mathcal O|},
\qquad
\eta=\frac{\beta}{2},
\qquad
T_{\max}
=
\left\lceil\frac{8\log|\mathcal O|}{\beta^2}\right\rceil+1.
\]
The following algorithm returns an
$(\Omega,\varepsilon)$-Decision-OI predictor with
probability at least $1-\delta$, after at most
$\lceil8\log|\mathcal O|/\beta^2\rceil$ updates.

For each round $0\leq t<T_{\max}$:
\begin{enumerate}
    \item Call the auditor on $p^{(t)}$ with failure
    probability $\delta/T_{\max}$, using logged
    observations independent of all preceding calls.
    If it returns $\perp$, terminate and return
    $p^{(t)}$.

    \item Otherwise, let $w^{(t)}$ be the returned
    witness and define
    \[
    \mathbf v^{(t)}(z)
    =
    \frac1m
    \bigl(p^{(t)}(o\mid z,a)\bigr)_{(a,o)\in\mathcal A\times\mathcal O}.
    \]
    For every $(z,a,o)$, update
    \[
    p^{(t+1)}(o\mid z,a)
    =
    \frac{
    p^{(t)}(o\mid z,a)
    \exp\bigl(\eta w^{(t)}_{(a,o)}
    (\mathbf v^{(t)}(z))\bigr)
    }{
    \sum_{o'\in\mathcal O}
    p^{(t)}(o'\mid z,a)
    \exp\bigl(\eta w^{(t)}_{(a,o')}
    (\mathbf v^{(t)}(z))\bigr)
    }.
    \]
\end{enumerate}
If the iteration limit is reached, return the current
predictor.
On the stated success event, the auditor returns
$\perp$ within $T_{\max}$ calls.
\end{lemma}

\begin{proof}
By independence of the batches, the auditor guarantee
applies conditionally on the preceding history at
every round.
A union bound over at most $T_{\max}$ calls shows
that all auditor outputs satisfy their guarantees
simultaneously with probability at least $1-\delta$.
We condition on this event.
Returning $\perp$ then certifies
$(\Omega,\varepsilon)$-Decision-OI.

To bound the number of updates, define the potential
\[
\Phi_t
=
\mathbb E_z
\left[
\frac1m\sum_{a\in\mathcal A}
D_{\mathrm{KL}}
\bigl(
p^*(\cdot\mid z,a)
\,\|\,p^{(t)}(\cdot\mid z,a)
\bigr)
\right].
\]
Uniform initialization gives
$\Phi_0\leq\log|\mathcal O|$, while
$\Phi_t\geq0$ for every $t$.

Whenever the auditor returns a witness,
\[
\mathbb E_{\mathcal D^{\mathrm{cal}}}
[\langle\mathbf y-\mathbf v,w^{(t)}(\mathbf v)\rangle]
\geq\frac{\beta}{2},
\]
where $\mathcal D^{\mathrm{cal}}$ denotes the distribution
induced by $p^{(t)}$.

For fixed $(z,a)$, Hoeffding's lemma and
$w^{(t)}_{(a,o)}\in[-1,1]$ imply
\[
\log\left(
\sum_{o\in\mathcal O}
p^{(t)}(o\mid z,a)
e^{\eta w^{(t)}_{(a,o)}(\mathbf v^{(t)}(z))}
\right)
\leq
\eta\sum_{o\in\mathcal O}
p^{(t)}(o\mid z,a)
w^{(t)}_{(a,o)}(\mathbf v^{(t)}(z))
+\frac{\eta^2}{2}.
\]
Substituting the update into the potential gives
\begin{align*}
\Phi_{t+1}-\Phi_t
&\leq
\frac{\eta}{m}\mathbb E_z
\left[
\sum_{a\in\mathcal A}\sum_{o\in\mathcal O}
\bigl(p^{(t)}(o\mid z,a)-p^*(o\mid z,a)\bigr)
w^{(t)}_{(a,o)}(\mathbf v^{(t)}(z))
\right]
+\frac{\eta^2}{2}
\\
&=
-\eta\,
\mathbb E_{\mathcal D^{\mathrm{cal}}}
[\langle\mathbf y-\mathbf v,w^{(t)}(\mathbf v)\rangle]
+\frac{\eta^2}{2}
\\
&\leq
-\frac{\eta\beta}{2}+\frac{\eta^2}{2}
=
-\frac{\beta^2}{8}.
\end{align*}
The equality follows from uniform logging and
$\mathbf v^{(t)}_{(a,o)}(z)=p^{(t)}(o\mid z,a)/m$.

Since $\Phi_0\leq\log|\mathcal O|$ and the potential
is nonnegative, there can be at most
$8\log|\mathcal O|/\beta^2$ updates.
Consequently, the auditor must return $\perp$ within
$T_{\max}$ calls, and its guarantee establishes
Decision-OI for the returned predictor.
\end{proof}

\subsection{Efficient Auditing for Weighted Calibration}

Lemma~\ref{applm:reduction} expresses Decision-OI violations
as weighted-calibration violations.
We approximate the associated weight functions by
polynomials with controlled RKHS norms and apply a
kernel auditing algorithm to obtain an efficient
Decision-OI auditor.

\begin{definition}[Multinomial kernel,
\citep{gopalan2024computationally}]
For a nonnegative integer $D$, let
$T_D=[k]^0\cup[k]^1\cup\cdots\cup[k]^D$.
For $\mathbf v=(v_1,\ldots,v_k)\in\Delta_k$
and a tuple $t=(t_1,\ldots,t_j)\in[k]^j$, define
$v_t=v_{t_1}\cdots v_{t_j}$, with the empty product
equal to $1$ when $j=0$.
Define
$\psi:\Delta_k\to\mathbb R^{1+k+\cdots+k^D}$
by assigning the coordinate $v_t$ to each $t\in T_D$.
The degree-$D$ multinomial kernel is
\begin{equation}
\operatorname{ker}_D(\mathbf v,\mathbf v')
=
\sum_{j=0}^D\langle\mathbf v,\mathbf v'\rangle^j
=
\langle\psi(\mathbf v),\psi(\mathbf v')\rangle.
\end{equation}
We denote its reproducing kernel Hilbert space by
$\Gamma(D)$ and its radius-$r$ ball by
\[
B_{\Gamma(D)}(r)
=
\{w\in\Gamma(D):\|w\|_{\Gamma(D)}\leq r\}.
\]
\end{definition}

\begin{lemma}[Kernel auditing,
\citep{gopalan2024computationally}, Theorem~33]
\label{lm:kernelauditing}
For every $r>0$, nonnegative integer $D$,
$\varepsilon>0$, and $\delta\in(0,1/3)$,
there exists an algorithm using
\[
n=\mathcal O\left(
\frac{kr^2(D+1)\log(1/\delta)}{\varepsilon^2}
\right)
\]
i.i.d. samples from a distribution $\mathcal D$
on $\Delta_k\times\mathcal E_k$ and running in time
$\operatorname{poly}(n,k,D)$. The algorithm always returns a function
\[
w'\in
\left(
B_{\Gamma(D)}\left(\frac1{\sqrt{D+1}}\right)
\right)^k.
\]
If there exists
$w\in(B_{\Gamma(D)}(r))^k$ such that
\[
\left|
\mathbb E_{(\mathbf v,\mathbf y)\sim\mathcal D}
[\langle\mathbf y-\mathbf v,w(\mathbf v)\rangle]
\right|>\varepsilon,
\]
then, with probability at least $1-\delta$,
\[
\mathbb E_{(\mathbf v,\mathbf y)\sim\mathcal D}
[\langle\mathbf y-\mathbf v,w'(\mathbf v)\rangle]
\geq\frac{\varepsilon}{3r\sqrt{D+1}}.
\]
Moreover, $w'$ takes values in $[-1,1]^k$.
\end{lemma}

Recall that $m=|\mathcal A|$, $k=|\mathcal A\times\mathcal O|$, and $B=1+c\geq1$ bounds the total loss.
Throughout the following analysis, we assume
$\tau>0$ and $\varepsilon\in(0,1/e)$.

\begin{lemma}\label{lm:approximation}
For $i\in[m]$, let $\sigma_i:[0,B]^m\to[0,1]$
be the $i$-th component of the softmax function
with inverse temperature $\tau$:
\[
\sigma_i(t_1,\ldots,t_m)
=
\frac{\exp(-\tau t_i)}
{\sum_{j=1}^m\exp(-\tau t_j)}.
\]
There exists a constant $C_0(B,\tau)>0$, independent
of $m$ and $k$, and a polynomial $p$ of degree at most
$C_0(B,\tau)\log^2(1/\varepsilon)$ such that
\[
|p(t_1,\ldots,t_m)-\sigma_i(t_1,\ldots,t_m)|
\leq\frac{\varepsilon}{mB}
\qquad
\text{for all }(t_1,\ldots,t_m)\in[0,B]^m.
\]
\end{lemma}

\begin{proof}
The proof proceeds in three steps.
We first approximate the exponential and reciprocal
functions separately, then combine these approximations
to obtain a polynomial approximation of the softmax
component.

\textbf{Step 1. Approximation of the exponential
function on $[0,B]$.}

Set
\[
d_1=
\left\lceil
\max\left\{
1,\;2e\tau B,\;
\frac{\log(8B/\varepsilon)+2\tau B}{\log2}
\right\}
\right\rceil,
\qquad
p_1(t)=\sum_{j=0}^{d_1}\frac{(-\tau t)^j}{j!}.
\]
By Taylor's theorem, for every $t\in[0,B]$,
\[
|p_1(t)-e^{-\tau t}|
\leq
\frac{(\tau B)^{d_1+1}}{(d_1+1)!}
\leq
\left(\frac{e\tau B}{d_1+1}\right)^{d_1+1}
\leq
2^{-(d_1+1)}
\leq
\frac{\varepsilon}{8B}e^{-2\tau B}.
\]

\textbf{Step 2. Approximation of the reciprocal
function on $[e^{-\tau B}/2,5/4]$.}

Set
\[
d_2=
\left\lceil
\frac52e^{\tau B}
\left(\log\frac{16B}{\varepsilon}+2\tau B\right)
\right\rceil,
\qquad
p_2(t)=\frac45\sum_{j=0}^{d_2}
\left(1-\frac{4t}{5}\right)^j.
\]
For $t\in[e^{-\tau B}/2,5/4]$, the geometric-series
identity gives
\begin{align*}
\left|p_2(t)-\frac1t\right|
&=
\frac45\sum_{j=d_2+1}^{\infty}
\left(1-\frac{4t}{5}\right)^j
=
\frac1t\left(1-\frac{4t}{5}\right)^{d_2+1}
\\
&\leq
2e^{\tau B}
\exp\left(-\frac25e^{-\tau B}(d_2+1)\right)
\leq
\frac{\varepsilon e^{-\tau B}}{8B}.
\end{align*}

\textbf{Step 3. Approximation of the softmax
component $\sigma_i$.}

For $(t_1,\ldots,t_m)\in[0,B]^m$, define
\[
p_3(t_1,\ldots,t_m)
=
\frac1m p_1(t_i)
p_2\left(\frac1m\sum_{j=1}^m p_1(t_j)\right).
\]
Write $E_j=e^{-\tau t_j}$ for $j\in[m]$.
Then
\[
\sigma_i(t_1,\ldots,t_m)
=
\frac1m\frac{E_i}{\frac1m\sum_{j=1}^mE_j},
\qquad
\left|
\frac1m\sum_{j=1}^m p_1(t_j)
-
\frac1m\sum_{j=1}^m E_j
\right|
\leq\frac{\varepsilon}{8B}e^{-2\tau B}.
\]
Since $B\geq1$ and $\varepsilon<1/e$, Step 1 implies
\[
\frac12e^{-\tau B}
\leq p_1(t)
\leq1+\frac{\varepsilon}{8B}e^{-2\tau B}
<\frac54
\qquad
\text{for every }t\in[0,B].
\]
Thus the average of the $p_1(t_j)$ lies in the
approximation interval from Step 2.
Moreover,
$\frac1m\sum_{j=1}^mE_j\geq e^{-\tau B}$.
Consequently,
\begin{align*}
&\left|
p_2\left(\frac1m\sum_{j=1}^m p_1(t_j)\right)
-
\frac1{\frac1m\sum_{j=1}^mE_j}
\right|
\\
&\leq
\left|
p_2\left(\frac1m\sum_{j=1}^m p_1(t_j)\right)
-
\frac1{\frac1m\sum_{j=1}^m p_1(t_j)}
\right|
+
\left|
\frac1{\frac1m\sum_{j=1}^m p_1(t_j)}
-
\frac1{\frac1m\sum_{j=1}^m E_j}
\right|
\\
&\leq
\frac{\varepsilon e^{-\tau B}}{8B}
+
\frac{
\frac{\varepsilon}{8B}e^{-2\tau B}
}{
\left(\frac1m\sum_{j=1}^m p_1(t_j)\right)
\left(\frac1m\sum_{j=1}^m E_j\right)
}
\\& \leq\frac{3\varepsilon}{8B}.
\end{align*}
It follows that
\begin{align*}
&|p_3(t_1,\ldots,t_m)-\sigma_i(t_1,\ldots,t_m)|
\\
&\leq
\frac1m p_1(t_i)
\left|
p_2\left(\frac1m\sum_{j=1}^m p_1(t_j)\right)
-
\frac1{\frac1m\sum_{j=1}^m E_j}
\right|
+
\frac1m
\frac{|p_1(t_i)-E_i|}
{\frac1m\sum_{j=1}^mE_j}
\\
&\leq
\frac1m
\left(
\frac54\frac{3\varepsilon}{8B}
+
\frac{\varepsilon e^{-\tau B}}{8B}
\right)
\\&\leq\frac{\varepsilon}{mB}.
\end{align*}

The polynomial $p_3$ has degree at most $d_1(d_2+1)$.
Since $\log(1/\varepsilon)>1$,
\[
d_1
\leq
\left[
1+\max\left\{
1,\;2e\tau B,\;
\frac{1+\log(8B)+2\tau B}{\log2}
\right\}
\right]\log(1/\varepsilon),
\]
and
\[
d_2+1
\leq
\left[
2+\frac52e^{\tau B}
\bigl(1+\log(16B)+2\tau B\bigr)
\right]\log(1/\varepsilon).
\]
Define
\[
C_0(B,\tau)
=
\left[
1+\max\left\{
1,\;2e\tau B,\;
\frac{1+\log(8B)+2\tau B}{\log2}
\right\}
\right]
\left[
2+\frac52e^{\tau B}
\bigl(1+\log(16B)+2\tau B\bigr)
\right].
\]
Then
\[
\deg(p_3)
\leq d_1(d_2+1)
\leq C_0(B,\tau)\log^2(1/\varepsilon).
\]
The constant is independent of $m$ and $k$.
Taking $p=p_3$ proves the claim.
\end{proof}

For an $m$-variate polynomial $p$ and vectors
$\mathbf u_1,\ldots,\mathbf u_m\in\mathbb R^k$, define
\[
p_{\mathbf u_1,\ldots,\mathbf u_m}(\mathbf v)
=
p\bigl(
\langle\mathbf u_1,\mathbf v\rangle,\ldots,
\langle\mathbf u_m,\mathbf v\rangle
\bigr).
\]

\begin{lemma}\label{lm:bound2}
Let
\[
p(t_1,\ldots,t_m)
=
\sum_{j=0}^d\sum_{I\in[m]^j}
\beta_I t_{i_1}\cdots t_{i_j}
\]
be a representation of an $m$-variate polynomial
of total degree at most $d$, where
$I=(i_1,\ldots,i_j)$ and $|I|=j$.
The empty product represents the constant term.
For vectors
$\mathbf u_1,\ldots,\mathbf u_m\in[-mB,mB]^k$,
the function $p_{\mathbf u_1,\ldots,\mathbf u_m}$
belongs to $\Gamma(d)$ and satisfies
\[
\|p_{\mathbf u_1,\ldots,\mathbf u_m}\|_{\Gamma(d)}^2
\leq
(d+1)(mB\sqrt{k})^{2d}
\left(\sum_{|I|\leq d}|\beta_I|\right)^2.
\]
\end{lemma}

\begin{proof}
For $I=(i_1,\ldots,i_j)$, substituting
$t_r=\langle\mathbf u_r,\mathbf v\rangle$ gives
\[
\beta_I
\prod_{\ell=1}^j\langle\mathbf u_{i_\ell},\mathbf v\rangle
=
\beta_I\sum_{t\in[k]^j}
A_{t,I}v_{t_1}\cdots v_{t_j},
\qquad
A_{t,I}
=
\prod_{\ell=1}^j(\mathbf u_{i_\ell})_{t_\ell}.
\]
For $j=0$, take $A_{t,I}=1$.
Since $|A_{t,I}|\leq(mB)^j$,
\[
p_{\mathbf u_1,\ldots,\mathbf u_m}(\mathbf v)
=
\sum_{j=0}^d\sum_{t\in[k]^j}
\left(\sum_{|I|=j}\beta_I A_{t,I}\right)
v_{t_1}\cdots v_{t_j}.
\]
This is a representation in the multinomial feature
space.
Its coefficient norm therefore bounds the RKHS norm:
\begin{align*}
\|p_{\mathbf u_1,\ldots,\mathbf u_m}\|_{\Gamma(d)}^2
&\leq
\sum_{j=0}^d\sum_{t\in[k]^j}
\left|\sum_{|I|=j}\beta_I A_{t,I}\right|^2
\\
&\leq
\sum_{j=0}^d
(mB\sqrt{k})^{2j}
\left(\sum_{|I|=j}|\beta_I|\right)^2
\\
&\leq
(d+1)(mB\sqrt{k})^{2d}
\left(\sum_{|I|\leq d}|\beta_I|\right)^2,
\end{align*}
where the last inequality uses $mB\sqrt{k}\geq1$.
\end{proof}

\begin{remark}
A polynomial can have multiple representations in
the ordered-monomial form used in
Lemma~\ref{lm:bound2}, as noted in
\citep{goel2017reliably}.
The bound holds for each specified representation.
In the coefficient calculations below, we combine
identical monomials before taking absolute coefficient
sums.
\end{remark}

We now bound the coefficient sum for the polynomial
approximation of the softmax function.

\begin{lemma}\label{lm:bound3}
There exists a constant $C_1(B,\tau)>0$, independent
of $m$ and $k$, such that, for every
$\varepsilon\in(0,1/e)$ and
$\mathbf u_1,\ldots,\mathbf u_m\in[0,mB]^k$,
there exists a polynomial $p(t_1,\ldots,t_m)$
satisfying the following conditions:
\begin{enumerate}
    \item For every $\mathbf v\in\Delta_k$ such that
    $\langle\mathbf u_i,\mathbf v\rangle\in[0,B]$
    for all $i\in[m]$,
    \[
    \left|
    \sigma_1\bigl(
    \langle\mathbf u_1,\mathbf v\rangle,\ldots,
    \langle\mathbf u_m,\mathbf v\rangle
    \bigr)
    -
    p_{\mathbf u_1,\ldots,\mathbf u_m}(\mathbf v)
    \right|
    \leq\frac{\varepsilon}{2mB}.
    \]

    \item
 It has degree  $\deg(p)\leq
    D:=\lceil C_1(B,\tau)\log^2(4/\varepsilon)\rceil$.

    \item The polynomial $p_{\mathbf u_1,\ldots,\mathbf u_m}$ induced by $p$ has RKHS norm 
    \[
    \|p_{\mathbf u_1,\ldots,\mathbf u_m}\|_{\Gamma(D)}^2
    \leq
    (D+1)\exp(D)(mB\sqrt{k})^{2D}.
    \]

    \item
    $|p(t_1,\ldots,t_m)|\leq1$
    for every $(t_1,\ldots,t_m)\in[0,B]^m$.
\end{enumerate}
\end{lemma}

\begin{proof}
Let $p_1$ and $p_2$ be the polynomials constructed
in Lemma~\ref{lm:approximation} with accuracy
parameter $\varepsilon/4$.
Define
\[
p(t_1,\ldots,t_m)
=
\frac{4B}{4mB+\varepsilon}
p_1(t_1)
p_2\left(\frac1m\sum_{j=1}^m p_1(t_j)\right),
\]
and
\[
p_0(t_1,\ldots,t_m)
=
\frac1m p_1(t_1)
p_2\left(\frac1m\sum_{j=1}^m p_1(t_j)\right).
\]
By Lemma~\ref{lm:approximation},
\[
|p_0(t_1,\ldots,t_m)-\sigma_1(t_1,\ldots,t_m)|
\leq\frac{\varepsilon}{4mB}
\]
on $[0,B]^m$.
Since $p=\frac{4mB}{4mB+\varepsilon}p_0$,
\begin{align*}
|p(t_1,\ldots,t_m)-\sigma_1(t_1,\ldots,t_m)|
&\leq
\frac{4mB}{4mB+\varepsilon}
|p_0(t_1,\ldots,t_m)-\sigma_1(t_1,\ldots,t_m)|
\\
&\quad+
\frac{\varepsilon}{4mB+\varepsilon}
|\sigma_1(t_1,\ldots,t_m)|
\\
&\leq
\frac{2\varepsilon}{4mB+\varepsilon}
\leq\frac{\varepsilon}{2mB}.
\end{align*}
This proves condition 1.
Writing $d_1=\deg(p_1)$ and $d_2=\deg(p_2)$,
the same construction gives
\[
\deg(p)
\leq d_1(d_2+1)
\leq C_0(B,\tau)\log^2(4/\varepsilon).
\]
Thus condition 2 holds whenever $C_1(B,\tau)\geq C_0(B,\tau)$.

To prove condition 3, write the standard monomial
expansion of a polynomial $P$ as
\[
P(t_1,\ldots,t_m)
=
\sum_{\substack{I\in\mathbb Z_{\geq0}^m\\|I|\leq d}}
\beta_I t_1^{i_1}\cdots t_m^{i_m},
\qquad
|I|=i_1+\cdots+i_m,
\]
and define
\[
A(P)=\sum_{|I|\leq d}|\beta_I|.
\]
We use the following elementary properties:

\textbf{(a)}
For polynomials $P$ and $Q$,
\[
A(P+Q)\leq A(P)+A(Q),
\qquad
A(PQ)\leq A(P)A(Q).
\]

\textbf{(b)}
If $P$ is univariate of degree $d_P$, then
\[
A(P\circ Q)
\leq A(P)\max\{A(Q),1\}^{d_P}.
\]

By the definition of $p_1$,
\[
A(p_1)=\sum_{j=0}^{d_1}\frac{\tau^j}{j!}\leq e^\tau.
\]
For $p_2$,
\[
p_2(t)
=
\frac45\sum_{j=0}^{d_2}
\sum_{s=0}^j\binom js
\left(-\frac45\right)^s t^s
=
\frac45\sum_{s=0}^{d_2}
\left(-\frac45\right)^s
\left(\sum_{j=s}^{d_2}\binom js\right)t^s.
\]
The coefficient of $t^s$ has sign $(-1)^s$, so
\[
A(p_2)
=
p_2(-1)
=
\left(\frac95\right)^{d_2+1}-1
\leq e^{d_2+1}.
\]
Moreover,
\[
A\left(\frac1m\sum_{j=1}^m p_1(t_j)\right)
\leq A(p_1)\leq e^\tau.
\]
Since $\frac{4B}{4mB+\varepsilon}\leq1$,
properties (a) and (b) imply
\begin{align*}
A(p)
&\leq
A(p_1)A(p_2)
\max\left\{
A\left(\frac1m\sum_{j=1}^m p_1(t_j)\right),1
\right\}^{d_2}
\\
&\leq e^\tau e^{d_2+1}e^{\tau d_2}
=
\exp\bigl((\tau+1)(d_2+1)\bigr).
\end{align*}
The construction in Lemma~\ref{lm:approximation}
ensures $d_1\geq\max\{1,\tau\}$.
Therefore,
\[
A(p)
\leq
\exp\bigl(2d_1(d_2+1)\bigr)
\leq
\exp\left(2C_0(B,\tau)\log^2(4/\varepsilon)\right).
\]
Choose $C_1(B,\tau)=4C_0(B,\tau)$.
Then
\[
D=\left\lceil C_1(B,\tau)\log^2(4/\varepsilon)\right\rceil,
\qquad
\deg(p)\leq D,
\qquad
A(p)^2\leq e^D.
\]
The standard monomial expansion gives a representation
as in Lemma~\ref{lm:bound2} with coefficient sum $A(p)$.
Applying that lemma with degree bound $D$ yields
\[
\|p_{\mathbf u_1,\ldots,\mathbf u_m}\|_{\Gamma(D)}^2
\leq
(D+1)(mB\sqrt{k})^{2D}A(p)^2
\leq
(D+1)(mB\sqrt{k})^{2D}\exp(D).
\]
This proves condition 3.

Finally, for every $(t_1,\ldots,t_m)\in[0,B]^m$,
\begin{align*}
|p(t_1,\ldots,t_m)|
&=
\frac{4mB}{4mB+\varepsilon}|p_0(t_1,\ldots,t_m)|
\\
&\leq
\frac{4mB}{4mB+\varepsilon}
\left(
|p_0(t_1,\ldots,t_m)-\sigma_1(t_1,\ldots,t_m)|
+
|\sigma_1(t_1,\ldots,t_m)|
\right)
\\
&\leq
\frac{4mB}{4mB+\varepsilon}
\left(1+\frac{\varepsilon}{4mB}\right)
=1.
\end{align*}
This proves condition 4.
\end{proof}

We now state our main theorem.

\begin{theorem}\label{thm:efficiency}
Under the uniform-logging and fixed-cost assumptions
of this section, fix $\tau>0$.
There exists a constant $C(B,\tau)>0$, depending only
on $B,\tau$ and independent of $|\mathcal A|,|\mathcal O|$,
such that the following holds.
For every $\varepsilon\in(0,1/e)$ and
$\delta\in(0,1/3)$, let
\[
D=\left\lceil C(B,\tau)\log^2(4/\varepsilon)\right\rceil,
\qquad
r=\sqrt{D+1}(mB\sqrt{ek})^D,
\]
where $m=|\mathcal A|$ and $k=m|\mathcal O|$.
There exists an
\[
\left(
\Omega,\varepsilon,
\frac{\varepsilon}{6mBr\sqrt{D+1}}
\right)
\]
auditor with witness for Decision-OI, with failure
probability at most $\delta$, sample complexity
\[
n=\mathcal O\left(
\frac{km^2B^2r^2(D+1)\log(1/\delta)}{\varepsilon^2}
\right),
\]
and running time $\operatorname{poly}(n,k,D)$.
The samples used in an auditor call are independent
of its input predictor.
\end{theorem}

\begin{remark}
Let $D$ and $r$ be as in Theorem~\ref{thm:efficiency}.
With uniform initialization
$p^{(0)}(o\mid z,a)=1/|\mathcal O|$,
multiplicative updates with factor
$\exp(\eta w^{(t)}_{(a,o)}(\mathbf v^{(t)}(z)))$,
and termination when the auditor returns $\perp$,
Lemma~\ref{lm:boosting} gives
\[
\mathcal O\left(
\frac{m^2B^2r^2(D+1)\log|\mathcal O|}{\varepsilon^2}
\right)
\]
updates on the joint success event of the auditor calls.
Each call uses independent samples, with failure
probabilities allocated across calls as in that lemma.

For fixed $B,\tau,\varepsilon$, this update bound is
polynomial in $m$ and $k$.
For fixed $m,k,B,\tau$, it is quasi-polynomial
in $1/\varepsilon$.
Combining the auditor with boosting therefore
preserves these qualitative efficiency guarantees.
\end{remark}

\begin{remark}
The updates are normalized separately for each action,
preserving total mass $1/m$ in each action block of
$\mathbf v$.
The averaged KL potential in Lemma~\ref{lm:boosting}
accounts for this block structure and yields the
stated iteration bound.
\end{remark}

\begin{proof}[Proof of Theorem~\ref{thm:efficiency}]
For simplicity, write $\mathcal A=\{a_1,\ldots,a_m\}$.
For a loss-cost pair $\omega=(\widetilde l,c)$, define
\[
s_a(z)
=
m\sum_{o\in\mathcal O}
\ell_a^\omega(o)\mathbf v_{(a,o)}(z),
\]
and let $\mathbf l_a^\omega\in[0,mB]^k$ have coordinates
\[
(\mathbf l_a^\omega)_{(a',o')}
=
m\ell_a^\omega(o')\mathbf 1\{a=a'\}.
\]
Since $\mathbf v_{(a,o)}(z)=p(o\mid z,a)/m$ and
$0\leq\ell_a^\omega(o)\leq B$,
\[
s_a(z)
=
\langle\mathbf l_a^\omega,\mathbf v(z)\rangle
=
\sum_{o\in\mathcal O}\ell_a^\omega(o)p(o\mid z,a)
\in[0,B].
\]
Consequently,
\[
\pi^\omega(a_i\mid z)
=
\sigma_i\bigl(
\langle\mathbf l_{a_1}^\omega,\mathbf v(z)\rangle,
\ldots,
\langle\mathbf l_{a_m}^\omega,\mathbf v(z)\rangle
\bigr).
\]

Take $C(B,\tau)=C_1(B,\tau)$ from
Lemma~\ref{lm:bound3} and set
\[
D=\left\lceil C(B,\tau)\log^2(4/\varepsilon)\right\rceil.
\]
If $p$ violates $(\Omega,\varepsilon)$-Decision-OI,
there exists $\omega\in\Omega$ such that
\[
\left|
\mathbb E_z\left[
\frac1m
\sum_{a\in\mathcal A}\sum_{o\in\mathcal O}
\frac{\ell_a^\omega(o)}B\pi^\omega(a\mid z)
\bigl(p^*(o\mid z,a)-p(o\mid z,a)\bigr)
\right]
\right|
>
\frac{\varepsilon}{mB}.
\]

By Lemma~\ref{lm:bound3}, with a permutation of
coordinates for each softmax component, there exists
an $m$-variate polynomial $q^{(a_i)}$ of degree at most
$D$ such that
\[
\left|
q^{(a_i)}_{\mathbf l_{a_1}^\omega,\ldots,\mathbf l_{a_m}^\omega}
(\mathbf v(z))
-
\pi^\omega(a_i\mid z)
\right|
\leq\frac{\varepsilon}{2mB},
\]
and
\[
\left\|
q^{(a_i)}_{\mathbf l_{a_1}^\omega,\ldots,\mathbf l_{a_m}^\omega}
\right\|_{\Gamma(D)}^2
\leq
(D+1)(mB\sqrt{k})^{2D}\exp(D).
\]
Here we used
$\langle\mathbf l_{a_i}^\omega,\mathbf v(z)\rangle\in[0,B]$
for every action $a_i$.
Define
\[
r
=
\sqrt{D+1}(mB\sqrt{k})^D\exp(D/2)
=
\sqrt{D+1}(mB\sqrt{ek})^D.
\]
Since $0\leq\ell_a^\omega(o)/B\leq1$, the tuple
\[
\left(
\frac{\ell_a^\omega(o)}B
q^{(a)}_{\mathbf l_{a_1}^\omega,\ldots,\mathbf l_{a_m}^\omega}
\right)_{(a,o)\in\mathcal A\times\mathcal O}
\]
belongs to $(B_{\Gamma(D)}(r))^k$.

We next bound the change in correlation caused by
the polynomial approximation.
For every $z$ and $a$, the normalized loss takes
values in $[0,1]$, so
\[
\left|
\sum_{o\in\mathcal O}
\frac{\ell_a^\omega(o)}B
\bigl(p^*(o\mid z,a)-p(o\mid z,a)\bigr)
\right|
\leq1.
\]
Grouping the outcome terms by action gives
\begin{align*}
&\left|
\mathbb E_z\left[
\frac1m\sum_{a\in\mathcal A}
\left(
q^{(a)}_{\mathbf l_{a_1}^\omega,\ldots,\mathbf l_{a_m}^\omega}
(\mathbf v(z))
-\pi^\omega(a\mid z)
\right)
\sum_{o\in\mathcal O}
\frac{\ell_a^\omega(o)}B
\bigl(p^*(o\mid z,a)-p(o\mid z,a)\bigr)
\right]
\right|
\\
&\leq
\frac1m\sum_{a\in\mathcal A}
\mathbb E_z\left[
\left|
q^{(a)}_{\mathbf l_{a_1}^\omega,\ldots,\mathbf l_{a_m}^\omega}
(\mathbf v(z))
-\pi^\omega(a\mid z)
\right|
\right]
\leq\frac{\varepsilon}{2mB}.
\end{align*}
Therefore,
\[
\left|
\mathbb E_z\left[
\frac1m
\sum_{a\in\mathcal A}\sum_{o\in\mathcal O}
\frac{\ell_a^\omega(o)}B
q^{(a)}_{\mathbf l_{a_1}^\omega,\ldots,\mathbf l_{a_m}^\omega}
(\mathbf v(z))
\bigl(p^*(o\mid z,a)-p(o\mid z,a)\bigr)
\right]
\right|
>
\frac{\varepsilon}{2mB}.
\]
Under uniform logging,
\[
\mathbb E[\mathbf y_{(a,o)}\mid z]
=
\frac1m p^*(o\mid z,a).
\]
Thus $(B_{\Gamma(D)}(r))^k$ contains a function
whose absolute correlation with $\mathbf y-\mathbf v$
under $\mathcal D^{\mathrm{cal}}$ exceeds
$\varepsilon/(2mB)$.

Write
\[
\beta=\frac{\varepsilon}{6mBr\sqrt{D+1}}.
\]
Applying Lemma~\ref{lm:kernelauditing} with accuracy
parameter $\varepsilon/(2mB)$ and failure probability
$\delta/2$ uses
\[
\mathcal O\left(
\frac{km^2B^2r^2(D+1)\log(2/\delta)}{\varepsilon^2}
\right)
\]
samples and produces a function
\[
w'\in
\left(
B_{\Gamma(D)}
\left(\frac1{\sqrt{D+1}}\right)
\right)^k.
\]
Whenever $p$ violates Decision-OI, with probability
at least $1-\delta/2$ this function satisfies
\[
\mathbb E_{\mathcal D^{\mathrm{cal}}}
[\langle\mathbf y-\mathbf v,w'(\mathbf v)\rangle]
\geq\beta.
\]
The kernel algorithm runs in time
$\operatorname{poly}(n,k,D)$, and its output always
takes values in $[-1,1]^k$.

On an independent batch of
$\mathcal O(\beta^{-2}\log(1/\delta))$ samples,
the auditor estimates this correlation to accuracy
$\beta/4$ with probability at least $1-\delta/2$.
It returns $w'$ if the empirical correlation is at
least $3\beta/4$, and returns $\perp$ otherwise.
On the joint success event, every returned witness
has true correlation at least $\beta/2$.
If the auditor returns $\perp$, the candidate's true
correlation is strictly less than $\beta$, which
implies that $p$ satisfies
$(\Omega,\varepsilon)$-Decision-OI.
The failure probability is therefore at most $\delta$.

Since
\[
\beta^{-2}
=
\frac{36m^2B^2r^2(D+1)}{\varepsilon^2},
\]
the total sample complexity is
\[
n=\mathcal O\left(
\frac{km^2B^2r^2(D+1)\log(1/\delta)}{\varepsilon^2}
\right).
\]
The kernel representation permits evaluation of $w'$
on the independent batch in time polynomial in
$n,k,D$, so the total running time remains
$\operatorname{poly}(n,k,D)$.
Each logged observation produces a sample from
$\mathcal D^{\mathrm{cal}}$ through Eq.~\eqref{eq:Dcal}.
By the definition of an auditor with witness for
Decision-OI, this establishes the claimed
$(\Omega,\varepsilon,\beta)$ guarantee.

For fixed $B,\tau,\varepsilon$, these complexities
are polynomial in $m=|\mathcal A|$ and
$k=|\mathcal A\times\mathcal O|$.
For fixed $B,\tau,m,k$, they are quasi-polynomial
in $1/\varepsilon$.
\end{proof}
\section{Theoretical Guarantee for Extensions in Section~\ref{sec:ext}}\label{app:ext}

\subsection{Extension to Feature-Based Collaboration Policies}
While Theorems~\ref{thm:upper} and \ref{thm:convergence} establish convergence and sample efficiency of COI-Boost, real-world human--AI collaboration systems typically process complex, high-dimensional states, such as medical images or lengthy textual contexts. To study the representation required by COI-Boost, we consider predefined collaboration policies operating on a fixed, lower-dimensional feature representation, such as the confidence scores of a black-box model~\citep{alur2024human,casacuberta2025selective}. We show that, when the costs and logging propensities also depend only on these features, the learned surrogate depends on the state through the same representation.

\begin{lemma}[Feature-Dependent Reduction of COI-Boost]
Assume that every policy in $\mathcal{H}$, every cost function in $\mathcal{C}$, and the logging propensities $\pi_{\mathrm{log}}(a\mid z)>0$ depend on the state $z$ only through a fixed feature map
$\psi(z)=[\psi_1(z),\ldots,\psi_k(z)]^\top\in\mathbb{R}^k$.
Suppose Algorithm~\ref{alg:coiboosting} uses the auditor constructed in Theorem~\ref{thm:upper}, which searches over $\Pi_{\mathrm{eval}}$.
Then, at every iteration $t$, the base predictive distribution
$p^{(t)}(o\mid z,a)$ depends on $z$ only through $\psi(z)$.
\end{lemma}




\begin{proof}
We proceed by mathematical induction. At $t=0$, the algorithm initializes
$p^{(0)}(o\mid z,a)=1/|\mathcal{O}|$, which does not depend on $z$.
Assume that at iteration $t$, the predictor
$p^{(t)}(o\mid z,a)$ depends on $z$ only through $\psi(z)$.

During the auditing step, the algorithm searches over
$\Pi_{\mathrm{eval}}$ for a violating policy and loss-cost pair.
By assumption, every base collaboration policy $h\in\mathcal{H}$
depends on $z$ only through $\psi(z)$.
Moreover, for $\omega=(\widetilde l,c)$ and $a=(r,s,g)$,
the total loss
$\ell^\omega_a(z,o)=\widetilde l(o)+r c(x)$
depends on $z$ only through $\psi(z)$.

The post-processing policy $\pi^\omega$ is defined by applying
softmax to the predicted action losses:
\[
\pi^\omega(a\mid z)
=
\frac{\exp\!\left(-\tau
\mathbb{E}_{o\sim p^{(t)}(\cdot\mid z,a)}
[\ell^\omega_a(z,o)]\right)}
{\sum_{a'\in\mathcal{A}}
\exp\!\left(-\tau
\mathbb{E}_{o\sim p^{(t)}(\cdot\mid z,a')}
[\ell^\omega_{a'}(z,o)]\right)}.
\]
By the induction hypothesis and the assumption on the costs,
both $p^{(t)}$ and $\ell^\omega_a(z,o)$ depend on $z$ only
through $\psi(z)$. Therefore, the predicted action losses
and $\pi^\omega(a\mid z)$ have the same feature dependence.

Consequently, every policy in $\Pi_{\mathrm{eval}}$ depends on
$z$ only through $\psi(z)$.
If the auditor returns $\perp$, the algorithm terminates.
Otherwise, let $(\pi^{(t)},\omega^{(t)},\sigma)$ be its returned
certificate. The update score is
\[
g^{(t)}(z,a,o)
=
\sigma
\frac{\pi^{(t)}(a\mid z)}{\pi_{\mathrm{log}}(a\mid z)}
\ell^{\omega^{(t)}}_a(z,o).
\]
By assumption, the logging propensities also depend on $z$
only through $\psi(z)$.
Thus the policy, logging propensity, and total loss all have
the same feature dependence, while $\sigma\in\{-1,1\}$
is fixed for this update.
It follows that $g^{(t)}$ depends on $z$ only through $\psi(z)$.

The multiplicative weights update is
\[
p^{(t+1)}(o\mid z,a)
=
\frac{p^{(t)}(o\mid z,a)
\exp\!\left(\eta g^{(t)}(z,a,o)\right)}
{\sum_{o'\in\mathcal{O}}p^{(t)}(o'\mid z,a)
\exp\!\left(\eta g^{(t)}(z,a,o')\right)}.
\]
Both the numerator and the normalization denominator depend on
$z$ only through $\psi(z)$.
Hence $p^{(t+1)}(o\mid z,a)$ has the same feature dependence,
completing the induction.
\end{proof}

We now consider randomized collaboration policies that
operate on a fixed feature representation $\psi(z)$.
Each policy forms linear action scores from these features
and converts them into action probabilities through a
fixed Lipschitz map. Under bounds on the features and
policy parameters, we use Rademacher complexity to
establish a sample-complexity guarantee for COI auditing
over the resulting policy class $\mathcal{H}_{\psi}$.

\begin{assumption}[Lipschitz Policies over Bounded Features]
\label{as:kernel}
Let $\psi:\mathcal{Z}\to\mathbb{R}^k$ be a fixed feature map
satisfying
$\sup_{z\in\mathcal{Z}}\|\psi(z)\|_2\leq B$.
Let $\phi:\mathbb{R}^{|\mathcal{A}|}\to\Delta(\mathcal{A})$
be a fixed map from action scores to action probabilities
satisfying
\[
\|\phi(u)-\phi(v)\|_2
\leq \Lambda\|u-v\|_2
\qquad
\text{for all }u,v\in\mathbb{R}^{|\mathcal{A}|}.
\]
Fix $W_{\max}>0$ and define the randomized policy family
\[
\mathcal{H}_{\psi}
=
\left\{
h(\cdot\mid z)=\phi(W\psi(z)):
W\in\mathbb{R}^{|\mathcal{A}|\times k},
\ \|W\|_F\leq W_{\max}
\right\},
\]
where
\[
\|W\|_F^2
=
\sum_{a\in\mathcal{A}}\sum_{j=1}^{k}W_{a,j}^2.
\]
The feature and Lipschitz bounds use Euclidean norms,
and the parameter constraint uses the Frobenius norm.
\end{assumption}
\begin{theorem}[Sample Complexity for Feature-Based Policies]
\label{thm:smooth_sample_complexity}
Suppose Assumptions~\ref{as:log}, \ref{as:kernel}, and
\ref{as:complexity}(a,b) hold, and the logging propensities
are known. Fix a finite inverse temperature $\tau>0$ and
a predictor $p^{(t)}$ independently of the audit batch.
For each $\omega\in\Omega$, let
$\pi^\omega=k^\omega(p^{(t)})$, and define
\[
\Pi_{\mathrm{eval}}(p^{(t)})
=
\bigl(\mathcal{H}_{\psi}\times\Omega\bigr)
\cup
\left\{(\pi^\omega,\omega):\omega\in\Omega\right\}.
\]
For $\varepsilon,\delta\in(0,1)$, draw $m$ independent
samples from the logging distribution. Whenever
\[
m\gtrsim
\frac{(1+B_c)^2}{\gamma^2\varepsilon^2}
\left[
|\mathcal{A}|\Lambda^2B^2W_{\max}^2
+(1+\tau)^2|\mathcal{O}|
+\log\frac{2|\mathcal{C}|}{\delta}
\right],
\]
where the implicit constant is universal, with probability
at least $1-\delta$,
\[
\sup_{(\pi,\omega)\in\Pi_{\mathrm{eval}}(p^{(t)})}
\left|
\widehat G_m(\pi;\omega;p^{(t)})
-G(\pi;\omega;p^{(t)})
\right|
\leq \frac{\varepsilon}{8}.
\]

Consequently, consider an auditor that maximizes the absolute
empirical COI gap over $\Pi_{\mathrm{eval}}(p^{(t)})$ to
additive accuracy $\varepsilon/8$. If the selected pair has
absolute empirical gap greater than $3\varepsilon/4$, the
auditor returns this pair together with the sign of its
empirical gap; otherwise, it returns $\perp$.
On the same event, the auditor satisfies:
\begin{enumerate}
    \item Every returned certificate $(\pi,\omega,\sigma)$
    satisfies
    \[
    \sigma\left(
    V(\pi;\omega;p^*)-V(\pi;\omega;p^{(t)})
    \right)>\varepsilon/2.
    \]
    \item If the auditor returns $\perp$, then $p^{(t)}$
    satisfies $(\Omega,\mathcal{H}_{\psi},\varepsilon)$-COI.
\end{enumerate}
\end{theorem}
\begin{remark}
The sample bound has no explicit dependence on the feature
dimension $k$ when the feature norm bound $B$, the Frobenius
norm bound $W_{\max}$, and the Lipschitz constant $\Lambda$
are fixed. These quantities may themselves depend on the
chosen feature representation, so the dimension dependence
is controlled through the stated norm and smoothness bounds.

The auditing guarantee assumes that the absolute empirical
COI gap can be maximized to additive accuracy $\varepsilon/8$.
The theorem establishes the sample complexity and validity
of the resulting auditor; its computational efficiency
additionally depends on the procedure used to solve this
empirical optimization problem.
\end{remark}


\begin{proof}
The auditor's objective is to estimate the difference in expected
loss between the true environment and the base predictive
distribution uniformly over
$(\pi,\omega)\in\Pi_{\mathrm{eval}}(p^{(t)})$.
Recall that the population and empirical COI gaps are
\[
G(\pi;\omega;p^{(t)})
=
V(\pi;\omega;p^*)-V(\pi;\omega;p^{(t)})
\]
and
\[
\widehat G_m(\pi;\omega;p^{(t)})
=
\frac{1}{m}\sum_{i=1}^{m}
\left[
\frac{\pi(a_i\mid z_i)}{\pi_{\mathrm{log}}(a_i\mid z_i)}
\ell^\omega_{a_i}(z_i,o_i)
-
\sum_{a\in\mathcal A}\pi(a\mid z_i)
\mathbb E_{o\sim p^{(t)}(\cdot\mid z_i,a)}
[\ell^\omega_a(z_i,o)]
\right].
\]
Since $p^{(t)}$ is independent of the audit batch, we may
condition on it throughout the proof. For each fixed
$(\pi,\omega)$, the empirical gap is an unbiased estimator
of the population gap.

By Assumption~\ref{as:complexity}(a,b),
$|\ell^\omega_a(z,o)|\leq 1+B_c$.
Set $M=(1+B_c)/\gamma$. Each scalar contribution to
$\widehat G_m$ has absolute value at most
\[
\frac{1+B_c}{\gamma}+(1+B_c)\leq 2M.
\]
We decompose the evaluation class as
$\Pi_{\mathrm{eval}}(p^{(t)})
=\Pi_{\mathcal H}\cup\Pi_{\mathrm{soft}}$, where
\[
\Pi_{\mathcal H}
=
\{(h,\omega):h\in\mathcal H_\psi,\ \omega\in\Omega\},
\qquad
\Pi_{\mathrm{soft}}
=
\{(k^\omega(p^{(t)}),\omega):\omega\in\Omega\}.
\]

\textbf{Step 1: Rademacher Complexity of $\Pi_{\mathcal H}$.}

Fix $\omega$ and the audit sample.
For $h(\cdot\mid z)=\phi(W\psi(z))$, the $i$-th term
in $\widehat G_m(h;\omega;p^{(t)})$ is
\[
\frac{\phi(W\psi(z_i))_{a_i}}
{\pi_{\mathrm{log}}(a_i\mid z_i)}
\ell^\omega_{a_i}(z_i,o_i)
-
\sum_{a\in\mathcal A}
\phi(W\psi(z_i))_a
\mathbb E_{o\sim p^{(t)}(\cdot\mid z_i,a)}
[\ell^\omega_a(z_i,o)].
\]
For two parameter matrices $W$ and $W'$, we have
\[
\begin{aligned}
&\Bigg|
\frac{\phi(W\psi(z_i))_{a_i}-\phi(W'\psi(z_i))_{a_i}}
{\pi_{\mathrm{log}}(a_i\mid z_i)}
\ell^\omega_{a_i}(z_i,o_i)
-
\sum_{a\in\mathcal A}
\bigl(\phi(W\psi(z_i))_a-\phi(W'\psi(z_i))_a\bigr)
\mathbb E_{o\sim p^{(t)}(\cdot\mid z_i,a)}
[\ell^\omega_a(z_i,o)]
\Bigg|\\
&\leq
\frac{1+B_c}{\pi_{\mathrm{log}}(a_i\mid z_i)}
\left|\phi(W\psi(z_i))_{a_i}-\phi(W'\psi(z_i))_{a_i}\right|
+
(1+B_c)\sum_{a\in\mathcal A}
\left|\phi(W\psi(z_i))_a-\phi(W'\psi(z_i))_a\right|\\
&\leq
(1+B_c)
\left(
\frac{1}{\pi_{\mathrm{log}}(a_i\mid z_i)}
+\sqrt{|\mathcal A|}
\right)
\left\|\phi(W\psi(z_i))-\phi(W'\psi(z_i))\right\|_2\\
&\leq
\Lambda(1+B_c)
\left(
\frac{1}{\pi_{\mathrm{log}}(a_i\mid z_i)}
+\sqrt{|\mathcal A|}
\right)
\left\|(W-W')\psi(z_i)\right\|_2\\
&\leq
2\Lambda M\left\|(W-W')\psi(z_i)\right\|_2.
\end{aligned}
\]
The first inequality uses the loss bound.
The second follows from Cauchy--Schwarz, and the third
uses the Euclidean Lipschitz assumption on $\phi$.
The last inequality follows from
$\pi_{\mathrm{log}}(a_i\mid z_i)\geq\gamma$
and $|\mathcal A|\gamma\leq 1$.
Thus each scalar contribution to the IPS estimator is
$2\Lambda M$-Lipschitz in the action-score vector
$W\psi(z_i)$ under the Euclidean norm.

Let $\sigma_i$ and $\sigma_{i,a}$ be independent
Rademacher signs, each taking values $+1$ and $-1$
with equal probability.
The empirical Rademacher complexity under consideration is
\[
\mathbb E_\sigma
\sup_{\|W\|_F\leq W_{\max}}
\frac{1}{m}\sum_{i=1}^{m}\sigma_i
\left[
\frac{\phi(W\psi(z_i))_{a_i}}
{\pi_{\mathrm{log}}(a_i\mid z_i)}
\ell^\omega_{a_i}(z_i,o_i)
-
\sum_{a\in\mathcal A}\phi(W\psi(z_i))_a
\mathbb E_{o\sim p^{(t)}(\cdot\mid z_i,a)}
[\ell^\omega_a(z_i,o)]
\right].
\]
Since each expression in brackets is $2\Lambda M$-Lipschitz
in the score vector $W\psi(z_i)$, the vector-contraction
inequality bounds this quantity by
\[
\frac{2\sqrt{2}\Lambda M}{m}
\mathbb E_\sigma
\sup_{\|W\|_F\leq W_{\max}}
\sum_{i=1}^{m}\sum_{a\in\mathcal A}
\sigma_{i,a}(W\psi(z_i))_a.
\]

We now bound the remaining linear expression.
Expanding the matrix-vector product gives
\[
\sum_{i=1}^{m}\sum_{a\in\mathcal A}
\sigma_{i,a}(W\psi(z_i))_a
=
\sum_{a\in\mathcal A}\sum_{j=1}^{k}
W_{a,j}
\left(\sum_{i=1}^{m}\sigma_{i,a}\psi_j(z_i)\right).
\]
By Cauchy--Schwarz and $\|W\|_F\leq W_{\max}$,
\[
\sup_{\|W\|_F\leq W_{\max}}
\sum_{i=1}^{m}\sum_{a\in\mathcal A}
\sigma_{i,a}(W\psi(z_i))_a
\leq
W_{\max}
\left(
\sum_{a\in\mathcal A}
\left\|\sum_{i=1}^{m}\sigma_{i,a}\psi(z_i)\right\|_2^2
\right)^{1/2}.
\]
Taking expectations and applying Jensen's inequality yields
\[
\mathbb E_\sigma
\sup_{\|W\|_F\leq W_{\max}}
\sum_{i=1}^{m}\sum_{a\in\mathcal A}
\sigma_{i,a}(W\psi(z_i))_a
\leq
W_{\max}
\left(
\sum_{a\in\mathcal A}
\mathbb E_\sigma
\left\|\sum_{i=1}^{m}\sigma_{i,a}\psi(z_i)\right\|_2^2
\right)^{1/2}.
\]
Independence and zero means of the signs eliminate the
cross terms, so
\[
\sum_{a\in\mathcal A}
\mathbb E_\sigma
\left\|\sum_{i=1}^{m}\sigma_{i,a}\psi(z_i)\right\|_2^2
=
|\mathcal A|\sum_{i=1}^{m}\|\psi(z_i)\|_2^2
\leq |\mathcal A|mB^2.
\]
Consequently, the empirical Rademacher complexity of the
scalar IPS contributions, with $\omega$ fixed, is at most
\[
\frac{2\sqrt{2}\Lambda M B W_{\max}\sqrt{|\mathcal A|}}
{\sqrt m}.
\]

For fixed $h$ and $c$, the estimation error
$\widehat G_m-G$ is affine in the loss vector
$\widetilde l\in[0,1]^{|\mathcal O|}$.
Its maximum absolute value is therefore attained at a
vertex of this cube. Applying standard Rademacher
generalization bounds uniformly over $h\in\mathcal H_\psi$
and taking a union bound over the $2^{|\mathcal O|}$ vertices
and $c\in\mathcal C$ gives, with probability at least
$1-\delta/2$,
\[
\sup_{(h,\omega)\in\Pi_{\mathcal H}}
\left|
\widehat G_m(h;\omega;p^{(t)})
-G(h;\omega;p^{(t)})
\right|
\lesssim
\frac{M}{\sqrt m}
\left(
\Lambda BW_{\max}\sqrt{|\mathcal A|}
+
\sqrt{|\mathcal O|+\log\frac{2|\mathcal C|}{\delta}}
\right),
\]
where $\lesssim$ denotes an inequality up to a universal
multiplicative constant.

\textbf{Step 2: Uniform Convergence.}

We next control $\Pi_{\mathrm{soft}}$.
For fixed $c$, changing the loss vector from $\widetilde l$
to $\widetilde l'$ changes each predicted action loss by
at most $\|\widetilde l-\widetilde l'\|_\infty$.
The softmax rule consequently satisfies
\[
\left\|
k^{(\widetilde l,c)}(p^{(t)})(\cdot\mid z)
-
k^{(\widetilde l',c)}(p^{(t)})(\cdot\mid z)
\right\|_1
\leq
2\tau\|\widetilde l-\widetilde l'\|_\infty.
\]

Combining this with the loss and propensity bounds,
the change in each sample's contribution to
$\widehat G_m$ satisfies
\[
\begin{aligned}
&\Bigg|
\frac{
k^{(\widetilde l,c)}(p^{(t)})(a_i\mid z_i)
\ell_{a_i}^{(\widetilde l,c)}(z_i,o_i)
-
k^{(\widetilde l',c)}(p^{(t)})(a_i\mid z_i)
\ell_{a_i}^{(\widetilde l',c)}(z_i,o_i)
}{
\pi_{\mathrm{log}}(a_i\mid z_i)
}\\
&\qquad-
\sum_{a\in\mathcal A}
\Big(
k^{(\widetilde l,c)}(p^{(t)})(a\mid z_i)
\mathbb E_{o\sim p^{(t)}(\cdot\mid z_i,a)}
[\ell_a^{(\widetilde l,c)}(z_i,o)]
-
k^{(\widetilde l',c)}(p^{(t)})(a\mid z_i)
\mathbb E_{o\sim p^{(t)}(\cdot\mid z_i,a)}
[\ell_a^{(\widetilde l',c)}(z_i,o)]
\Big)
\Bigg|\\
&\leq
\left(1+\frac{1}{\gamma}\right)
\left(1+2\tau(1+B_c)\right)
\|\widetilde l-\widetilde l'\|_\infty
\leq
4M(1+\tau)\|\widetilde l-\widetilde l'\|_\infty.
\end{aligned}
\]

An $\ell_\infty$ cover of the loss cube therefore induces
a cover of these scalar contributions under the normalized
empirical $L_2$ metric.
Centering at the loss vector $\widetilde l=0$ removes a fixed
offset without changing the empirical Rademacher complexity.
Applying Dudley's entropy integral then shows that this
complexity, for fixed $c$, is
$\lesssim M(1+\tau)\sqrt{|\mathcal O|/m}$.

Using the envelope bound $2M$ and a union bound over
$c\in\mathcal C$, we obtain, with probability at least
$1-\delta/2$,
\[
\sup_{\omega\in\Omega}
\left|
\widehat G_m(k^\omega(p^{(t)});\omega;p^{(t)})
-
G(k^\omega(p^{(t)});\omega;p^{(t)})
\right|
\lesssim
\frac{M}{\sqrt m}
\left(
(1+\tau)\sqrt{|\mathcal O|}
+
\sqrt{\log\frac{2|\mathcal C|}{\delta}}
\right).
\]

Combining the two bounds, with probability at least
$1-\delta$,
\[
\sup_{(\pi,\omega)\in\Pi_{\mathrm{eval}}(p^{(t)})}
\left|
\widehat G_m(\pi;\omega;p^{(t)})
-G(\pi;\omega;p^{(t)})
\right|
\lesssim
\frac{1+B_c}{\gamma\sqrt m}
\left(
\Lambda BW_{\max}\sqrt{|\mathcal A|}
+
(1+\tau)\sqrt{|\mathcal O|}
+
\sqrt{\log\frac{2|\mathcal C|}{\delta}}
\right).
\]
Thus the uniform estimation error is at most
$\varepsilon/8$ whenever
\[
m\gtrsim
\frac{(1+B_c)^2}{\gamma^2\varepsilon^2}
\left(
|\mathcal A|\Lambda^2B^2W_{\max}^2
+
(1+\tau)^2|\mathcal O|
+
\log\frac{2|\mathcal C|}{\delta}
\right),
\]
where the universal implicit constant in this sample-size
requirement is sufficiently large.

On this event, consider an $\varepsilon/8$-approximate
maximizer $(\pi,\omega)$ of the absolute empirical gap.
If its absolute empirical gap exceeds $3\varepsilon/4$,
return this pair with
$\sigma=\operatorname{sign}
(\widehat G_m(\pi;\omega;p^{(t)}))$.
Its signed true gap satisfies
\[
\sigma G(\pi;\omega;p^{(t)})
\geq
\left|\widehat G_m(\pi;\omega;p^{(t)})\right|
-\frac{\varepsilon}{8}
>
\frac{5\varepsilon}{8}
>
\frac{\varepsilon}{2}.
\]
Otherwise, the supremum of the absolute empirical gap
is at most $7\varepsilon/8$, so the supremum of the
absolute true gap is at most $\varepsilon$.
Returning $\perp$ therefore certifies
$(\Omega,\mathcal H_\psi,\varepsilon)$-COI.
This completes the proof.
\end{proof}

\subsection{Extension to Multi-Agent Settings}

The proposed collaborative omnimechanism framework also
accommodates routing among $K$ human expert agents with
different query costs. We incorporate the expert identity
into the action space by writing $a=(r,s,g,j)$, where
$j\in[K]$ identifies the expert queried when $r=1$, and
$r=0$ corresponds to automation. Thus each deferred action
queries a single expert. Under this product representation,
the enlarged action space $\mathcal A'$ has size
$2K|\mathcal S||\mathcal G|$. Differences in experts'
responses and impressionability are represented by the
action-dependent outcome distribution $p^*(o\mid z,a)$.

Let $\mathcal C$ be a finite family of expert-cost profiles
$c=(c_1,\ldots,c_K)$, with $0\leq c_j(x)\leq B_c$ uniformly
over $c\in\mathcal C$, $j\in[K]$, and $x$.
For $\omega=(\tilde l,c)$, the total loss is
\[
    \ell_a^\omega(z,o)=\tilde l(o)+r c_j(x).
\]
Since each deferred action queries only one expert,
the total loss remains bounded by $1+B_c$.

Let $\mathcal H'$ denote the resulting collaboration policy
class, with Natarajan dimension $d_{\mathcal H'}$.
Assume that the logging propensities are known and satisfy
$\pi_{\mathrm{log}}(a\mid z)\geq\gamma'>0$ for every
$a\in\mathcal A'$ and state $z$.
For fixed finite inverse temperature $\tau>0$, under
the remaining assumptions of the auditing theorem,
the sample-complexity analysis applies to this enlarged
action space and expert-dependent loss.
In particular, a sufficient number of samples per audit is
\[
m\gtrsim
\frac{(1+B_c)^2}{(\gamma')^2\varepsilon^2}
\left[
d_{\mathcal H'}\log|\mathcal A'|
+(1+\tau)^2|\mathcal O|
+\log\frac{2|\mathcal C|}{\delta}
\right],
\]
where $|\mathcal C|$ counts the admissible expert-cost
profiles.

The statistical cost of accommodating additional experts
is therefore governed by the enlarged policy class,
action space, logging coverage, and family of cost profiles.
Computational efficiency additionally depends on the
procedure used to optimize the empirical auditing objective
over $\mathcal H'$.



\section{Additional Details for Experiment 1: Impressionable Human Decisions}
\label{app:experimental_details}

\subsection{Dataset}
\label{app:dataset}
We use the human-AI interaction dataset of \citep{vodrahalli2022uncalibrated}. The dataset contains more than 30,000 human-AI interactions across several binary classification tasks. Each interaction records the human's initial response, the AI recommendation, and the human's final response after observing the AI advice. Responses are encoded on a signed confidence scale in $[-1,1]$: the sign indicates whether the selected label agrees with the ground-truth label, and the magnitude indicates the confidence strength. Thus, positive values correspond to correct predictions, negative values correspond to incorrect predictions, and values near zero indicate low confidence. The dataset also includes participant-level demographic and survey variables, which are used to fit the behavioral simulator following the original setup, but are not used in our omnimechanism design.

We focus on the Art and Sarcasm tasks. These two tasks provide sufficient interactions to support our full experimental pipeline: fitting a behavioral simulator, optimizing policies, and evaluating the learned policies on held-out examples. The remaining tasks contain data only for training the behavioral simulator, but not additional held-out interactions for reliable policy optimization and evaluation. We therefore exclude them from the main experiments.

For each task, we reserve a separate participant pool for fitting the behavioral simulator, as in \citep{vodrahalli2022uncalibrated}. The remaining participants are split approximately equally between policy training and held-out evaluation using seed 42. All interactions from a participant remain in the same partition, and the simulator-fitting, policy-training, and evaluation pools are mutually disjoint.



\subsection{Human Behavioral Simulator}
\label{app:behavioral_simulator}

We use the two-stage behavioral simulator from \citep{vodrahalli2022uncalibrated} as a proxy for human behavior. Given the initial human response $r_1$ and the AI advice, the simulator first predicts the probability that the human changes their response after seeing the advice, denoted by $p_{\mathrm{act}}$. Conditional on activation, it predicts the signed response change
\[
    \Delta = \mathrm{sign}(r_1)(r_2-r_1).
\]
The corresponding activated response is
\[
    \hat r_2^{\mathrm{act}}
    =
    \mathrm{clip}\bigl(r_1+\mathrm{sign}(r_1)\hat\Delta,\,-1,\,1\bigr).
\]
If the human is not activated, the final response remains $r_1$.

For any transformed AI advice, the simulator therefore induces a mixture over two possible final responses: $r_1$ with probability $1-p_{\mathrm{act}}$, and $\hat r_2^{\mathrm{act}}$ with probability $p_{\mathrm{act}}$.
For a collaboration policy with human weight $w$ and AI weight $1-w$, the final output is the weighted combination of the human response and the original AI signal. Its expected accuracy is
\[
\mathrm{Acc.}
=
\mathbb{E}\left[
(1-p_{\mathrm{act}})
\mathbf{1}\{w r_1+(1-w)r_{\mathrm{AI}}>0\}
+
p_{\mathrm{act}}
\mathbf{1}\{w\hat r_2^{\mathrm{act}}+(1-w)r_{\mathrm{AI}}>0\}
\right].
\]
Here, $w=1$ corresponds to using the human response alone.
For the AI-only route, accuracy is
$\mathbb{E}[\mathbf{1}\{r_{\mathrm{AI}}>0\}]$.




\subsection{Omnimechanism Design}
\label{app:omnimechanism_design}
Both the boosting-based and neural implementations instantiate the same omnimechanism objective: learning a base predictive distribution over outcomes for each state-action pair, which can then be post-processed into policies for different downstream costs. They differ only in how the base predictive distribution is represented and optimized.

\textbf{State representation.}
The observable state $z$ contains only pre-interaction information available before deciding how to use the AI signal. We use
\[
    z =
    \bigl(
    |r_1|,\ |r_{\mathrm{AI}}|,\ 
    \mathbf{1}\{\mathrm{sign}(r_1)=\mathrm{sign}(r_{\mathrm{AI}})\}
    \bigr),
\]
where $r_1$ is the human's initial signed confidence and $r_{\mathrm{AI}}$ is the raw signed AI signal. The first two coordinates measure the human and AI confidence strengths, while the third records whether the human and AI initially agree on the label. This representation removes the direct correctness information carried by the signs of the signed responses, while still retaining whether the human and AI initially agree.

For the boosting-based implementation, we discretize the two confidence coordinates with step size $s_z$ and keep the agreement coordinate binary. In the main experiments, we use $s_z=0.1$. For the neural implementation, we use the same three state features in continuous form, together with action features.

\textbf{Action space.}
An action $a$ specifies how the AI advice is used. It consists of three components: a deferral decision, a signal transformation, and a collaboration function. The deferral decision has two possible outcomes: either the system directly outputs the AI prediction, or it routes the instance to the human. We refer to these as the AI-only route and the Human+AI route.

Under the AI-only route, the final output is the raw AI signal $r_{\mathrm{AI}}$. Under the Human+AI route, the AI signal is first transformed and then passed to the behavioral simulator to produce a simulated human response.

We follow the signal transformation in \citep{vodrahalli2022uncalibrated}, parameterized by $(\alpha,\beta)$. We convert the signed AI signal $r_{\mathrm{AI}}$ into a probability
\[
    p_{\mathrm{AI}} = \frac{r_{\mathrm{AI}}+1}{2},
\]
which we clip to $[10^{-6},1-10^{-6}]$ before converting to a logit:
\[
    A = \log\frac{p_{\mathrm{AI}}}{1-p_{\mathrm{AI}}}.
\]
Using the logistic sigmoid $\sigma(u)=(1+e^{-u})^{-1}$,
the transformed AI confidence is
\[
    \phi_{\alpha,\beta}(A)
    =
    \sigma\left(\mathrm{sign}(A)(\alpha |A|+\beta)\right),
\]
and the corresponding transformed signed signal is
\[
    r'_{\mathrm{AI}}
    =
    2\phi_{\alpha,\beta}(A)-1.
\]
We use the parameter grids $\alpha\in\{0.5,1.0,1.5,2.0\}$
and $\beta\in\{0,0.5,1.0,1.5,2.0\}$.



The collaboration function is used only under the Human+AI route. Let $\hat r_{\mathrm{human}}$ denote the simulator-induced human response after observing the transformed AI signal. The final routed output is
\[
    r_{\mathrm{final}}
    =
    w \hat r_{\mathrm{human}} + (1-w)r_{\mathrm{AI}},
\]
where $w\in(0,1]$ is a collaboration weight. We use a grid over $w$; with step size $0.5$, the candidate values are $w\in\{0.5,1.0\}$. The case $w=1$ corresponds to using only the simulated human final response.

In our policy class $\cH$, deferral functions are implemented by a threshold rule on AI confidence: it routes the instance to AI-only when $|r_{\mathrm{AI}}|$ exceeds a threshold, and otherwise routes it to Human+AI. We also include always-human and always-AI as special cases. Although there may be many different deferral functions, each ultimately outputs only one of the two route outcomes for any fixed state: AI-only or Human+AI. Thus, deferral functions define candidate policies over states, while the action outcome itself is determined by the binary route together with the chosen signal transformation and collaboration function.

The following remark explains how this experimental action-space
construction fits the finite-action framework underlying our
sample-complexity guarantees.

\begin{remark}[Context-dependent signal presentation]
\label{rem:context-dependent-signals}
Let $\xi\in\Xi$ denote the full observable pre-interaction input, including the instance and the initial human and AI responses, and let $z=\psi(\xi)$ denote the three-dimensional feature representation defined above. Let $\mathcal U$ be a fixed finite family of measurable signal-transformation rules $u:\Xi\to\mathcal S$, where $\mathcal S$ is the ambient signal space. We index the implemented interventions by
\[
    \bar{\mathcal A}
    \subseteq \{0,1\}\times\mathcal U\times\mathcal G.
\]
The policy selects an action $(b,u,g)$ using $z$, while the selected transformation is evaluated on the full input: the implemented action is $(b,u(\xi),g)$. Here distinct rules may produce the same displayed signal.

Assume that the logging policy also selects indexed actions using only $z$, as is the case for uniform logging. The outcome distribution $p^*(\cdot\mid z,a)$ then averages over full inputs conditional on $z$. Under the remaining assumptions of the  corresponding auditing theorem, including known logging propensities
and coverage over $\bar{\mathcal A}$, the finite-action
sample-complexity guarantees apply with $|\bar{\mathcal A}|$
and the complexity of the induced policy class.
Thus, the displayed signal may depend on the full input
even though policy selection and outcome prediction use
the reduced features $z$.
\end{remark}

\textbf{Outcome representation.}
For each state-action pair $(z,a)$, the base predictive distribution predicts a four-class advice-relative outcome
\[
    o \in \{0,1,2,3\}.
\]
Each outcome records two binary quantities: whether the final prediction agrees with the AI signal, and whether the AI signal is correct. We define the classes as
\[
\begin{array}{ccl}
0 &: & \mathrm{final\_agree\_AI}=0,\quad \mathrm{AI\_correct}=0, \\
1 &: & \mathrm{final\_agree\_AI}=0,\quad \mathrm{AI\_correct}=1, \\
2 &: & \mathrm{final\_agree\_AI}=1,\quad \mathrm{AI\_correct}=0, \\
3 &: & \mathrm{final\_agree\_AI}=1,\quad \mathrm{AI\_correct}=1.
\end{array}
\]
This representation is sufficient to determine whether the final output is correct. If the AI signal is incorrect, then disagreeing with it is correct, so class $0$ is correct and class $2$ is incorrect. If the AI signal is correct, then agreeing with it is correct, so class $3$ is correct and class $1$ is incorrect. Therefore, classes $0$ and $3$ correspond to correct final outputs, while classes $1$ and $2$ correspond to errors.

For losses based on prediction error, we use
\[
    l_{\mathrm{err}}(o)
    =
    \mathbf{1}\{o\in\{1,2\}\}.
\]
This error loss is combined with human-involvement cost when computing COI violations during auditing, and the same outcome representation is used for post-processing under different downstream costs.

\textbf{Training targets.}
For Human+AI actions, we replace hard correctness with
$\sigma(T r_{\mathrm{final}})$ and average over the simulator's
activation mixture. We map the resulting soft correctness to
the four outcome classes above, obtaining $q^{(0)}$.
For AI-only actions, $q^{(0)}$ is one-hot.
Both implementations use the smoothed target
\[
q_{\mathrm{sim}}
=(1-4\lambda)q^{(0)}+\lambda\mathbf{1}_4,
\]
with $T=5$ and $\lambda=10^{-4}$.
This smoothing is used for learning; reported accuracy is computed
using hard sign-based correctness.

\subsection{Boosting-Based Omnimechanism}
\label{app:boosting_omnimechanism}

We implement the boosting-based omnimechanism by discretizing the state space and maintaining a tabular base predictive distribution over outcomes for each discrete state-action pair. Specifically, after discretizing the confidence coordinates of $z$, we maintain
\[
    \tilde p^{(t)}(o\mid z,a),
\]
where $o\in\{0,1,2,3\}$ is the outcome described above and $a$ is an action.

For a downstream specification $\omega=(l_{\mathrm{err}},c)$, we define the action-dependent loss
\[
    \ell_a^\omega(z,o)
    =
    l_{\mathrm{err}}(o)
    +
    c\cdot \mathbf{1}\{a \text{ routes to human}\},
\]
where
\[
    l_{\mathrm{err}}(o)
    =
    \mathbf{1}\{o\in\{1,2\}\}.
\]
Here $l_{\mathrm{err}}$ denotes the prediction-error component, while $\ell_a^\omega$ denotes the full downstream loss including the human-involvement cost.

At each boosting round, we audit the current base predictive distribution against a collection of candidate policies. This collection includes all fixed policies in the candidate class $\cH$, as well as the current optimal post-processing policy induced by the predictor for each downstream cost parameter $c$:
\[
    \pi^\omega(z)
    \in
    \arg\min_a
    \mathbb{E}_{\tilde p^{(t)}(\cdot\mid z,a)}
    \left[
        \ell_a^\omega(z,o)
    \right].
\]
For a candidate mechanism $h$ and downstream specification $\omega$, we compute the empirical COI violation
\[
    \varepsilon(h,\omega)
    =    
    \mathbb{E}_{\mathrm{sim}}
    \left[
        \ell_{\pi_h(z)}^\omega(z,o)
    \right]-\mathbb{E}_{\tilde p^{(t)}}
    \left[
        \ell_{\pi_h(z)}^\omega(z,o)
    \right], 
\]
where the simulator expectation is estimated using the behavioral simulator and the predictor expectation is computed from $\tilde p^{(t)}(\cdot\mid z,\pi_h(z))$. We then select the candidate with the largest absolute violation:
\[
    (h_t,\omega_t)
    =
    \arg\max_{h,\omega}
    |\varepsilon(h,\omega)|.
\]
If the maximum violation is below a tolerance threshold $\varepsilon$, the procedure terminates.

Otherwise, we update the base predictive distribution on the action selected by the violating policy. Let
\[
    \sigma_t = \mathrm{sign}(\varepsilon(h_t,\omega_t)).
\]
For each discrete state $z$, only the entry corresponding to the selected action $a=\pi_{h_t}(z)$ is updated:
\[
    \tilde p^{(t+1)}(o\mid z,a)
    \propto
    \tilde p^{(t)}(o\mid z,a)
    \exp\left(
\eta\,\sigma_t\,l_{\mathrm{err}}(o)
    \right),
    \qquad a=\pi_{h_t}(z),
\]
followed by normalization over $o$. Entries for actions not selected by $h_t$ remain unchanged. The human-involvement cost affects which policies are selected during auditing and post-processing, but it is outcome-independent and therefore is not included in the multiplicative update.

After training, the learned base predictive distribution is reused for different downstream costs without retraining. In the experiments, we use deterministic post-processing, selecting the action with the smallest predicted expected loss. Thus, for any $\omega=(l_{\mathrm{err}},c)$, the induced policy is obtained by post-processing:
\[
    \pi^\omega(z)
    \in
    \arg\min_a
    \mathbb{E}_{\tilde p(\cdot\mid z,a)}
    \left[
        \ell_a^\omega(z,o)
    \right].
\]
In our experiments, we use discretization step $0.1$, multiplicative-weights learning rate $\eta=0.15$, and COI tolerance $\varepsilon=0.05$. We also record the KL divergence from the smoothed simulator target to the learned predictive distribution over boosting rounds as a diagnostic of distribution-level convergence, as shown in Figure~\ref{fig:boosting-comparison}.

\subsection{Neural Omnimechanism}
\label{app:neural_omnimechanism}

We also implement a neural version of the omnimechanism to avoid discretizing the state space. It uses the same state, primitive action, and outcome definitions as above, but represents the base predictive distribution with a neural network:
\[
    \tilde p_\theta(o\mid z,a).
\]
The network takes the continuous state $z$ and primitive action features $(\alpha,\beta,\mathrm{route},w)$ as input, and outputs logits over the four outcome classes.

For each training example and primitive action, we query the
behavioral simulator to construct the target distribution
$q_{\mathrm{sim}}$ defined above.






The neural predictor is trained on the targets $q_{\mathrm{sim}}$ by minimizing soft-label cross entropy:
\[
    \mathcal{L}_{\mathrm{NN}}
    =
    -
    \mathbb{E}_{(z,a)}
    \sum_o
    q_{\mathrm{sim}}(o\mid z,a)
    \log \tilde p_\theta(o\mid z,a).
\]
We use a three-layer perceptron with two 64-unit hidden layers, ReLU activations, and dropout $0.1$, trained with Adam at learning rate $10^{-3}$ and batch size 512. We reserve $30\%$ of the policy-training participants for validation and select the checkpoint with the lowest validation cross entropy. We also audit empirical COI violations on the validation set, using simulator losses based on accuracy.



At evaluation time, we use the same deterministic post-processing rule as in Appendix~\ref{app:boosting_omnimechanism}: for each downstream specification $\omega=(l_{\mathrm{err}},c)$, the neural omnimechanism induces the policy
\[
    \pi^\omega(z)
    \in
    \arg\min_a
    \mathbb{E}_{\tilde p_\theta(\cdot\mid z,a)}
    \left[
        \ell_a^\omega(z,o)
    \right],
\]
using the same loss definition and evaluation protocol as in Appendix~\ref{app:boosting_omnimechanism}. The selected action is then evaluated with the behavioral simulator rather than with the neural prediction itself.

\subsection{Baselines and Metrics}
\label{app:baselines_metrics}

We compare the post-processing policies induced by our omnimechanism with three baselines: \textbf{No Action}, \textbf{Manipulation}, and \textbf{Global Best}.

\textbf{No Action.}
This corresponds to the basic Human+AI setting in which every example is deferred to the human with original AI advice and final prediction is human's final response.

\textbf{Manipulation.}
This baseline is the AI-advice manipulation method from \citep{vodrahalli2022uncalibrated}. It always routes the instance to the human and optimizes only the global AI signaling function
$\phi_{\alpha,\beta}$.
The optimization is performed over the continuous $(\alpha,\beta)$ space using the same parameter range as our action space.

\textbf{Global Best.}
This baseline selects the best single policy from the candidate
policy class $\cH$ for each downstream specification $\omega$
and holds that policy fixed across inputs. The policy class $\cH$ contains global combinations of the three action components introduced above: a deferral function, a signaling function, and a collaboration function. The deferral functions include always-human, always-AI, and threshold-based rules that defer to the human when the AI confidence is below a threshold; the thresholds are chosen on a grid with step size $0.2$. The signaling functions are drawn from the grid of transformations
$\phi_{\alpha,\beta}$, and the collaboration functions are linear
combinations parameterized by the grid of weights $w$, as described
in Appendix~\ref{app:omnimechanism_design}. For each cost parameter $c$, Global Best is chosen on the policy-optimization data using simulator utility and then evaluated on the held-out evaluation data.
Unlike Global Best, which uses a fixed routing rule, signaling
function, and collaboration weight, our post-processing can
select the complete action separately for each input.



\textbf{Metrics.}
All methods are evaluated on the held-out evaluation set using the behavioral simulator. Accuracy is computed using the simulator-induced activation mixture defined in Appendix~\ref{app:behavioral_simulator}.
We report human involvement rate as
\[
    \mathrm{Hum.}
    =
    \mathbb{E}\left[
        \mathbf{1}\{a \text{ routes to human}\}
    \right],
\]
and utility for cost parameter $c$ as
\[
    Util._c
    =
    \mathrm{Acc.}
    -
    c\cdot \mathrm{Hum.}
\]

We also assess stability across random seeds. Across the evaluated
runs, the empirical standard deviation of the reported metrics
is below one percentage point.



\subsection{Supplementary Experiment: Varying AI Advice Accuracy}
\label{app:varying_ai_accuracy}

We conduct an additional experiment to further validate the effectiveness of our methods under different AI advice accuracy levels. Starting from the original AI advice distribution, whose average accuracy is around $80\%$, we apply a global downward shift to the signed AI signal and clip the result to $[-1,1]$. This produces two additional settings in which the average AI advice accuracy is approximately $70\%$ and $60\%$. We focus on downward shifts to examine settings in which human consultation may become more valuable. The realized AI-only accuracies are $69.8\%$ and $73.7\%$
for Art and Sarcasm in the first setting, and $59.6\%$ and
$63.1\%$ in the second.



For each shifted setting, we rerun the full experimental pipeline, including training the boosting-based and neural omnimechanisms and reselecting the corresponding baselines. This allows us to compare our methods with the strongest available baselines under each AI-advice accuracy level, and to examine whether the learned post-processing policies still achieve competitive utility when the reliability of the AI signal changes.

\begin{table}[!t]
\centering
\caption{Policy performance after shifting AI advice to approximately $70\%$ accuracy. We report accuracy (Acc.), utility ($\mathrm{Util.}=\mathrm{Acc.}-c\cdot\mathrm{Hum.}$), and human-involvement rate (Hum.). The best and second-best utilities in each setting are shown in \textbf{bold} and \underline{underline}, respectively.}
\label{tab:shift-ai-70}
\vspace{0.5em}
\renewcommand{\arraystretch}{0.95}
\setlength{\tabcolsep}{5pt}
\resizebox{0.9\linewidth}{!}{
\begin{tabular*}{\textwidth}{@{\extracolsep{\fill}}llccc ccc}
\toprule
& & \multicolumn{3}{c}{\textbf{Art}} & \multicolumn{3}{c}{\textbf{Sarcasm}} \\
\cmidrule(lr){3-5} \cmidrule(lr){6-8}
$c$ & \textbf{Method}
& \textbf{Acc.} & \textbf{Util.} & \textbf{Hum.}
& \textbf{Acc.} & \textbf{Util.} & \textbf{Hum.} \\
\midrule
\multirow{5}{*}{0.0}
& No Action                 & 0.695 & 0.695 & 1.000 & 0.699 & 0.699 & 1.000 \\
& Manipulation             & 0.697 & 0.697 & 1.000 & 0.710 & 0.710 & 1.000 \\
& Global Best              & 0.702 & \textbf{0.702} & 0.182 & 0.743 & 0.743 & 0.532 \\
& Ours (boosting)          & 0.698 & \underline{0.698} & 0.709 & 0.757 & \textbf{0.757} & 0.450 \\
& Ours (NN)                & 0.698 & \underline{0.698} & 0.001 & 0.745 & \underline{0.745} & 0.157 \\
\midrule
\multirow{5}{*}{0.1}
& No Action                 & 0.695 & 0.595 & 1.000 & 0.699 & 0.599 & 1.000 \\
& Manipulation             & 0.697 & \underline{0.597} & 1.000 & 0.710 & 0.610 & 1.000 \\
& Global Best              & 0.698 & \textbf{0.698} & 0.000 & 0.737 & \textbf{0.737} & 0.000 \\
& Ours (Boosting)          & 0.698 & \textbf{0.698} & 0.000 & 0.737 & \textbf{0.737} & 0.000 \\
& Ours (NN)                & 0.698 & \textbf{0.698} & 0.000 & 0.741 & \underline{0.734} & 0.076 \\
\midrule
\multirow{5}{*}{0.2}
& No Action                & 0.695 & 0.495 & 1.000 & 0.699 & 0.499 & 1.000 \\
& Manipulation             & 0.697 & \underline{0.497} & 1.000 & 0.710 & 0.510 & 1.000 \\
& Global Best              & 0.698 & \textbf{0.698} & 0.000 & 0.737 & \textbf{0.737} & 0.000 \\
& Ours (Boosting)          & 0.698 & \textbf{0.698} & 0.000 & 0.737 & \textbf{0.737} & 0.000 \\
& Ours (NN)                & 0.698 & \textbf{0.698} & 0.000 & 0.740 & \underline{0.733} & 0.035 \\
\bottomrule
\end{tabular*}}
\vspace{-0.5em}
\end{table}

\begin{table}[!ht]
\centering
\caption{Policy performance after shifting AI advice to approximately $60\%$ accuracy. We report accuracy (Acc.), utility ($\mathrm{Util.}=\mathrm{Acc.}-c\cdot\mathrm{Hum.}$), and human-involvement rate (Hum.). The best and second-best utilities in each setting are shown in \textbf{bold} and \underline{underline}, respectively.}
\label{tab:shift-ai-60}
\vspace{0.5em}
\renewcommand{\arraystretch}{0.95}
\setlength{\tabcolsep}{5pt}
\resizebox{0.9\linewidth}{!}{
\begin{tabular*}{\textwidth}{@{\extracolsep{\fill}}llccc ccc}
\toprule
& & \multicolumn{3}{c}{\textbf{Art}} & \multicolumn{3}{c}{\textbf{Sarcasm}} \\
\cmidrule(lr){3-5} \cmidrule(lr){6-8}
$c$ & \textbf{Method}
& \textbf{Acc.} & \textbf{Util.} & \textbf{Hum.}
& \textbf{Acc.} & \textbf{Util.} & \textbf{Hum.} \\
\midrule
\multirow{5}{*}{0.0}
& No Action                 & 0.618 & \textbf{0.618} & 1.000 & 0.673 & 0.673 & 1.000 \\
& Manipulation             & 0.618 & \textbf{0.618} & 1.000 & 0.671 & 0.671 & 1.000 \\
& Global Best              & 0.615 & {0.615} & 0.409 & 0.681 & \underline{0.681} & 0.797 \\
& Ours (boosting)          & 0.617 & \underline{0.617} & 0.602 & 0.683 & \textbf{0.683} & 0.752 \\
& Ours (NN)                & 0.616 & {0.616} & 0.273 & 0.677 & 0.677 & 0.336 \\
\midrule
\multirow{5}{*}{0.1}
& No Action                 & 0.618 & 0.518 & 1.000 & 0.673 & 0.573 & 1.000 \\
& Manipulation             & 0.618 & 0.518 & 1.000 & 0.671 & 0.571 & 1.000 \\
& Global Best              & 0.596 & \underline{0.596} & 0.000 & 0.668 & {0.641} & 0.272 \\
& Ours (Boosting)          & 0.596 & \underline{0.596} & 0.000 & 0.663 & \underline{0.643} & 0.202 \\
& Ours (NN)                & 0.605 & \textbf{0.603} & 0.021 & 0.669 & \textbf{0.651} & 0.184 \\
\midrule
\multirow{5}{*}{0.2}
& No Action                & 0.618 & \underline{0.418} & 1.000 & 0.673 & 0.473 & 1.000 \\
& Manipulation             & 0.618 & \underline{0.418} & 1.000 & 0.671 & 0.471 & 1.000 \\
& Global Best              & 0.596 & \textbf{0.596} & 0.000 & 0.631 & \underline{0.631} & 0.000 \\
& Ours (Boosting)          & 0.596 & \textbf{0.596} & 0.000 & 0.631 & \underline{0.631} & 0.000 \\
& Ours (NN)                & 0.596 & \textbf{0.596} & 0.000 & 0.668 & \textbf{0.639} & 0.145 \\
\bottomrule
\end{tabular*}}
\vspace{-0.5em}
\end{table}



Tables~\ref{tab:shift-ai-70} and~\ref{tab:shift-ai-60} report the results. When AI accuracy is reduced to roughly $70\%$, the learned policies remain competitive with Global Best across both tasks. On Art, the differences between methods are small, and for nonzero costs both Global Best and our methods quickly move toward AI-only behavior. On Sarcasm, where the value of human involvement is more state-dependent, our methods obtain the best or near-best utilities across cost values.

When AI accuracy is reduced to roughly $60\%$, the trade-off between accuracy and human involvement becomes more pronounced. When human involvement is free, the learned policies make
substantial use of human input and improve accuracy over AI-only. As the human cost increases, the policies reduce human involvement and become more selective. On Art, our methods closely match the strongest utilities across cost values, indicating that the learned policies can recover the appropriate low-deferral behavior when human involvement is not cost-effective. On Sarcasm, both implementations improve upon Global Best at
$c=0.1$; at $c=0.2$, NN retains selective human involvement and
higher utility, while Boosting uses AI-only and matches Global Best.



\textbf{Overall, this supplementary experiment shows that our conclusions are not specific to the original AI accuracy level.} After retraining under each shifted AI-advice distribution, the omnimechanism-based policies remain competitive with the strongest global baselines and reflect the expected trade-off between accuracy and human cost.

\section{Additional Details for Experiment 2: Radiologist--AI Collaboration}
\label{app:radiology_experiment}

\subsection{Data and Action Construction}
Collab-CXR contains probabilistic assessments from 227 professional
radiologists on 324 retrospective chest X-rays, collected under four
information environments and three experimental designs
\citep{moehring2025dataset}.  Our analysis uses Design~1, in which 159
radiologists each read 60 unique cases, with 15 cases assigned to each
randomized information environment.  A radiologist encounters each patient at
most once, yielding one-shot assessments that match our action-selection
setting.  Designs~2 and~3 study repeated or sequential assessments and are not
included in this experiment.

We study Airspace Opacity and Cardiomediastinal Abnormality, the two
pre-registered top-level pathologies for which AI predictions are available in
the original analysis.  The original experiment varies both AI assistance and
the availability of clinical history to study how these information sources
affect radiologists' assessments.  We restrict attention to conditions without
clinical history and consider three actions: the direct algorithmic probability
(\textsc{AI}), a radiologist's probability based on the X-ray alone
(\textsc{H}), and a radiologist's probability after observing both the X-ray
and the AI assessment (\textsc{AI+H}).  We charge a human-involvement cost $c$
whenever a radiologist is consulted: \textsc{AI} incurs no such cost, while
\textsc{H} and \textsc{AI+H} each incur cost $c$.
Clinical-history conditions could be incorporated as additional actions but
may induce additional cost.  We omit them to keep the action space focused on
how AI and radiologists are combined, with a single cost for radiologist
involvement.

After restricting to the two conditions without clinical history, the 159
radiologists contribute $159\times30=4{,}770$ X-ray reads in total:
$159\times15=2{,}385$ under \textsc{H} and the same number under
\textsc{AI+H}.  Each read contains an assessment for both focal pathologies,
yielding $2{,}385\times2=4{,}770$ pathology-level probability reports per
human action and 9,540 across the two actions.  Both human actions cover all
$324\times2=648$ patient--pathology inputs.  The direct AI probability is
unique within each patient--pathology pair.

\subsection{Omnimechanism Design}
Let $y\in\{0,1\}$ denote the binary diagnostic-standard label supplied with
each patient--pathology pair.  The accompanying study evaluates diagnostic
probabilities using the average deviation from this standard
\citep{agarwal2023combining}.  For a reported probability
$\rho\in[0,1]$, we encode the following fractional four-state target:
\begin{equation}
\begin{split}
q_{00}&=(1-y)(1-\rho), \qquad q_{01}=y(1-\rho),\\
q_{10}&=(1-y)\rho,   \qquad \; \;   \qquad q_{11}=y\rho.
\end{split}
\label{eq:radiology-four-state}
\end{equation}
We interpret $\rho$ as a distribution over the binary diagnostic label,
assigning probability $1-\rho$ to label $0$ and probability $\rho$ to label
$1$.  The first
index of $q$ denotes this predicted label, while the second denotes the
diagnostic standard.  The diagonal mass gives the diagnostic score used in the
main text:
\begin{equation}
S(\rho,y)=q_{00}+q_{11}=1-|\rho-y|.
\label{eq:radiology-deviation-mass}
\end{equation}
Maximizing $S$ is therefore equivalent to minimizing the deviation measure. 
The fractional representation retains the
reported probability rather than converting it to a binary prediction.

For \textsc{H} and \textsc{AI+H}, we construct
Eq.~\ref{eq:radiology-four-state} for every reader and average the four masses
within each patient--pathology--action cell.  For \textsc{AI}, we construct the
target once from the direct AI probability.  This gives
$324\times2\times3=1{,}944$ cells, which are weighted equally during training.
Repeated radiologist assessments estimate the empirical outcome distribution
within a cell; reader identity is not included as a model input.

\subsection{Neural Omnimechanism}
Patients are split 80/20 using seed 42, stratified over the four joint binary
label combinations for the two pathologies, so that their relative frequencies
are approximately preserved across the training and test sets.  The resulting
split contains 259 training patients and 65 test patients.  All actions
and pathologies associated with an X-ray remain in the same split.

We extract one 1,152-dimensional representation per X-ray using the
MedSigLIP-448 vision encoder \citep{sellergren2025medgemma}.  A 32-component
PCA transform is fit on the training embeddings and then used to project both
training and test embeddings, retaining 83.27\% of the training embedding
variance.  During cross-validation, the same procedure is performed within
each fold: PCA is fit on the fold's training patients and applied to its
validation patients.  After L2 normalization, the model input concatenates the
image vector, scalar AI probability, two-dimensional pathology one-hot
encoding, and three-dimensional action one-hot encoding, for 38 input
features.

The neural omnimechanism is a two-hidden-layer MLP with width 32, ReLU
activations, and dropout 0.1 after the first hidden layer.  A four-way softmax
produces $\widehat q(z,a)$.  We minimize fractional-target cross-entropy using
AdamW, learning rate $10^{-3}$, weight decay $10^{-4}$, and batch size 128.
We select the number of training epochs by five-fold patient-grouped,
joint-label-stratified cross-validation, choosing the earliest epoch whose
mean validation loss lies within one standard error of the minimum.  The final
PCA transform and neural model are fit using all training patients, without
access to the test set.

\subsection{Evaluation and Additional Results}
For each input, the same trained model is evaluated with all three actions. 
At cost $c$, it chooses
\begin{equation}
\pi_c(z)\in\arg\max_{a\in\{\textsc{AI},\textsc{H},\textsc{AI+H}\}}
\left\{\widehat q_{00}(z,a)+\widehat q_{11}(z,a)
-c\mathbf{1}\{a\ne\textsc{AI}\}\right\}.
\label{eq:radiology-policy}
\end{equation}
The comparator is Global Best: separately for each $c$, we use training cells
to select one action maximizing mean
$S(\rho,y)-c\cdot\mathrm{Hum.}$ and apply that single action to every test input.
Both the input-dependent policy and Global Best are fixed before evaluation on
the test patients.

\begin{table}[!t]
\centering
\caption{Actions selected by the post-processed policy on the Collab-CXR test
set.  Each pathology contains 65 test X-rays.}
\label{tab:radiology-selection}
\vspace{0.4em}
\renewcommand{\arraystretch}{0.95}
\setlength{\tabcolsep}{6pt}
\resizebox{0.72\linewidth}{!}{
\begin{tabular}{clrrrr}
\toprule
$c$ & Pathology & \textsc{AI} & \textsc{H} & \textsc{AI+H} & Hum. \\
\midrule
\multirow{2}{*}{0.00} & Airspace Opacity          &  4 & 39 & 22 & 0.938 \\
                      & Cardiomediastinal Abnorm. & 14 &  4 & 47 & 0.785 \\
\midrule
\multirow{2}{*}{0.02} & Airspace Opacity          & 30 & 29 &  6 & 0.538 \\
                      & Cardiomediastinal Abnorm. & 64 &  0 &  1 & 0.015 \\
\midrule
\multirow{2}{*}{0.05} & Airspace Opacity          & 60 &  5 &  0 & 0.077 \\
                      & Cardiomediastinal Abnorm. & 65 &  0 &  0 & 0.000 \\
\midrule
\multirow{2}{*}{0.10} & Airspace Opacity          & 65 &  0 &  0 & 0.000 \\
                      & Cardiomediastinal Abnorm. & 65 &  0 &  0 & 0.000 \\
\midrule
\multirow{2}{*}{0.20} & Airspace Opacity          & 65 &  0 &  0 & 0.000 \\
                      & Cardiomediastinal Abnorm. & 65 &  0 &  0 & 0.000 \\
\bottomrule
\end{tabular}}
\vspace{-0.5em}
\end{table}

Table~\ref{tab:radiology-selection} gives the action composition underlying
the performance results in Table~\ref{tab:radiology-main}.  With no human
cost, the policy uses both unaided and AI-assisted radiologist assessments.
As $c$ increases, human involvement declines first for Cardiomediastinal
Abnormality and then for Airspace Opacity.  At $c\geq0.10$, all test inputs are
assigned to \textsc{AI}.

\section{Additional Details for Experiment 3: AI-Assisted House-Price Prediction}
\label{app:ames_experiment}

\subsection{Data and Action Construction}
We use the AMES house-price judgment data of \citet{guo2025value}.
The released trial table contains 437 participants and 5,244 paired judgments
on 240 distinct houses. Each participant evaluates 12 houses, first providing
all 12 unaided estimates and then revising these estimates after receiving AI
assistance. Participants are assigned to one of six information conditions,
crossing two AI models with three explanation types: no explanation, SHAP,
and ILIV+SHAP. The assigned condition remains fixed across a participant's
assisted trials.

For each participant--house pair, let $H_0$ and $H_1$ denote the initial and
assisted human estimates, $M_1$ and $M_2$ the two AI predictions, and $y>0$
the sale price. Human estimates in the released table are multiplied by
1,000 to place all predictions and labels on the same dollar scale. The
context is $z=(x,H_0,M_1,M_2)$, where $x$ contains the six house features:
year built, above-ground living area, garage capacity, number of fireplaces,
year remodeled, and overall quality. Sale price, assisted human responses,
participant identity, and reward variables are excluded from the policy
inputs.

We study a sequential decision made after observing $H_0$ and we consider the action space $\mathcal{A}$ consisting of the two parts below. The first action set is defined by  $\mathcal A_0=\{\textsc{H}_0,\textsc{AI}_1,\textsc{AI}_2\}$ where the actions directly output $H_0$, $M_1$, and $M_2$, respectively. The second action set is defined by  $\mathcal A_1=\{\textsc{Rev}_{j,e}:j\in\{1,2\},\, e\in \{\text{none},\text{SHAP},\text{ILIV+SHAP}\}\}$, where the actions request a human
revision after displaying model $j$ with explanation $e$, and use that
revised estimate as the final output. Thus
$\mathcal A=\mathcal A_0\cup\mathcal A_1$ contains nine actions. We charge
a human-involvement cost $c$ only for actions in $\mathcal A_1$, while the initial estimate is already available to every policy, so its elicitation
cost is treated as sunk. All six revision actions correspond to conditions
present in the original experiment; we do not introduce manipulated AI predictions beyond these conditions.

The main analysis retains all 437 participants in the released table. We
also report a sensitivity analysis using the original filtering rules, which
exclude participants who never revise, have long runs of repeated answers,
or give extreme overestimates in both rounds. These rules retain 421
participants and 5,052 paired judgments. We use the unfiltered population
for the main analysis because these exclusions depend partly on assisted
responses.

\subsection{Omnimechanism Design}
Let $v$ denote an action's final price estimate. Following \citet{guo2025value}, we use absolute percentage error (APE):
\begin{equation}
d_{\mathrm{APE}}(v,y)=\left|\frac{v-y}{y}\right|.
\label{eq:ames-continuous-loss}
\end{equation}
The total decision loss is
$d_{\mathrm{APE}}(v,y)+c\,r(a)$, where $r(a)=\mathbf{1}\{a\in\mathcal A_1\}$.
We report loss directly, with lower values indicating better performance.
The cost $c$ is measured in APE units.

To obtain a finite outcome representation, we discretize the signed relative
error $R=(v-y)/y$ into 13 bins. The finite breakpoints are
\begin{equation*}
\begin{split}
(-1.00,-0.75,-0.50,-0.30,-0.15,-0.05, 0.05,0.15,0.30,0.50,0.75,1.00).
\end{split}
\end{equation*}
The outer endpoints are $-\infty$ and $+\infty$, and finite bins include
their lower endpoint. For an outcome $R$, let $b(R)\in\{1,\ldots,13\}$ denote
the resulting bin index. Writing the finite breakpoints as $t_1<\cdots<t_{12}$, we use the midpoint
of each interior bin, $\rho_k=(t_{k-1}+t_k)/2$ for $k=2,\ldots,12$.
For the lower tail, we set $\rho_1=-1$ at the limiting boundary implied
by positive estimates and sale prices ($R>-1$); for the open-ended
upper bin $[1,\infty)$, we use $\rho_{13}=1.5$, the midpoint of the
reference interval $[1,2]$, as a fixed heuristic representative.
The neural omnimechanism predicts $\widehat p_\theta(k\mid z,a)$,
and the corresponding discrete loss is
$\ell_{\mathrm{APE}}(k)=|\rho_k|$. This loss approximates
Eq.~\eqref{eq:ames-continuous-loss} for action selection; final evaluation
uses the continuous errors rather than the bin representatives.




For each policy-training context and action in $\mathcal A_0$, the observed output gives a one-hot target on $b(R)$. For an action in $\mathcal A_1$, a behavioral simulator generates $B=256$ revised estimates $\widetilde H_1^{(1)}(z,a),\ldots,\widetilde H_1^{(B)}(z,a)$. We construct the averaged target for omnimechanism based on the behavioral simulator: \begin{equation}
q_k(z,a)=\frac{1}{B}\sum_{m=1}^{B}
\mathbf{1}\!\left\{
b\!\left(\frac{\widetilde H_1^{(m)}(z,a)-y}{y}\right)=k
\right\}.
\label{eq:ames-soft-target}
\end{equation}
Each context contributes one target distribution for each of the nine
actions, weighted equally during training. This produces 9,396
context--action training targets in the main split. The sale price is used
to construct these supervised targets, but is not an input to the predictor.

\subsection{Neural Omnimechanism}
We split participants into disjoint simulator-training, simulator-audit,
policy-training, and evaluation sets, stratifying by their assigned AI and
explanation condition. With seed 42, the sets contain 197, 66, 87, and 87
participants, respectively, corresponding to 2,364, 792, 1,044, and 1,044
paired judgments. All trials from a participant remain in the same split.
The split evaluates generalization to held-out participants on the same
house bank; houses are not required to be disjoint across splits.

The behavioral simulator predicts the log-relative update
$\log(H_1/H_0)$ using a two-hidden-layer MLP of width 64 with ReLU
activations and a scalar output. Its inputs are the six house features,
$\log H_0$, the logarithm of the displayed AI prediction, relative
AI--human disagreement $(M_j-H_0)/H_0$ and its magnitude, and one-hot
encodings of the model and explanation. The model is fit by squared error
on simulator-training participants. To generate stochastic responses, we
add residuals sampled from a pooled distribution obtained by four-fold
participant-grouped, condition-stratified prediction within that training
set. We transform back to dollars and clip simulated estimates to
$[1,10^6]$. The same residual draws are shared across candidate revision
actions for a context. The separate simulator-audit split is used only for
held-out diagnostics.

The neural omnimechanism takes the six house features, the logarithms of
$H_0$, $M_1$, and $M_2$, and a nine-dimensional action one-hot encoding,
giving 18 input features. Numerical context features are standardized using
training data. A two-hidden-layer MLP of width 64 with ReLU activations and
a 13-way softmax produces $\widehat p_\theta(\cdot\mid z,a)$. We minimize
soft-target cross-entropy,
\begin{equation}
\mathcal L_{\mathrm{NN}}(\theta)
=-\frac{1}{|\mathcal{A}| N_{\mathrm{train}}}
\sum_{i=1}^{N_{\mathrm{train}}}\sum_{a\in\mathcal A}
\sum_{k=1}^{13}q_k(z_i,a)\log\widehat p_\theta(k\mid z_i,a).
\label{eq:ames-cross-entropy}
\end{equation}
The simulator and omnimechanism both use AdamW with learning rate
$10^{-3}$, weight decay $10^{-4}$, and batch size 256. For each model, an
internal 20\% participant-level, condition-stratified validation split of
its training data selects the epoch with the lowest validation loss, with
at most 250 epochs and early-stopping patience 25. Standardization is fit
within the internal training split during selection. We then refit each
model and its standardization on all of its designated training data for
the selected number of epochs: 9 for the simulator and 200 for the
omnimechanism in the main run. The omnimechanism is trained once and is
reused for all costs.

\subsection{Evaluation and Additional Results}
Following the deterministic post-processing used in Experiments 1 and 2,
the same trained predictor is evaluated with every action. For each cost $c$,
it chooses
\begin{equation}
\pi_c(z)\in\arg\min_{a\in\mathcal A}
\left\{\sum_{k=1}^{13}\widehat p_\theta(k\mid z,a)\ell_{\mathrm{APE}}(k)
+c\,r(a)\right\}.
\label{eq:ames-policy}
\end{equation}
We evaluate costs
$c\in\{0,0.005,0.010,0.025,0.050,0.075,0.100\}$.
The comparator is Global Best: separately for each cost, we use
policy-training contexts to select the single action with the lowest mean
continuous loss plus cost, evaluating revision actions with simulator
rollouts. That action is then applied to every evaluation context. Global
Best selects $\textsc{AI}_2$ at all tested costs in the
main run. Both policies are fixed before evaluation.

Simulator evaluation uses the observed outputs for $\mathcal A_0$ and
1,024 fresh rollouts per context and revision action for $\mathcal A_1$.
It scores the selected outputs by Eq.~\eqref{eq:ames-continuous-loss},
independently of the omnimechanism's predicted bin probabilities. We also
evaluate the policies on held-out logged human responses. Each row reveals
only one of the six assisted outcomes, so a selected revision can be scored
directly only when its action matches the participant's logged condition.
Let $A_i$ be the logged action, $a_i=\pi_c(z_i)$ the selected
action, and $v_i(a)$ the known output for $a\in\mathcal A_0$. Using the
nominal equal-assignment propensity $e_a=1/6$ for revision actions, the
logged estimate is
\begin{equation}
\begin{split}
\widehat d_i(a)
&=\begin{cases}
d_{\mathrm{APE}}(v_i(a),y_i), & a\in\mathcal A_0,\\
\dfrac{\mathbf{1}\{A_i=a\}}{e_a}\,d_{\mathrm{APE}}(H_{1,i},y_i),
& a\in\mathcal A_1,
\end{cases}\\
\widehat L_c(\pi)
&=\frac{1}{N}\sum_{i=1}^{N}
\left[\widehat d_i(a_i)+c\,r(a_i)\right].
\end{split}
\label{eq:ames-ipw}
\end{equation}
The cost is charged whenever revision is selected, including unmatched
rows. We use 2,000 paired participant-bootstrap replicates for uncertainty
intervals, keeping each participant's 12 judgments together. These intervals
condition on the fitted models and the observed house bank. Since treatment
is assigned once per participant, interpreting the logged estimates for
instance-dependent intervention choices requires no cross-trial carryover
within the assisted round.

\begin{table}[!t]
\centering
\caption{AMES evaluation loss (APE plus marginal revision cost) on 87
held-out participants and 1,044 paired judgments. Sim. uses
simulator rollouts; Logged uses Eq.~\ref{eq:ames-ipw}. Global Best (GB)
selects $\textsc{AI}_2$, so its simulated and logged losses coincide.
Lower values are better.}
\label{tab:ames-results}
\vspace{0.4em}
\renewcommand{\arraystretch}{0.95}
\setlength{\tabcolsep}{7pt}
\begin{tabular}{crrr}
\toprule
$c$ & NN Sim. & NN Logged & GB \\
\midrule
0.000 & 0.1053 & 0.1061 & 0.1335 \\
0.005 & 0.1054 & 0.1065 & 0.1335 \\
0.010 & 0.1054 & 0.1055 & 0.1335 \\
0.025 & 0.1057 & 0.1058 & 0.1335 \\
0.050 & 0.1057 & 0.1066 & 0.1335 \\
0.075 & 0.1057 & 0.1065 & 0.1335 \\
0.100 & 0.1056 & 0.1056 & 0.1335 \\
\bottomrule
\end{tabular}
\vspace{-0.5em}
\end{table}

Table~\ref{tab:ames-results} reports simulator and logged performance.
At zero cost, the NN policy achieves APE 0.1053 under simulator evaluation
and 0.1061 under logged evaluation, compared with 0.1335 for Global Best.
The paired logged APE difference, NN minus Global Best, is $-0.0274$
with a 95\% interval of $[-0.0346,-0.0189]$.

\begin{table}[!t]
\centering
\caption{Actions selected by the AMES policy on the 1,044 evaluation
contexts. Rev. aggregates the six revision actions. Matched counts selected
revisions whose action equals the logged condition; it is a subset of Rev.}
\label{tab:ames-selection}
\vspace{0.4em}
\renewcommand{\arraystretch}{0.95}
\setlength{\tabcolsep}{7pt}
\begin{tabular}{crrrrr}
\toprule
$c$ & $\textsc{H}_0$ & $\textsc{AI}_1$ & $\textsc{AI}_2$ & Rev. & Matched \\
\midrule
0.000 & 67 & 473 & 475 & 29 & 5 \\
0.005 & 69 & 473 & 475 & 27 & 5 \\
0.010 & 69 & 477 & 481 & 17 & 3 \\
0.025 & 70 & 478 & 484 & 12 & 2 \\
0.050 & 70 & 480 & 486 & 8 & 2 \\
0.075 & 70 & 481 & 491 & 2 & 1 \\
0.100 & 70 & 481 & 493 & 0 & 0 \\
\bottomrule
\end{tabular}
\vspace{-0.5em}
\end{table}

Table~\ref{tab:ames-selection} gives the action composition underlying
these results. At zero cost, the NN policy requests revision on 29
contexts (2.78\%). Additional human involvement decreases with cost and
disappears at $c=0.10$. The policy can still use the already-observed initial
human estimate at this cost. Restricting the
same predictor to $\mathcal A_0$ gives zero-cost logged APE 0.1056;
the full policy's difference is $+0.0005$, with a paired 95\% interval
of $[-0.0047,0.0080]$. Thus the primary-run gain over Global Best is
largely explained by input-dependent routing among the available estimates.

We repeat the pipeline with seeds 7 and 2026 to assess sensitivity to
participant splitting and training. At zero cost, logged APE is 0.1388
and 0.1026, respectively, compared with Global Best values 0.1377 and
0.1329. Under the 421-participant filtering sensitivity
with seed 42, logged APE is 0.1117 compared with 0.1357 for Global Best.
These checks show that the primary-run improvement is not uniform across
splits. The finite-bin
approximation and limited factual support for selected revisions are
therefore relevant when interpreting the simulator and logged results.

\clearpage




\end{document}